\documentclass[]{bytedance_seed}
\usepackage[toc,page,header]{appendix}

\usepackage{minitoc}
\usepackage{amsfonts}
\usepackage{amssymb}
\usepackage{tabularx}
\usepackage{listings}
\usepackage{xcolor}
\usepackage{cancel}
\newtheorem{theorem}{Theorem}
\newtheorem{lemma}{Lemma}

\definecolor{codegreen}{rgb}{0,0.6,0}
\definecolor{codegray}{rgb}{0.5,0.5,0.5}
\definecolor{codepurple}{rgb}{0.58,0,0.82}
\definecolor{backcolour}{rgb}{0.95,0.95,0.92}
\definecolor{boxblue}{RGB}{57,89,163}
\definecolor{boxbluebg}{RGB}{230,237,250} 

\lstdefinestyle{mystyle}{
    backgroundcolor=\color{backcolour},   
    commentstyle=\color{codegreen},
    keywordstyle=\color{magenta},
    numberstyle=\tiny\color{codegray},
    stringstyle=\color{codepurple},
    basicstyle=\ttfamily\footnotesize,
    breakatwhitespace=false,         
    breaklines=true,                 
    captionpos=b,                    
    keepspaces=true,                 
    numbers=none,                    
    numbersep=5pt,                  
    showspaces=false,                
    showstringspaces=false,
    showtabs=false,                  
    tabsize=2
}
\newcommand{\vect}[1]{\bm{#1}}

\title{Understanding Transformer from the Perspective of Associative Memory}

\author[*, \dagger]{Shu Zhong}
\author[*]{Mingyu Xu}
\author[*]{Tenglong Ao}
\author[\dagger]{Guang Shi}

\affiliation[]{ByteDance Seed}

\contribution[*]{Equal contribution}
\contribution[\dagger]{Corresponding authors}

\abstract{
In this paper, we share our reflections and insights on understanding Transformer architectures through the lens of associative memory—a classic psychological concept inspired by human cognition. We start with the basics of associative memory (think simple linear attention) and then dive into two dimensions:

\textbf{Memory Capacity}: How much can a Transformer really remember, and how well? We introduce retrieval SNR to measure this and use a kernel perspective to mathematically reveal why Softmax Attention is so effective. We also show how FFNs can be seen as a type of associative memory, leading to insights on their design and potential improvements.

\textbf{Memory Update}: How do these memories learn and evolve? We present a unified framework for understanding how different Transformer variants (like DeltaNet and Softmax Attention) update their "knowledge base". This leads us to tackle two provocative questions: 1. \textit{Are Transformers fundamentally limited in what they can express, and can we break these barriers? } 2. \textit{If a Transformer had infinite context, would it become infinitely intelligent?}

We want to demystify Transformer architecture, offering a clearer understanding of existing designs. This exploration aims to provide fresh insights and spark new avenues for Transformer innovation.
}

\date{\today}
\correspondence{Shu Zhong at \email{zhongshu@bytedance.com}, Guang Shi at \email{shiguang@bytedance.com}}

\begin{document}
\maketitle

\section{Introduction}
Transformer models~\cite{vaswani2017attention} have emerged as foundational architectures in modern artificial intelligence (AI), demonstrating remarkable versatility and groundbreaking performance across diverse domains, including natural language processing~\cite{brown2020gpt3}, computational biology~\cite{jumper2021alphafold}, computer vision~\cite{dosovitskiy2020vit}, and multimodal tasks~\cite{radford2021clip, radford2023whisper}. Despite their evident success, the internal mechanisms underlying Transformers remain complex and not fully understood, posing significant challenges for theoretical interpretation and principled improvements.

In our journey towards demystifying Transformer architectures, we have discovered that the concept of associative memory, drawn from human cognition, provides a uniquely intuitive and insightful perspective. Associative memory, in psychological terms, describes the brain's ability to form and recall connections between seemingly unrelated entities~\cite{suzuki2005associative}.

Interestingly, we find that Transformer architectures—including their core components, the Self-Attention mechanism and Feed-Forward Networks (FFNs)—can be elegantly understood as implementations of associative memory. More specifically, the Attention mechanism dynamically forms short-term, contextual associations, while FFNs encode persistent, long-term associative memories distilled during training. This unified viewpoint enables a deeper, more structured understanding of Transformer behavior and offers practical insights for architectural innovation.

Our analysis begins with the classical cumulative outer-product model of associative memory, essentially the simplest representation of linear attention mechanisms. From this starting point, we systematically examine the Transformer architecture along two fundamental dimensions: {memory capacity} and {memory update}. For each dimension, we clarify how existing designs fit into a generalized associative memory framework and highlight theoretical and empirical differences. Through this lens, we also uncover opportunities for novel model designs and improvements.

\textbf{Memory Capacity.} To understand and quantify memory capacity, we introduce the retrieval Signal-to-Noise Ratio (SNR) as a precise theoretical measure of associative memory retrieval quality. From a kernel perspective, we mathematically demonstrate why Softmax Attention (with exponential kernel) significantly outperforms Linear Attention in retrieval accuracy and memory capacity. Furthermore, we unify the Transformer FFN as an associative memory employing a ReLU-like kernel and analyze its properties. From this unified perspective, we raise a question:
\begin{itemize}
    \item \textit{Is it possible to achieve a unified understanding of the diverse designs for Attention and FFNs?}
\end{itemize}
By comparing Attention (exponential kernel) and FFN, we provide a principled explanation for their architectural differences, discuss empirical phenomena such as polysemanticity, and propose potential avenues for enhancing both components through kernel design, multihead mechanisms, sparsity, and gating functions.

\textbf{Memory Update.} We then explore the mechanisms by which associative memories can be updated. Starting again from the classical linear model, we propose a general recurrent form of memory update and its corresponding optimization objective. Under this generalized framework, we reinterpret various Transformer designs—including Linear Attention~\cite{katharopoulos2020linear_attn}, DeltaNet~\cite{schlag2021linear,yang2024deltanet}, Gated Attention~\cite{yang2023gated}, and Softmax Attention—as specific instances with distinct memory update strategies. 
By examining various memory update strategies, we raise the following questions:
\begin{itemize}
    \item \textit{Does Transformers suffer from an intrinsic limitation in expressive power, and can it be improved?}
    \item \textit{If a Transformer’s context length grows infinitely, will the model itself become infinitely intelligent?}
\end{itemize}
Regarding the first question, we discuss model expressivity because, when measured by circuit complexity, different memory update strategies can significantly affect a model's expressivity. For instance, DeltaNet, which employs a delta-rule update strategy, demonstrates superior expressivity compared to Transformers using Softmax Attention~\cite{grazzi2024unlocking,siems2025deltaproduct,merrill2024illusion}. Based on this insight, we introduce a novel associative memory model that combines the high retrieval accuracy of Softmax Attention with the delta-rule update mechanism of DeltaNet, thereby achieving improved memory management and stability. Through theoretical proof, we demonstrate that the proposed model's expressivity exceeds that of standard Transformers, revealing considerable potential.

For the second question, a common intuition suggests that ``Transformers supporting longer contexts can unlock greater intelligence.'' For example, reasoning language models can solve increasingly complex problems when given more time to think. This raises the question: is it theoretically feasible to build a Transformer supporting unlimited context to achieve greater potential intelligence? In-context learning actually corresponds to continuous updates of associative memory. From associative memory update perspective, we discover that theoretically, as context increases, memory may gradually converge, leading to degradation in in-context learning. This indicates potential theoretical challenges in constructing Transformers supporting infinite context. Additionally, we provide an in-depth analysis of several mitigating factors that could prevent memory convergence.

While several existing studies have explored associative-memory interpretations of Transformer architectures~\cite{ramsauer2020hopfield, wang2025test_time_regression, behrouz2025itsconnectedjourneytesttime}, we hope our exploration offers fresh insights, a clearer theoretical understanding, and inspiration for future Transformer innovations. In the conclusion section, our exploration culminates in a forward-looking discussion, where we pose several fundamental questions about the future trajectory of AI architectures, inspired by these associative memory principles.\footnote{Consider this post less as a formal research paper and more as a blog-style sharing of our current reflections, intended to spark discussion as one might in a collaborative team meeting.}
\section{Associative Memory}
\emph{What is associative memory?} According to Wikipedia~\footnote{\url{https://en.wikipedia.org/wiki/Associative_memory_(psychology)}}, associative memory refers to the ability to learn and remember relationships between different entities, even if these entities themselves are not directly related. For instance, a person who has visited Paris and seen the Eiffel Tower is likely to form an association between Paris and Eiffel Tower in the brain. Consequently, when Paris is mentioned in the future, he will likely recall the Eiffel Tower. Describing this process academically, Paris serves as a \emph{key} that triggers the recall of a corresponding \emph{value} stored in associative memory, in this case, the Eiffel Tower. This capability underpins human cognitive functions, enabling us to efficiently retrieve related information (values) from a single idea or perception (key).

\emph{How to formalize the associative memory?} Let us define a matrix $\vect{S}$ to represent associative memory. $\vect{S}$ is updated continuously and varies with time. For instance, by adding a subscript $t$ to get $\vect{S}_t$, we represent the associative memory at time $t$. We can define the ``information'' to be stored at each moment as a key-value vector pair $(\vect{k}, \vect{v})$, where $\vect{k} \in \mathbb{R}^{d_k}, \vect{v} \in \mathbb{R}^{d_v}, \vect{S} \in \mathbb{R}^{d_v \times d_k}$.

How are key-value pairs specifically memorized by $\vect{S}$? A classical modeling approach is to directly accumulate the outer product matrices of key-value pairs:
\begin{align}
    \vect{S}_{t} = \sum_{i=1}^{t} \vect{v}_i\vect{k}_{i}^\top. 
    \label{eqn:classical_mem}
\end{align}

Then how to retrieve the stored value $\vect{v}$ from memory $\vect{S}$ based on key $\vect{k}$? We can define a function $f_{\vect{S}} : \mathbb{R}^{d_k} \to \mathbb{R}^{d_v}$ parameterized by $\vect{S}$, such that $f_{\vect{S}}(\vect{k}_i)$ approximates $\vect{v}_i$ as closely as possible for any stored $(\vect{k}_i, \vect{v}_i)$. $f_{\vect{S}}$, also called the associative map, essentially simulates the recall process of memory. The recall process is typically failure tolerant~\cite{hinton2014parallel_models_associative_mem,kohonen1972correlation_maxtrix_mem}. Even if $\vect{k}_i$ is perturbed somewhat, defined as query $\vect{q}_i \approx \vect{k}_i$, $f_{\vect{S}}(\vect{q}_i)$ can still recall a similar $\vect{v}_i$. A classical $f_{\vect{S}}$ is a simple linear transformation, specifically $f_{\vect{S}}(\vect{q}) = \vect{S} \vect{q}$.

In summary, we find that an associative memory model is determined by two components: the update mechanism of $\vect{S}$ and the associative map $f_{\vect{S}}$.

\emph{Transformer can actually be interpreted as associative memory.} Taking the attention layer, a core component of Transformer, as an example, its output $\vect{o}_t$ at time $t$ can be calculated as:
\begin{align}
    \vect{o}_t = \sum_{i=1}^t \frac{\exp(\vect{k}_i^\top \vect{q}_t / \sqrt{d_k})}{\sum_{j=1}^t \exp(\vect{k}_j^\top \vect{q}_t / \sqrt{d_k})} \vect{v}_i.
\end{align}
Define the feature mapping corresponding to the exp kernel function as $\phi(\cdot)$, such that $\exp(\vect{k}^\top \vect{q} / \sqrt{d_k}) = \phi(\vect{k})^\top \phi(\vect{q})$.
For simplicity, we ignore the normalization term. The calculation of $\vect{o}_t$ can be re-written as:
\begin{align}
    \vect{o}_t &= \sum_{i=1}^t \phi(\vect{k}_i)^\top \phi(\vect{q}_t) \vect{v}_i \nonumber \\
    &= \underbrace{\sum_{i=1}^t \vect{v}_i \phi(\vect{k}_i)^\top}_{\vect{S}_t} \phi(\vect{q}_t).
\end{align}
Thus, we discover that the attention layer is a form of associative memory, where $\vect{S}_t = \sum_{i=1}^t \vect{v}_i \phi(\vect{k}_i)^\top$ and the associative map $f_{\vect{S}}(\vect{q}) = \vect{S} \phi(\vect{q})$.

In fact, another core component of Transformer, the feed-forward network (FFN), can also be viewed as a form of associative memory (see the discussion on the ReLU kernel in Sec.~\ref{subsubsec:kernel_trick}).

Although both the attention layer and the feed-forward network (FFN) maintain memory that can be categorized as associative memory, they differ in terms of the lifespan of the stored information. Specifically, the attention layer maintains a short-term contextual memory organized in an associative manner. During inference, this memory, known as the key-value (KV) cache, is discarded once inference is completed. In contrast, the FFN maintains a persistent, long-term associative memory. This memory is compressed via gradient descent during training and encodes knowledge relevant to the training dataset. Typically, it remains unchanged after training concludes.

Since Transformer belongs to a type of associative memory, we can understand Transformer and its various variants from the perspective of associative memory. As a memory, it is natural to examine its properties from two dimensions: memory capacity (Sec.~\ref{subsec:mem_cap}) and memory update (Sec.~\ref{subsec:mem_update}). To keep the discussion general, i.e., covering as many model designs as possible, for each dimension we will start from the basic model corresponding to Eq.~\ref{eqn:classical_mem}, gradually extending it and analyzing different influencing factors.

\subsection{Memory Capacity}
\label{subsec:mem_cap}
Specifically, considering $\vect{q}_t=\vect{k}_i$, retrieving any $\vect{v}_i$ is performed as:
\begin{align}
    \vect{o}_t = \vect{S}_t \vect{k}_i = \underbrace{\vect{v}_i (\vect{k}_i^\top  \vect{k}_i)}_{\text{signal term}} + \underbrace{\sum_{j \ne i} \vect{v}_j (\vect{k}_j^\top  \vect{k}_i)}_{\text{noise term}}.
    \label{eqn:associative_recall}
\end{align}
To achieve accurate retrieval, it is required that vectors $\{\vect{k}_1, \dots, \vect{k}_t\}$ be orthonormal, implying that in Eq. \ref{eqn:associative_recall}, $\vect{k}_i^\top  \vect{k}_i = 1$ and $\vect{k}_j^\top  \vect{k}_i = 0$ for $j \ne i$.

What is the capacity of the associative memory $\vect{S}_t$? That is, under the condition of accurate recall, what is the maximum number of key-value pairs it can store? Intuitively, the linear space spanned by vectors $\vect{k}_i$ has only $d_k$ orthogonal bases. If the number of stored key-value pairs $n \gg d_k$, the ``noise'' term $\sum_{j \ne i} \vect{v}_j (\vect{k}_j^\top  \vect{k}_i)$ in Eq.~\ref{eqn:associative_recall} becomes non-zero, resulting in increased retrieval error.

To quantitatively analyze the performance of fuzzy associative memory retrieval, we define a metric similar to Signal-to-Noise Ratio (SNR). Specifically, consider the general retrieval operation:
\begin{align}
    \vect{S}_t \vect{k}_i = c\vect{v}_i + \vect{r},
\end{align}
where the first term represents the signal, scaled by factor $c$, and the second term represents the noise $\vect{r}$. The inverse SNR is thus defined as:
\begin{align}
    \mathrm{SNR}^{-1} = \mathbb{E}_{\vect{v}_j, \vect{k}_j}\left[\frac{\|\vect{r}\|^2}{c^2\|\vect{v}_i\|^2}\right].
    \label{eqn:def_SNR}
\end{align}
This indicator intuitively measures the proportion of noise relative to the signal. A larger value indicates a higher noise component, resulting in lower retrieval accuracy. We consider a scenario where key vectors $\vect{k}_j \in \mathbb{R}^{d_k}$ and value vectors $\vect{v}_j \in \mathbb{R}^{d_v}$ are independently and identically distributed (i.i.d.) standard Gaussian vectors. Under this assumption, we can quantitatively calculate the inverse SNR, providing an explicit measure of retrieval accuracy degradation. 

For the linear form in Eq.~\ref{eqn:associative_recall}, the signal term is $c=\vect{k}_i^\top  \vect{k}_i$,  and the noise is given by $\vect{r}=\sum_{j \ne i} \vect{v}_j (\vect{k}_j^\top  \vect{k}_i)$. As shown in the appendix \ref{sec:linear_kernel}, this leads to the following approximate expression:
\begin{align}
    \boxed{\mathrm{SNR}^{-1}_{\text{Linear}} \approx \frac{N}{d_k}}.
    \label{eqn:linear_SNR}
\end{align}

This result intuitively indicates that the noise increases linearly with the number of stored key-value pairs $N$\footnote{We use $N$ instead of the earlier variable $t$ to emphasize the total \textit{number} of stored key-value pairs.} and decreases inversely with the dimensionality of the key vectors $d_k$. Practically,  this implies that for reliable associative recall in linear form, the number of stored pairs should be on the order of or less than the dimensionality of the key vectors: $d_k \gtrsim N$.

Beyond this threshold, retrieval accuracy deteriorates rapidly due to the accumulation of noise, severely restricting the effective capacity of the associative memory. This also explains why linear Transformers often underperform in precise long-context retrieval tasks \cite{li2025minimax,poli2024mechanistic}.  

We next introduce two strategies to improve associative memory capacity: kernel mapping and sparsity. These techniques are widely used in attention mechanisms and feedforward layers, respectively. While traditionally treated as separate design choices, we show that they are not mutually exclusive and, in fact, share substantial structural similarities. 

\subsubsection{Kernel Mapping}
\label{subsubsec:kernel_trick}
Since the memory capacity is limited by $d_k$, we can apply a kernel function to map vectors to a higher-dimensional space, allowing query-key matching to occur in this expanded space. 
Intuitively, in higher-dimensional spaces, vectors become more distinguishable from each other, making them easier to separate, reducing retrieval errors, and enabling memory to store more values $\vect{v}$. This insight is closely related to the Johnson–Lindenstrauss lemma and superposition theory \cite{henighan2023superposition, elhage2022superposition}.

By introducing a kernel function $\kappa(\vect{x},\vect{y})=\phi(\vect{x})^\top \phi(\vect{y})$, the mapping process becomes $f_{\vect{S}_{t}}(\vect{k}_t) = \vect{S}_{t}\phi(\vect{k}_t)$. Correspondingly, the associative memory and the recall operation are:
\begin{align}
    \vect{S}_t = \sum_{i=1}^{t} \vect{v}_i \phi(\vect{k}_{i})^\top ,
\end{align}
\begin{align}
    \vect{o}_t = \vect{S}_t \phi(\vect{q}_t) = \sum_{i=1}^{t} \vect{v}_i \left(\phi(\vect{k}_{i})^\top  \phi(\vect{q}_t) \right) = \sum_{i=1}^{t} \vect{v}_i \kappa(\vect{k}_{i}, \vect{q}_t) .
    \label{eqn:kernel_recall}
\end{align}

We now introduce the SNR analysis under the kernelized setting. Following the derivation from the appendix \ref{sec:inverse_snr}, we get:
\begin{align}
    \mathrm{SNR^{-1}_\kappa} =
    N\frac{
    \mathbb{E}_{k_j}[\kappa^2(\vect{k}_{j}, \vect{k}_{i})]}{\kappa^2(\vect{k}_{i}, \vect{k}_{i})}.
    \label{eqn:snr_kernel}
\end{align}
In the special case where the kernel function is linear, i.e., $\kappa(\vect{x},\vect{y})=\vect{x}^\top \vect{y}$, the kernelized formulation recovers the original linear-case inverse SNR in Eq.~\ref{eqn:linear_SNR}. Next, we analyze two kernel functions particularly significant in associative memory: the exponential (Exp) kernel and the ReLU kernel, due to their close connections to the attention mechanism and feed-forward network (FFN) commonly employed in modern transformer architectures.

\paragraph{\textbf{Exp Kernel.}} We define the exponential kernel as $\kappa_\text{exp}(\vect{x},\vect{y}) = \exp(\frac{\vect{x}^\top \vect{y}}{\tau})$, where $\tau > 1$ is a temperature parameter. Typically, $\tau$ is set to $\sqrt{d_k}$ to ensure appropriate scaling, since the dot product $\vect{x}^\top \vect{y}$ has variance $d_k$ (assuming independent components with zero mean and unit variance). With this choice, the kernel formulation directly corresponds to the standard Softmax Attention mechanism\footnote{Note that the normalization term is omitted since it does not affect the computation of the inverse SNR.}: $\sum_{i=1}^{t} \vect{v}_i \exp(\vect{k}_{i}, \vect{q}_t)$
. 
In fact, the exponential kernel admits an explicit feature mapping into an infinite-dimensional Hilbert space, which can be represented as an infinite power series\footnote{In fact, due to the limited precision, our experiments have found that this expansion practically truncates at around the 7th to 8th order, rather than continuing infinitely.}:
\begin{align}
    \phi(\cdot) = 
    \begin{bmatrix}
        1 \\
        (\cdot) \\
        \frac{(\cdot)^{\otimes 2}}{\sqrt{2!}} \\
        \vdots\\
        \frac{(\cdot)^{\otimes \infty}}{\sqrt{\infty!}}
    \end{bmatrix},
    \label{eqn:softmax_kernel}
\end{align} 
where $(\cdot)^{\otimes m}$ denotes the $m$-fold Kronecker product of the vector with itself. We omit $\tau$ here for clarity. Specifically, we have,
\begin{align}
    \vect{S}_t \phi(\vect{q}_t) 
    = \sum_{i=1}^{t} \left(\phi(\vect{q}_{j})^\top  \phi(\vect{k}_i) \right) \vect{v}_i 
    = \sum_{i=1}^{t} \sum_{m=0}^{\infty} \frac{(\vect{q}_t^{\otimes m})^\top  (\vect{k}_i^{\otimes m})}{m!} \vect{v}_i 
     = \sum_{i=1}^{t} \sum_{m=0}^{\infty} \frac{(\vect{q}_t^\top  \vect{k}_i)^m}{m!} \vect{v}_i 
    = \sum_{i=1}^{t} \exp(\vect{q}_t^\top  \vect{k}_i) \vect{v}_i.
    \label{eqn:exp_taylor_expanding}
\end{align} 

Under the same assumption that keys and values are i.i.d. standard Gaussian vectors, as derived in the appendix \ref{sec:exp_kernel}, the inverse SNR of exp kernel can be approximated as:
\begin{align}
    \boxed{
    \mathrm{SNR^{-1}_{\text{exp}}} \approx
    \frac{N}{\exp(\frac{2(\tau-1)}{\tau^2}d_k)}.
}
\end{align}

This result shows that exponential kernel significantly reduces the required feature dimensionality $d_k$ for accurate retrieval: from \(\mathcal{O}(N)\) in the linear case (Eq.~\ref{eqn:linear_SNR}) to \(\mathcal{O}(\log^2 N)\) when \(\tau=\sqrt{d_k}\) with an exponential kernel. 

This theoretical finding can be supported by several experimental observations:
\begin{enumerate}
    \item \textit{Dimension Scaling in LLMs:} In practical LLM training with Softmax Attention, increasing dimensionality $d$ becomes necessary as context length $N$ grows to maintain retrieval accuracy, directly aligning with our derived SNR relationship \cite{xu2024kv} .
    
    \item \textit{Temperature Effects:} Reducing the temperature parameter $\tau$—while keeping it above 1—improves retrieval precision, consistent with findings by \cite{peng2023yarn,wang2024length}.
    
    \item \textit{Multi-head Efficiency Trade-off:} We hypothesize that Softmax Attention's superior retrieval capability allows it to trade dimensionality for increased expressiveness through multi-head architectures, while Linear Attention benefits more from higher dimensionality. Our preliminary experiment in Sec.~\ref{sec.capex} support this conclusion.
\end{enumerate}

\paragraph{\textbf{ReLU Kernel.}} We show that the standard feed-forward network (FFN), which consists of two linear layers without bias and a nonlinear ReLU activation can be interpreted as an associative memory defined by a ReLU kernel\footnote{Other types of FFN activation functions can be analyzed using similar methods; here we select the most representative ones. Also, we do not analyse the SiLU activation separately, as its behavior is similar to ReLU and does not affect our conclusions.}. Consider the FFN given by $\text{FFN}(\mathbf{x}) = \vect{W}_V^\top \,\text{ReLU}\Bigl(\vect{W}_K\,\mathbf{x}\Bigr),$
where \(\mathbf{x} \in \mathbb{R}^{d}\) is the input, \(\vect{W}_K,\vect{W}_V \in \mathbb{R}^{m \times d}\) are weight matrices, and ReLU function is applied element-wise: $\text{ReLU}(z)=\max\{0,z\}.$
With the above formulation, we can directly map the FFN to an associative memory:
\begin{align}
    \text{FFN}(\vect{x}) 
    &= \vect{W}_V^\top \,\text{ReLU}\Bigl(\vect{W}_K\,\vect{x}\Bigr) \notag \\
    &= \sum_{i=1}^{m} \underbrace{{\vect{W}^\top_{V_i}}\vphantom{{\vect{W}_K}_{i}}}_{\vect{v}_i} \underbrace{\text{ReLU}\vphantom{{\vect{W}_K}_{i}}}_{\kappa(\cdot,\cdot)}\bigl(
    \underbrace{{\vect{W}_{K_i}}}_{\vect{k}_i} \underbrace{\vect{x}\vphantom{{\vect{W}_K}_{i}}}_{\vect{q}_t}\bigr).
    \label{eqn:relu_recall}
\end{align}

which is consistent with Eq.~\ref{eqn:kernel_recall}, where the rows in two linear layers \({W_U}_i, {W_D}_i\) define $m$ memory keys $\vect{k_i}$ and values $\vect{v_i}$, and the kernel function $\kappa_\text{ReLU}(\vect{x}, \vect{y}) = \text{ReLU}(\vect{x}^\top \vect{y})$. 
From this perspective, we can observe that the memory mechanisms underlying FFN and Attention are fundamentally unified. \emph{The key distinction is that Attention maintains context-dependent associative memories (e.g., the surname and given name of a new person in context), while FFNs encode persistent memories that reflect more general world knowledge (e.g., the name of a famous celebrity).} This structural similarity suggests that their designs can be mutually inspired and has already been recognized~\cite{vaswani2017attention}. 
We will discuss this potential integrations in the subsequent section.

Following the derivation presented in the appendix \ref{sec:relu_kernel}, we approximate the inverse SNR for the ReLU kernel as:
\begin{align}
    \boxed{
    \mathrm{SNR}^{-1}_{\text{ReLU}} \approx \frac{N}{2 d_k}
    }.
\end{align}
Compared to the linear case (Eq.~\ref{eqn:linear_SNR}), the ReLU kernel improves retrieval performance by suppressing negative-similarity query-key pairs to zero. However, its precision still falls significantly short of the exponential kernel employed in Attention mechanisms. We will discuss this next. 

\subsubsection{Vignette 1: Rethinking Attention and FFN from the Perspective of Associative Memory}

\begin{table}[h]
\centering
\caption{Comparison of typical architectural components between Attention and FFN.}
\begin{tabular}{l|c|c}
\toprule
Component     & Attention & FFN \\
\midrule
Kernel        & Exp & ReLU / SiLU \\
Normalization & Yes & None \\
Multihead     & Yes & None \\
Sparsity      & None & MoE \\
Gating          & None & SwiGLU Gating \\
\bottomrule
\end{tabular}
\label{tab:attn_ffn}
\end{table}

The symmetry between Attention and FFN in Transformers is intriguing, because—at least in principle—any design for one of these modules can be implemented exactly in the other. This symmetry compels us to delve more deeply into their differences and, at the same time, inspires us to discover new architectural designs.

Tab.~\ref{tab:attn_ffn} summarizes several of the most common, classic designs shared by the two modules. In what follows we focus on their differences at the kernel, and then briefly touch on normalization, multihead, sparsity, and gating. A deeper exploration of these topics is left to future work; we also encourage readers to ponder why such differences exist and whether they can be leveraged to improve both sides.

\paragraph{\textbf{Kernel}.} As mentioned above, the distinct kernels employed by the two modules raise an intriguing question: {why do feed-forward networks (FFNs) adopt the ReLU kernel instead of a more precise one like Exp kernel?}

We propose the following hypothesis:
\emph{A kernel with lower retrieval precision encourages a more polysemantic key–value memory: multiple unrelated facts can be stored under the same key space.} This is commonly referred to as superposition~\cite{henighan2023superposition, elhage2022superposition} in interpretability research. This enlarges the amount of abstract, compressed knowledge that fits into a fixed-width associative memory.
Conversely, Attention layers frequently need to pinpoint a specific token from the current context (e.g. induction heads~\cite{olsson2022context}), and therefore benefit from a highly precise, more monosemantic kernel. FFNs, in contrast, must distill the entire training corpus into static weights; the ability to superpose many fragments of information under one vector is therefore advantageous, even at the cost of noisier recall.

A piece of empirical evidence comes from Anthropic’s replacement of ReLU with the Softmax Linear Unit (SoLU) activation~\cite{elhage2022solu}: \begin{align} \kappa_{\text{SoLU}}(\vect{x},\vect{y}) = (\vect{x}^{\top}\vect{y})\exp\left(\frac{\vect{x}^{\top}\vect{y}}{\tau}\right), \end{align} which can be viewed as a multiplicative blend of the linear and exponential kernels. They observed that hidden representations became markedly \emph{monosemantic}, yet model accuracy on knowledge-heavy benchmark TriviaQA deteriorated.

In Appendix~\ref{sec:solu_kernel}, we derive the inverse SNR for the SoLU kernel: \begin{align} \boxed{ \mathrm{SNR}^{-1}_{\text{SoLU}} \approx \frac{5N}{d_k \exp(2\sqrt{d_k})}. } \end{align} This indicates that SoLU offers a higher retrieval precision than ReLU, consistent with its observed monosemantic behavior. 
Together, these findings support our broader hypothesis: the choice of kernel fundamentally mediates a tradeoff between knowledge density (amount of information stored) and retrieval precision (specificity of memory access) in neural architectures.
We leave a fuller exploration of kernel choice for knowledge retention versus contextual recall to future work.

\paragraph{\textbf{Normalization}.} 
Compared with the associative-recall formulation introduced earlier (Eq.~\ref{eqn:kernel_recall}), the standard Softmax Attention includes an additional normalization factor:
\begin{align}
\vect{S}_t \phi(\vect{q}_t)\;
=\;
\frac{\sum_{i=1}^{t} \vect{v}_i\,\exp \bigl(\vect{q}_t^\top \vect{k}_i\bigr)}
     {\sum_{i=1}^{t} \exp \bigl(\vect{q}_t^\top \vect{k}_i\bigr)},
\end{align}
that the FFN omits. The most closely related work to ours on incorporating normalization into FFNs is NormFormer~\cite{shleifer2021normformer}, which introduces a LayerNorm operation immediately after the nonlinear activation (kernel function), resulting in improved training stability. Because many linear-attention variants\cite{li2025minimax,sun2023retentive,yang2024deltanet} have shown that this factor can be replaced by a simple per-head norm, we regard normalization primarily as a mechanism for training stability rather than a fundamental design requirement. 

\paragraph{\textbf{Multihead}.} Multihead attention slice $\vect{q}, \vect{k}, \vect{v}$ into $n_h$ heads and use output matrix $\vect{W}_O \in \mathbb{R}^{d\times d_hn_h}$ to combine per-head result:
\begin{align}
    [\vect{q}_{j,1};\vect{q}_{j,2};\dots;\vect{q}_{j,n_h}] = \vect{q}_t, \\
    [\vect{k}_{i,1};\vect{k}_{i,2};\dots;\vect{k}_{i,n_h}] = \vect{k}_i, \\
    [\vect{v}_{i,1};\vect{v}_{i,2};\dots;\vect{v}_{i,n_h}] = \vect{v}_i, \\
    \vect{o}_{t,\cdot} = \sum_{i=1}^{t}\vect{v}_{i,\cdot}\,\kappa(\vect{q}_{t,\cdot},\vect{k}_{i,\cdot}
            ), \\
    \vect{o}_{t}   = \vect{W}_{O}
          [\vect{o}_{t,1};\vect{o}_{t,2};\dots;\vect{o}_{t,n_h}].
\end{align}
This slicing mechanism allows multihead attention to compute $n_h$ distinct similarity patterns for the same query-key pair. As in our earlier analysis of the exponential kernel, the multihead mechanism can trade off retrieval precision for expressivity when using stronger kernels, or alternatively, enhance superposition to improve knowledge capacity.
Moreover, from an intuitive perspective, the output projection matrix $\vect{W}_O$ in multi-head attention can reduce the noise term in the Eq.~\ref{eqn:associative_recall}
\footnote{For instance, in Multi-Query Attention (MQA), if $\vect{W}_O$ learns to compute a simple average over $n_h$ heads with shared keys and values, the variance of the noise term is reduced by a factor of $n_h$ compared to a single-head attention with the same key/value dimensionality.}.
In the original Transformer paper, the authors experimented with replacing the entire FFN layer with multihead attention\cite{vaswani2017attention}, observing some performance improvements. Given that attentions have been significantly optimized in modern GPUs\cite{dao2022flashattention,dao2023flashattention2,shah2024flashattention3}, we argue that introducing multihead mechanisms into FFNs, i.e. slicing $\vect{x}, {\vect{W}_K}_{i}, {\vect{W}_V}_i$ in Eq.~\ref{eqn:relu_recall}, deserves renewed consideration, and we leave exploration of this idea to future work.

\paragraph{\textbf{Sparsity}.} It is easy to see that when numbers of keys and values becomes large, the keys relevant to each query will inevitably be sparse.
The simplest approach to incorporating sparsity in FFN is through the Mixture-of-Experts (MoE) mechanism~\cite{shazeer2017outrageously,fedus2022switch}. Consider an FFN with $E$ expert subnetworks, each parameterized by separate weight matrices $\vect{W}_K^{(e)}, \vect{W}_V^{(e)} \in \mathbb{R}^{(m/E)\times d}$. A gating function $g_e(\vect{x})$ (typically sparse) routes the input to a small subset (often one) of these experts:
\begin{align}
    \text{FFN}_{\text{MoE}}(\vect{x}) 
    &= \sum_{e=1}^{E} g_e(\vect{x})\,
    \left(\sum_{i\in \text{Expert}(e)}{{\vect{W}^{(e)}_{V_i}}}\,\text{ReLU}({\vect{W}^{(e)}_{K_i}}\vect{x})\right).
\end{align}

To achieve similar sparsity within Attention, one can partition key-value pairs into disjoint expert subsets and implement sparse routing based on queries:
\begin{align}
    \vect{o}_{t,\text{MoE}} 
    &= \sum_{e=1}^{E} g_e(\vect{q}_t, \{\vect{k}_i\}_{i\in \text{Expert}(e)})\left(\sum_{i\in \text{Expert}(e)}\vect{v}_i\,\kappa(\vect{q}_t,\vect{k}_i)\right).
\end{align}
The core difference between Attention and FFN in the MoE mechanism is that the keys and values ($\vect{k}$ and $\vect{v}$) in Attention are dynamic, whereas the expert parameters in FFN are static. Therefore, the gating function $g_e$ in Attention needs to depend on the set of key vectors $\{\vect{k}_i\}_{i\in \text{Expert}(e)}$. For instance, the MoBA architecture~\cite{lu2025moba} defines the gating function as (normalization terms omitted for clarity):
\begin{align}
    g_e(\vect{q}_t, \{\vect{k}_i\}_{i\in \text{Expert}(e)}) 
    &= 
    \begin{cases}
        1 & s_e(\vect{q}_t) \in \text{Topk}\left(\{s_{e'}(\vect{q}_t)\mid e'\in [E]\}, k\right) \\
        0 & \text{otherwise}
    \end{cases}, \\[8pt]
    s_e(\vect{q}_t) &= \langle \vect{q}_t, \text{mean-pool}(\{\vect{k}_i\}_{i\in \text{Expert}(e)}) \rangle.
\end{align}

\paragraph{\textbf{Gating.}} Another prevalent architectural choice in FFNs is gating, as exemplified by the SwiGLU activation~\cite{shazeer2020glu}. SwiGLU gating introduces multiplicative interactions between two linear transformations of the input, defined as:
\begin{align}
    \text{FFN}_\text{SwiGLU}(\vect{x}) 
    &= \vect{W}_V^\top \bigr(\left(\vect{W}_{G}\vect{x}\right) \odot \text{Swish}\left(\vect{W}_{U}\vect{x}\right)\bigl),
\end{align}
where \(\vect{W}_K, \vect{W}_G, \vect{W}_V \in \mathbb{R}^{m\times d}\) are parameter matrices, \(\odot\) denotes element-wise multiplication, and Swish is given by \(\text{Swish}(z) = z \cdot \text{Sigmoid}(z)\) .

Expressed in the associative-memory framework, SwiGLU can be rewritten as:
\begin{align}
    \text{FFN}_\text{SwiGLU}(\vect{x}) 
    &= \sum_{i=1}^{m}
    \underbrace{{\vect{W}_V}_i}_{\vect{v}_i}
    \underbrace{({\vect{W}_G}_i^\top \vect{x})}_{g_i(\vect{x})}
    \underbrace{\text{Swish}\vphantom{{\vect{W}_K}_{i}}}_{\kappa(\cdot,\cdot)}
    \bigl(
    \underbrace{{\vect{W}_K}_i^\top}_{\vect{k}_i}
    \underbrace{\vect{x}\vphantom{{\vect{W}_K}_{i}}}_{\vect{q}_t}
    \bigr),
\end{align}
where \(({\vect{W}_G}_i^\top \vect{x})\) is the additionally introduced gating term \(g_i(\vect{x})\), which can amplify or suppress the response of query to the specific key. Intuitively, this gating mechanism provides a dynamic, input-dependent modulation.
Recent linear transformers have experimented with gating mechanisms within attention layers, such as Gated Linear Attention~\cite{yang2023gated}. The gating term is defined as (normalization terms omitted): 
\begin{align}
    \vect{o}_{t,\text{gated}} &=  \sum_{i=1}^{t} \vect{v}_i\,{g_i}(\vect{x}_{i:t})\,\kappa(\vect{q}_t, \vect{k}_i), \\
    {g_i}(\vect{x}_{i:t})&= \prod_{j=i+1}^{t} \alpha_j
\end{align}
where \(\alpha_j\) parameterized via a linear layer
followed by sigmoid on hidden stats \(\vect{x}_j\). A more recent work, FoX~\cite{lin2025fox}, also incorporates gating into Softmax attention, adopting a formulation essentially identical to the one we present here.

Interestingly, gating mechanisms in attention are usually interpreted as a form of forgetting mechanism and thus typically represented as a cumulative product of gate values between 0 and 1 (implemented via sigmoid functions). However, observing the gating in FFNs, we notice the absence of such cumulative mechanisms. We conjecture that defining the gating function simply as 
\(g_i(\vect{x}_{i:t}) = {\vect{W}_G(\vect{x}_i)}^\top \vect{x}_t\) may achieve a similar effect while being simpler to implement. We leave the investigation of this hypothesis for future work. As for why gating mechanisms are effective in general, we encourage further research and exploration into their underlying principles and theoretical explanations.

\subsection{Memory Update}
\label{subsec:mem_update}
\emph{How is associative memory updated?} For example, by reformulating Eq.~\ref{eqn:classical_mem}, we obtain:
\begin{align}
    \vect{S}_{t} &= \underbrace{\left( \sum_{i=1}^{t-1} \vect{v}_i \vect{k}_i^\top \right)}_{\vect{S}_{t-1}} + \vect{v}_t \vect{k}_t^\top \nonumber \\
    &= \vect{S}_{t-1} + \vect{v}_t \vect{k}_t^\top.
    \label{eqn:classical_mem_update}
\end{align}
This indicates that, for classical associative memory, the update at time step $t$ is directly realized by adding the outer product of the latest vectors $\vect{v}_t$ and $\vect{k}_t$. We typically refer to Eq.~\ref{eqn:classical_mem_update} as the recurrent form of the classical associative memory. If we interpret this equation as a single-step gradient descent update on $\vect{S}$ (ignoring the learning rate), we have:
\begin{align}
    \vect{S}_{t} = \vect{S}_{t-1} - \underbrace{(-\vect{v}_t \vect{k}_t^\top)}_{\frac{\partial \mathcal{L}_t}{\partial \vect{S}_{t-1}}},
\end{align}
where $\mathcal{L}_t$ represents the optimization objective corresponding to the memory update, which can be derived explicitly:
\begin{align}
    \mathcal{L}_t(\vect{S}_{t-1}) = -\langle \vect{S}_{t-1} \vect{k}_t, \vect{v}_t\rangle,
    \label{eqn:inner_product_loss}
\end{align}
with $\langle \cdot, \cdot \rangle$ denoting the inner product. Intuitively, this objective aims to update $\vect{S}$ such that recalling $\vect{v}_t$ from the updated memory using $\vect{k}_t$ is as effective as possible. However, a clear drawback of this objective is that, besides promoting a large inner product between $\vect{S}\vect{k}_t$ and $\vect{v}_t$, it also encourages the norm of $\vect{S}\vect{k}_t$ to grow without bound, potentially causing numerical instability. In practice, this issue can be mitigated by, for instance, introducing decay mechanisms on $\vect{S}$~\cite{yang2023gated}. Various improved approaches to memory updating will be discussed in detail below.

To facilitate discussions on different memory update methods, we first define a general recurrent form of associative memory as follows:
\begin{align}
    \vect{S}_{t} = \vect{A}_t \vect{S}_{t-1} \vect{B}_t + \vect{C}_t,
    \label{eqn:general_recurrent_form}
\end{align}
where $\vect{A}_t$, $\vect{B}_t$, and $\vect{C}_t$ are parameter matrices. For example, Eq.~\ref{eqn:classical_mem_update} is a special case of Eq.~\ref{eqn:general_recurrent_form} when $\vect{A}_t = \vect{I}$, $\vect{B}_t = \vect{I}$, and $\vect{C}_t = \vect{v}_t \vect{k}_t^\top$. What optimization objective corresponds to Eq.~\ref{eqn:general_recurrent_form}? Referring to Appendix~\ref{subsec:general_ass_mem_opt_obj}, we can construct the associated objective function as:
\begin{align}
    \mathcal{L}_{t}(\vect{S}_{t-1}) = \frac{1}{2}\operatorname{tr}(\vect{S}_{t-1}^\top \vect{S}_{t-1}) - \frac{1}{2}\operatorname{tr}(\vect{S}_{t-1}^\top \vect{A}_t \vect{S}_{t-1}\vect{B}_t) - \operatorname{tr}(\vect{C}_t^\top \vect{S}_{t-1}),
    \label{eqn:general_loss}
\end{align}
where $\operatorname{tr}(\cdot)$ denotes the trace operation. Note that this construction has a constraint: matrices $\vect{A}_t$ and $\vect{B}_t$ must be symmetric. By specifying different combinations of $\vect{A}_t$, $\vect{B}_t$, and $\vect{C}_t$, we obtain distinct recurrent forms and corresponding optimization objectives. Each combination corresponds to a novel associative memory model, enabling us to interpret these models explicitly from the perspective of memory updating.

\begin{table}[t]
    \centering
    \caption{Different forms of memory update. Different associative memory models, their corresponding parameter matrices ($\vect{A}_t$, $\vect{B}_t$, and $\vect{C}_t$) in recurrent form (Eq.~\ref{eqn:general_recurrent_form}), and optimization objectives $\mathcal{L}_t(\vect{S}_{t-1})$. Detailed derivations can be found in Appendix~\ref{subsec:ass_mem_opt_obj_cases}.}
    \label{tab:mem_update_forms}

    \renewcommand{\arraystretch}{1.8} 
    \newcolumntype{X}{>{\raggedright\arraybackslash}X}
    \newcolumntype{Y}{>{\raggedleft\arraybackslash}X}
    \newcolumntype{Z}{>{\centering\arraybackslash}X}
    \begin{tabular}{l|l|l|l|l}
        \toprule
        Model & $\vect{A}_t$ & $\vect{B}_t$ & $\vect{C}_t$ & $\mathcal{L}_t(\vect{S}_{t-1})$ \\
        \toprule
        Linear Attention~\cite{katharopoulos2020linear_attn} & $\vect{I}$ & $\vect{I}$ & $\vect{v}_t \vect{k}_t^\top$ & $-\langle \vect{S}_{t-1}\vect{k}_t, \vect{v}_t\rangle$ \\
        \hline
        \multirow{2}{*}{\parbox{2.8cm}{Gated Linear \\ Attention~\cite{yang2023gated}}} & \multirow{2}{*}{$\text{diag}(\vect{\lambda}_t)$} & \multirow{2}{*}{$\vect{I}$} & \multirow{2}{*}{$\vect{v}_t \vect{k}_t^\top$} & $-\langle \vect{S}_{t-1}\vect{k}_t, \vect{v}_t\rangle$ \\
        & & & & $+ \frac{1}{2}\|\text{diag}(\sqrt{1-\vect{\lambda}_t})\vect{S}_{t-1}\|_F^2$ \\
        \hline
        DeltaNet~\cite{schlag2021linear,yang2024deltanet} & $\vect{I}$ & $\vect{I} - \vect{k}_t \vect{k}_t^\top$ & $\vect{v}_t \vect{k}_t^\top$ & $\frac{1}{2}\|\vect{S}_{t-1} \vect{k}_t - \vect{v}_t\|^2$ \\
        \hline
        \multirow{3}{*}{\parbox{2.8cm}{DeltaNet\\ $+$ Momentum}} & \multirow{3}{*}{$\vect{I}$} & \multirow{3}{*}{$\vect{I} - \vect{k}_t \vect{k}_t^\top$} & 
        \multirow{3}{*}{\parbox{4cm}{
        $\eta_t \vect{C}_{t-1} + \vect{v}_t \vect{k}_t^\top = $ \vspace{0.2cm} \\
        $\displaystyle\sum\limits_{i=1}^{t} \left( \prod\limits_{j=i+1}^{t} \eta_j \right) \vect{v}_i \vect{k}_i^\top$
        }} 
        & 
        \multirow{3}{*}{\parbox{3.8cm}{
        $\displaystyle\tfrac{1}{2}\|\vect{S}_{t-1} \vect{k}_t\|^2$ \vspace{0.2cm} \\
        $- \displaystyle\sum\limits_{i=1}^{t} \left(\prod\limits_{j=i+1}^{t} \eta_j\right) \vect{v}_i^\top \vect{S}_{t-1} \vect{k}_i$
        }} \\
        & & & & \\
        & & & & \\
        \hline
        \multirow{2}{*}{\parbox{2.8cm}{Softmax Attention\\ w/o Norm}}
         & \multirow{2}{*}{$\vect{I}$} & \multirow{2}{*}{$\vect{I}$} & \multirow{2}{*}{$\vect{v}_t \phi(\vect{k}_t)^\top$} & \multirow{2}{*}{$-\langle \vect{S}_{t-1}\phi(\vect{k}_t), \vect{v}_t\rangle$} \\ & & & & \\
        \hline
        \multirow{2}{*}{\parbox{2.8cm}{Softmax Attention\\ w/ Norm$^*$~\cite{vaswani2017transformer}}} & \multirow{2}{*}{$\frac{t-1}{t}\vect{I}$} & \multirow{2}{*}{$\vect{I}$} & \multirow{2}{*}{$\frac{1}{t}\vect{v}_{t} \phi(\vect{k}_{t})^\top$} & $- \frac{1}{t} \langle \vect{S}_{t-1}\phi(\vect{k}_{t}), \vect{v}_{t}\rangle$ \\
        & & & & $+ \frac{1}{2t} \|\vect{S}_{t-1}\|^2_F$ \\
        \hline
        \multirow{2}{*}{\parbox{2.8cm}{Gated Softmax\\ Attention$^*$~\cite{lin2025fox}}} & \multirow{2}{*}{$\frac{t-1}{t}\text{diag}(\vect{\lambda}_{t})$} & \multirow{2}{*}{$\vect{I}$} & \multirow{2}{*}{$\frac{1}{t}\vect{v}_{t} \phi(\vect{k}_{t})^\top$} & $- \frac{1}{t} \langle \vect{S}_{t-1}\phi(\vect{k}_{t}), \vect{v}_{t}\rangle$ \\
        & & & & $+ \frac{1}{2t} \|\text{diag}(\sqrt{1-\vect{\lambda}_{t}})\vect{S}_{t-1}\|^2_F$ \\
        \bottomrule
    \end{tabular}

    \vspace{0.5em}
    \begin{flushleft}
        {\footnotesize Both any element of $\vect{\lambda}$ and $\eta$ $ \in (0, 1)$.} \\
        {\footnotesize ``Softmax Attention'' refers to the vanilla causal softmax attention, excluding variants such as sparse softmax attention. ``Norm'' denotes the normalization factor of the softmax operation: $\sum_{i} \exp(\vect{q}_t^\top \vect{k}_i)$. The feature mapping of softmax attention $\phi(\cdot)$ is Eq.~\ref{eqn:softmax_kernel}.} \\
        {\footnotesize * $t$ is sufficiently large.}
    \end{flushleft}
\end{table}

Table~\ref{tab:mem_update_forms} summarizes various associative memory models, their corresponding parameter matrices ($\vect{A}_t$, $\vect{B}_t$, and $\vect{C}_t$) in recurrent form (Eq.~\ref{eqn:general_recurrent_form}), and optimization objectives $\mathcal{L}_t(\vect{S}_{t-1})$.
\begin{itemize}
    \item \textbf{Linear Attention}. The optimization objective of linear attention is unbounded, which may lead to numerical instability due to the exploding norm of $\vect{S}\vect{k}$. Gated linear attention~\cite{yang2023gated} introduces a memory decay term $\text{diag}(\vect{\lambda})$, which corresponds to adding a regularization term on the Frobenius norm of $\vect{S}$ in the optimization objective. This mitigates the exploding norm issue at the expense of memory forgetting.

    \item \textbf{DeltaNet}. DeltaNet~\cite{schlag2021linear,yang2024deltanet} replaces the inner-product objective with an L2 regression objective, inherently imposing norm regularization on $\vect{S}\vect{k}$. In recurrent form, this corresponds to introducing a new decay term $\vect{I} - \vect{k}\vect{k}^\top$. The effect of this decay is to retrieve and attenuate historical values associated with similar keys before memory write-in, thus preventing norm explosion after updates. Additionally, first-order momentum can be incorporated into key-value pair updates to adjust the memory updating dynamics. \citet{behrouz2024titans} found that incorporating momentum positively impacts model performance.\footnote{The model in \cite{behrouz2024titans} is not strictly equivalent to DeltaNet; thus, this finding should be considered only as a reference.}

    \item \textbf{Softmax Attention}. Softmax attention~\cite{vaswani2017transformer} improves memory recall performance by introducing an exponential kernel. Without the normalization term in the softmax operation, the optimization objective remains unbounded. With the normalization term, for sufficiently long contexts (large $t$), according to the derivations in Appendix~\ref{subsec:ass_mem_opt_obj_cases}, the objective includes a regularization on the Frobenius norm of $\vect{S}$, which alleviates numerical instability. Additionally, the objective approximately introduces a factor of $\frac{1}{t}$. As $t$ increases, we have $\frac{1}{t} \to 0$, causing $\mathcal{L}_t \to 0$ and consequently leading to the gradual vanishing of gradients with respect to $\vect{S}$. This may suppress memory updates (discussed in detail in Sec.~\ref{subsubsec:infinite_context}). Similar to gated linear attention, we can also introduce memory decay for softmax attention by incorporating a gating factor $\text{diag}(\vect{\lambda}_t)$ into the parameter matrix $\vect{A}_t$. \citet{lin2025fox} found that this improves the performance of softmax attention.
\end{itemize}

\subsubsection{Vignette 2: Softmax Attention with Delta Rule}
\label{subsubsec:delta_attn}
What insights can we gain from the comparison of different associative memory models summarized in Table~\ref{tab:mem_update_forms}? Two core points underpin these distinct models:
\begin{itemize}
    \item \emph{How to improve the accuracy of memory recall?} One effective approach is the kernel trick. For instance, softmax attention employs an exponential kernel function. As discussed in Sec.~\ref{subsubsec:kernel_trick}, this method efficiently enhances memory recall accuracy by performing additional kernel computations without increasing spatial complexity.

    \item \emph{How to effectively manage memory update to avoid such as the explosive growth of the spectral norm of $\vect{S}$?} It can be mathematically proved that memory models of the form $\vect{S}_t = \vect{S}_{t-1} + \vect{v}_i\vect{k}_i^\top$ inevitably experience growth in the spectral norm of $\vect{S}$ over time, resulting in numerical instability, regardless of whether the kernel trick is utilized. 
    
    Softmax attention addresses this issue by incorporating a normalization term, which is approximately equivalent to scaling the newly written information $\vect{v}_t$ by a decaying factor $\frac{1}{t}$. This implies that information closer to the future is increasingly attenuated, i.e., the write-in term $\frac{1}{t} \vect{v}_t \vect{k}_t^\top \to 0$, thereby slowing down the growth of the spectral norm of $\vect{S}$. The assumption that information closer to the future is less important is unreasonable.
    
    Generally, it is more reasonable to assume that older information is less important. Therefore, reversing the decay direction makes more sense. Gated linear attention implements this reversed decay by introducing an exponential decay factor $\text{diag}(\vect{\lambda})$ in its recurrent form. 
    
    Nevertheless, older information is not necessarily always less relevant. A smarter approach would be first to assess whether the new value $\vect{v}_t$ overlaps with historical information, erase the overlapping parts accordingly, and then write the new value into memory. This strategy, referred to as the delta rule, is precisely the memory update mechanism employed by DeltaNet. 
\end{itemize}

Inspired by the discussion above, \emph{can we propose a new associative memory model that integrates the advantages of both the kernel trick and the delta rule?} Such a model would achieve more intelligent memory update management while simultaneously ensuring high memory recall accuracy.

To achieve it, a straightforward approach is to introduce the delta rule into softmax attention. As shown in Table~\ref{tab:mem_update_forms}, the core of the delta rule is the decay term $\vect{I} - \vect{k}\vect{k}^\top$ in the recurrent form. Thus, replacing $\vect{B}_t$ in the recurrent form of softmax attention and incorporating the kernel trick would successfully introduce the delta rule into softmax attention:
\begin{align}
    {\vect{S}}_t = {\vect{S}}_{t-1} \underbrace{\left({\vect{I}} - \phi(\vect{k}_t)\phi(\vect{k}_t)^\top\right)}_{\text{Delta rule decay}} + \vect{v}_t \phi(\vect{k}_t)^\top.
    \label{eqn:softmax_delta_rule_update}
\end{align}
For simplicity, the normalization term of the softmax operation is omitted here without affecting the derivation~\footnote{As indicated in Table~\ref{tab:mem_update_forms}, the normalization term of the softmax operation scales $\vect{A}_t$ and $\vect{C}_t$ in the recurrent form by a factor related to $\frac{1}{t}$ as $t$ becomes sufficiently large. This scaling factor is independent and does not influence the main derivation.}. In practice, the normalization term is still retained to avoid numerical instability. 

Eq.~\ref{eqn:softmax_delta_rule_update} appears complicated. Is it possible to construct an associative memory model that is exactly equivalent to the model represented by Eq.~\ref{eqn:softmax_delta_rule_update}, yet possesses a simpler recurrent form? Ideally, the recurrent form should resemble aligned softmax attention, i.e., $\vect{A}_t = \vect{B}_t = \vect{I}$, while $\vect{C}_t$ remains an outer-product matrix. To achieve this, we introduce a new variable $\vect{u}_t$, which depends on the historical $\{\vect{k}\}_{i=1}^t$ and $\{\vect{v}\}_{i=1}^t$. Consequently, we construct a new recurrent formulation:
\begin{align}
    \vect{S}_t = \vect{S}_{t-1} + \vect{u}_t \phi(\vect{k}_t)^\top.
\end{align}

To ensure equivalence between this memory model and the one defined by Eq.~\ref{eqn:softmax_delta_rule_update}, we solve for $\vect{u}_t$ using the method of undetermined coefficients (see Appendix~\ref{sec:appendix_delta_rule_with_kernel} for details). This yields the following condition for $\vect{u}_t$:
\begin{align}
    \vect{u}_t &= -\sum_{i=1}^{t-1} \phi(\vect{k}_i)^\top \phi(\vect{k}_t) \vect{u}_i + \vect{v}_t \nonumber \\
    &= -\sum_{i=1}^{t-1} \exp(\vect{k}_i^\top \vect{k}_t) \vect{u}_i + \vect{v}_t.
    \label{eqn:delta_attn_u}
\end{align}

Compared to softmax attention, $\vect{v}_t$ is now ``replaced'' by $\vect{u}_t$ to form an outer product with $\phi(\vect{k}_t)$ before being written into memory. Intuitively, from Eq.~\ref{eqn:delta_attn_u}, $\vect{u}_t$ is derived from $\vect{v}_t$ through an additional ``processing'' step. Specifically, the similarity between $\vect{k}_t$ and historical keys is first computed to weight the sum of historical $\{\vect{u}_i\}_{i=1}^{t-1}$, yielding $-\sum_{i=1}^{t-1} \exp(\vect{k}_i^\top \vect{k}_t) \vect{u}_i$. This term represents information in the memory that overlaps with $\vect{v}_t$. We remove this ``redundant'' information from $\vect{v}_t$ to obtain $\vect{u}_t$. This process is exactly the delta rule.

After the memory $\vect{S}$ is updated, the corresponding memory recall method remains consistent with the form of softmax attention:
\begin{align}
    \vect{o}_t = \vect{S}_t \phi(\vect{q}_t) = \sum_{i=1}^t \exp(\vect{q}_t^\top \vect{k}_i) \vect{u}_i.
    \label{eqn:delta_attn_o}
\end{align}

In summary, the new model exhibits minimal formal changes compared to softmax attention, merely replacing the original $\vect{v}$ with $\vect{u}$ after an additional "processing" step according to the delta rule. Through this slight modification, we successfully incorporate the delta rule into softmax attention.

\paragraph{\textbf{General Form.}} 
We can actually build the new model in a more general form by modifying the following components:
\begin{itemize}
    \item \emph{Kernel function.} The kernel functions used for computing $\vect{u}$ (Eq.~\ref{eqn:delta_attn_u}) and for memory retrieval (Eq.~\ref{eqn:delta_attn_o}) need not be identical; different kernels can be adopted independently.

    \item \emph{Gate.} Gates can be introduced separately into the two parts of computing $\vect{u}$ (Eq.~\ref{eqn:delta_attn_u}): the term $-\sum_{i=1}^{t-1} \phi(\vect{k}_i)^\top \phi(\vect{k}_t) \vect{u}_i$ and the vector $\vect{v}_t$.

    \item \emph{Similarity calculation.} In computing $\vect{u}$ (Eq.~\ref{eqn:delta_attn_u}), a new vector, rather than $\vect{k}_t$, can be introduced to compute similarities with $\{\vect{k}_i\}_{i=1}^{t-1}$, thereby controlling memory writing and erasure.
\end{itemize}

In summary, the general form of the new model can be written as:
\begin{align}
    \vect{u}_t &= \alpha_t \vect{v}_t - \beta_t \sum_{i=1}^{t-1} \kappa_1(\vect{k}_i, \vect{w}_t) \vect{u}_i, \\
    \vect{o}_t &= \sum_{i=1}^t \kappa_2(\vect{k}_i, \vect{q}_t) \vect{u}_i,
\end{align}
where $\kappa_1(\cdot, \cdot)$ and $\kappa_2(\cdot, \cdot)$ are independent kernel functions; $\alpha_t$ and $\beta_t$ are independent gates; and $\vect{w}_t$ is the newly introduced retrieval vector. And we call the model characterized by this general form \emph{DeltaFormer}.

In fact, both the Transformer and DeltaNet are special cases of the above general form. Specifically, if we set $\beta_t=0$, $\alpha_t=1$, and use $\vect{o}_t = \frac{\sum_{i=1}^t \exp(\vect{k}_i^\top \vect{q}_t/\sqrt{d})\vect{u}_i}{\sum_{i=1}^t \exp(\vect{k}_i^\top \vect{q}_t/\sqrt{d})}$, we recover the standard Transformer. If we set $\alpha_t=1$, $\vect{w}_t = \vect{k}_t$, and both $\kappa_1(\cdot, \cdot)$ and $\kappa_2(\cdot, \cdot)$ as linear inner-product kernels, we recover exactly the DeltaNet.

\paragraph{\textbf{Model Expressivity.}}
Intuitively, neural networks can be viewed as circuits. Their expressivity can be measured via circuit complexity. Circuit complexity studies the minimal circuit size or depth required to solve computational problems, thus characterizing the intrinsic computational difficulty involved.

If we identify the circuit complexity corresponding to a given neural network, we can determine the class of computational problems that the network can solve, thereby characterizing its expressivity.

Circuit complexity typically focuses on Boolean circuits. A Boolean circuit is a directed acyclic graph composed of logic gates. It is defined by four parameters: 1) size, the total number of gates in the circuit; 2) depth, the number of gates along the longest path from input to output; 3) fan-in, the maximum number of inputs each gate can accept (either constant or unlimited); and 4) gate types: commonly AND, OR, NOT gates.

Depending on various combinations of these parameters, circuits are usually classified into three categories: $NC$ (Nick's Class), $AC$ (Alternating Class), and $TC$ (Threshold Class). Circuit depth characterizes parallelism~\footnote{Intuitively, for a circuit with a single layer, the output can be computed in one step. However, if the circuit has $k$ layers, it requires $k$ sequential steps to obtain the output. This implies that the number of circuit layers is inversely related to the degree of parallelism.} and is important. It is explicitly indicated by a superscript $k$. For example, $NC^k$ denotes circuits of depth $\mathcal{O}(\log^k(n))$, where $n$ is the input length.

We provide formal definitions of these three circuit classes:
\begin{itemize}
    \item \emph{$NC^k$}. Circuits with polynomial size in $n$ and depth $\mathcal{O}(\log^k(n))$, using AND, OR, NOT gates with bounded fan-in (typically limited to 2 or a finite constant). Intuitively, this is analogous to hiring numerous workers to perform tasks in parallel, but each worker handles only a limited number of tasks at once (small fan-in). Classical computational problems solvable by $NC$ circuits include integer addition ($k=1$) and matrix multiplication ($k=2$).

    \item \emph{$AC^k$}. Circuits with polynomial size in $n$, depth $\mathcal{O}(\log^k(n))$, and unlimited fan-in gates (AND, OR, NOT). Intuitively, this corresponds to hiring fewer but more powerful workers, each capable of processing numerous tasks simultaneously (unlimited fan-in). These circuits can perform simple logical judgments like ``does at least one input equal 1?'' or ``do all inputs equal 1?'' within constant depth, requiring infinite fan-in to achieve these computations efficiently.

    \item \emph{$TC^k$}. Circuits with polynomial size in $n$ and depth $\mathcal{O}(\log^k(n))$, employing special threshold gates. A threshold gate counts the number of inputs equal to 1, outputting 1 if this count meets or exceeds a certain threshold, and 0 otherwise. Threshold gates have unlimited fan-in. Intuitively, these circuits are more powerful, enabling efficient "majority voting" and thus handling more complex counting and decision tasks. Classical problems solvable by $TC$ circuits include numerical comparison, counting, and majority determination.
\end{itemize}

\begin{figure}[t]
  \centering
  \includegraphics[width=0.45\textwidth]{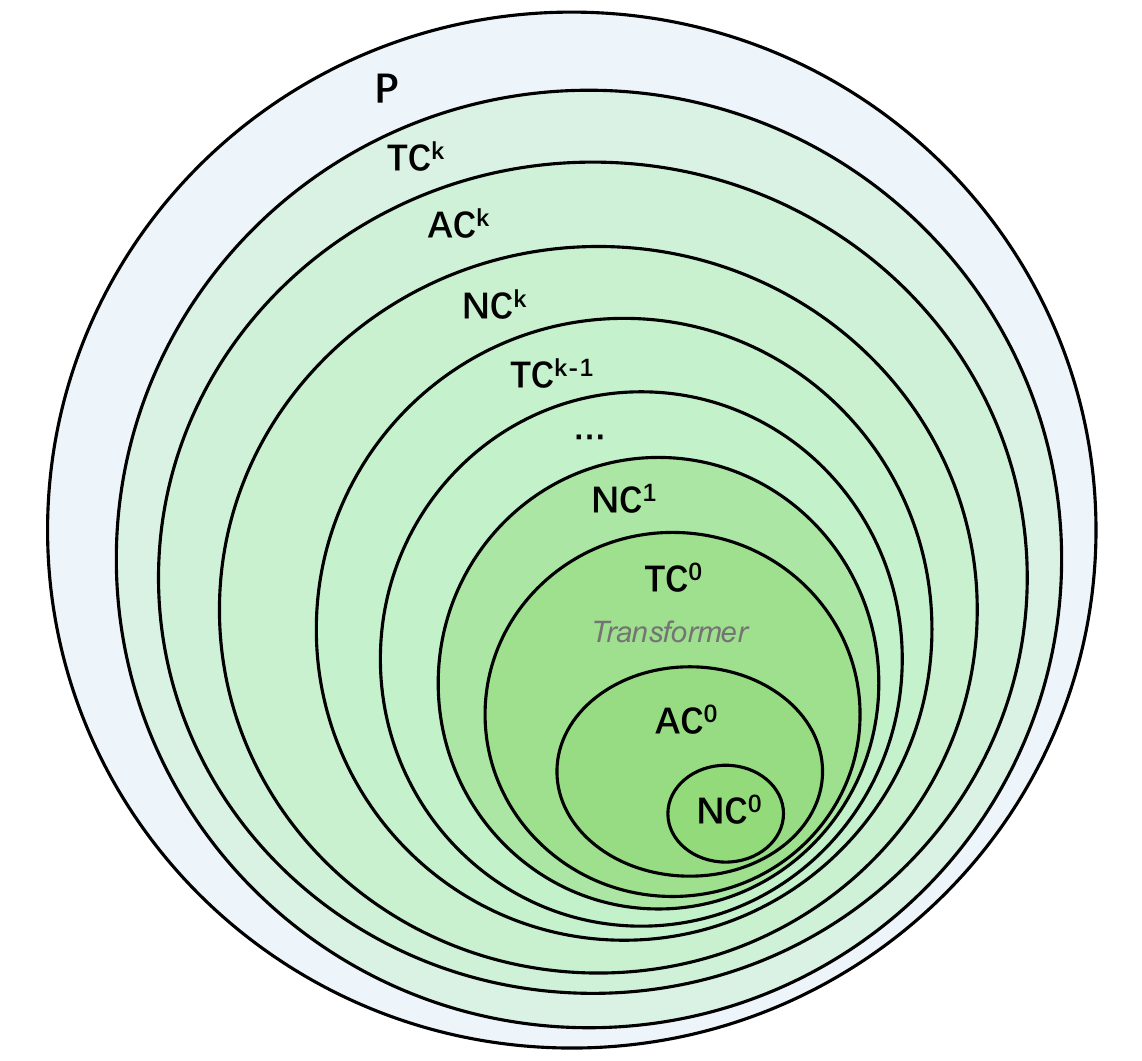} 
  \caption{The hierarchical relationship among different circuit classes. The complexity of problems solvable by the three circuit classes $NC$, $AC$, and $TC$ is contained within $P$. Specifically, we have: $NC^0 \subset AC^0 \subset TC^0 \subseteq NC^1 \subseteq AC^1 \subseteq TC^1 \subseteq NC^2 \subseteq \cdots \subseteq P.$}
  \label{fig:circuit_complexity}
\end{figure}

Due to the difference in fan-in restrictions (unlimited fan-in in $AC$ versus limited fan-in in $NC$) and the more expressive threshold gates in $TC$ as compared to $AC$, we arrive at the following hierarchy among the three circuit classes (Also demonstrated in Figure~\ref{fig:circuit_complexity}.):
\begin{align}
    NC^0 \subset AC^0 \subset TC^0 \subseteq NC^1 \subseteq AC^1 \subseteq TC^1 \subseteq NC^2 \subseteq \cdots \subseteq P.
\end{align}

To gain intuitive clarity about the relationships between these three circuit classes, consider two illustrative examples:
\begin{itemize}
    \item \emph{$AC^0$ vs. $TC^0$: Can we compute the sum of $n$ numbers in a single step?} Despite unlimited fan-in, $AC^0$ can only perform simple logical checks, such as "is at least one input equal to 1?", yet it cannot directly count how many inputs equal 1. Hence, $AC^0$ cannot compute the sum of $n$ numbers in a single step. In contrast, $TC^0$ introduces threshold gates, enabling counting of many bits (0 or 1) within constant depth, thus allowing summation of multiple numbers in a single (constant-depth) step.

    \item \emph{$TC^0$ vs. $NC^1$: Can we perform arbitrary swaps within a sequence of arbitrary length?} A key limitation of $TC^0$ is its constant depth: each output can only depend on information within a constant-sized neighborhood of the input, making arbitrary-length element swaps impossible. Conversely, $NC^1$ circuits have logarithmic depth, allowing the construction of a binary-tree structure to perform swaps over arbitrary lengths efficiently.
\end{itemize}

So, what is the circuit complexity corresponding to the Transformer? Given sequence length $n$, \citet{merrill2023parallelism} have shown that a constant-depth Transformer without chain-of-thought reasoning can only solve problems within $TC^0$. Meanwhile, several recent works~\cite{grazzi2024unlocking,siems2025deltaproduct,merrill2024illusion} demonstrate that the expressivity of DeltaNet exceeds $TC^0$.

By proving that DeltaFormer proposed in this section can perform tracking of $n$ elements~\footnote{Tracking $n$ states essentially refers to the task of swapping elements within a sequence of length $n$.}, for $n \geq 5$ (see Appendix~\ref{sec:state_tracking_delt_attn} for details), \emph{the expressivity of DeltaFormer surpasses $TC^0$ and theoretically reaches $NC^1$.}

\paragraph{\textbf{Algorithm Implementation.}}
We begin by analyzing the algorithm's complexity. Let vectors $\vect{q}, \vect{k}, \vect{v}, \vect{u} \in \mathbb{R}^d$~\footnote{For the inner product, $\vect{q}$ and $\vect{k}$ must have the same dimension. Eq.~\ref{eqn:delta_attn_u} implies that $\vect{v}$ and $\vect{u}$ also share the same dimension. Although the dimensions of $\vect{k}$ and $\vect{v}$ may differ, this variance is generally bounded by a constant factor. For simplicity, we assume their dimensions are equal herein.}, and $T$ be the sequence length.
\begin{itemize}
    \item \emph{Training phase.} First, we compute the sequence $\{\vect{u}\}_{i=1}^{T}$. Observing Eq.~\ref{eqn:delta_attn_u}, the computational complexity~\footnote{Computational complexity is defined here as the number of multiplication operations.} for each $\vect{u}$ is $\mathcal{O}(Td)$. Thus, the computational complexity for the entire sequence amounts to $\mathcal{O}(T^2 d)$. Since $\vect{u}_t$ depends on the history $\{\vect{u}\}_{i=1}^{t-1}$, the sequential time complexity is $\mathcal{O}(T)$. Once $\{\vect{u}\}_{i=1}^{T}$ is computed, the operations in Eq.~\ref{eqn:delta_attn_o} incur computational and time complexities of $\mathcal{O}(T^2 d)$ and $\mathcal{O}(1)$, respectively. Therefore, the overall computational and time complexities during training are $\mathcal{O}(T^2 d)$ and $\mathcal{O}(T)$, respectively.

    \item \emph{Inference phase.} First, we compute $\vect{u}_t$ and integrate it with $\vect{k}_t$ to update $\vect{S}_t$. The corresponding computational and time complexities are $\mathcal{O}(Td)$ and $\mathcal{O}(1)$, respectively. Then, $\vect{q}_t$ is employed to recall the final output $\vect{o}_t$ from $\vect{S}_t$, with computational and time complexities of $\mathcal{O}(d^2)$ and $\mathcal{O}(1)$, respectively. Consequently, the overall computational and time complexities during inference are $\mathcal{O}(Td + d^2)$ and $\mathcal{O}(1)$, respectively.
\end{itemize}
According to the above analysis, the training-stage time complexity of $\mathcal{O}(T)$ is significantly worse than the $\mathcal{O}(1)$ complexity of softmax attention. Thus, improving training parallelism is crucial. To address this, we can partition the sequence into chunks and aim for fully parallel computation within each chunk. Consequently, the overall training time complexity can be optimized to $\mathcal{O}(T/C)$, where $C$ denotes the chunk size. Detailed implementation specifics and pseudo-code can be found in Appendix~\ref{sec.appendix.chunkwise}. Additionally, we provide a reference code for the DeltaFormer state tracking toy experiment in Appendix~\ref{subsec:deltaformer_toy_model}.

\subsubsection{Vignette 3: What Will Happen in Infinitely Long In-Context Learning?}
\label{subsubsec:infinite_context}
As shown in Table~\ref{tab:mem_update_forms}, different associative memory models correspond to distinct memory update optimization objectives. Taking the softmax attention as an example, during test-time decoding—i.e., in-context learning (ICL)—the corresponding memory $\vect{S}$ undergoes single-step gradient descent updates. Consider an extreme scenario: \emph{as the length of ICL extends indefinitely, what would be the behavior of the entire optimization process?} Exploring this limiting case may provide insights into predicting the technological evolution of models designed for extremely long contexts.

\paragraph{\textbf{Single-head DeltaNet.}}
Specifically, we first consider a representative simple model: DeltaNet with single-head attention~\footnote{The optimization objective corresponding to vanilla linear attention is unbounded. As $t$ increases, the norm of $\vect{Sk}$ explodes, causing model collapse. This implies that linear attention (without optimizations such as memory decay) cannot perform very long in-context learning. The optimization objective for DeltaNet is bounded, avoiding the aforementioned issues, making it a more reasonable starting point for analysis and comparison.}. The optimization objective of DeltaNet is:
\begin{align}
    \mathcal{L}_t(\vect{S}_{t-1}) = \frac{1}{2}\|\vect{S}_{t-1}\vect{k}_t - \vect{v}_t\|^2.
\end{align}
Let us attempt to derive the optimal memory $\vect{S}^{*}$ corresponding to this optimization objective:
\begin{align}
    \vect{S}^{*} = \arg\min_{\vect{S}} \mathbb{E}_{\vect{x}}\left[\frac{1}{2}\|\vect{S}\vect{W}_k\vect{x}-\vect{W}_v\vect{x}\|^2\right],
\end{align}
where $\vect{x}$ denotes the input variable, $\vect{k} = \vect{W}_k\vect{x}$, $\vect{v} = \vect{W}_v\vect{x}$, and $\{\vect{W}_k, \vect{W}_v\}$ are fixed model weights at test time. According to Appendix~\ref{sec:associative_mem_analytical_solution}, the analytical solution for $\vect{S}^{*}$ is:
\begin{align}
    \vect{S}^{*} = \left(\vect{W}_v\,\mathbb{E}_{\vect{x}}[\vect{x}\vect{x}^{\top}]\,\vect{W}_k^{\top}\right)\left(\vect{W}_k\,\mathbb{E}_{\vect{x}}[\vect{x}\vect{x}^{\top}]\,\vect{W}_k^{\top}\right)^{-1}.
    \label{eqn:deltanet_analytical_solution}
\end{align}
Since $\vect{W}_k$ is typically invertible~\footnote{Square matrices are typically full rank, and therefore invertible.} and $\mathbb{E}_{\vect{x}}[\vect{x}\vect{x}^{\top}]$ is generally positive definite. Thus, the analytical solution simplifies to:
\begin{align}
    \vect{S}^{*} = \vect{W}_v\vect{W}_k^{-1}.
\end{align}
This indicates that, ideally, as memory updates progress, memory $\vect{S}$ converges to the fixed value $\vect{S}^{*}$. When the memory is no longer updated, it means that no new information is written into the memory, and the recall result depends solely on the fixed $\vect{S}^{*}$. If $\vect{S}^{*}$ is viewed as part of the model weights, the single-head DeltaNet degenerates into a bi-gram model, where the output is determined only by the current input. \emph{Theoretically, as the context length approaches infinity, the single-head DeltaNet loses its in-context learning capability.}

\paragraph{\textbf{Multi-head DeltaNet.}}
In the case of multi-head setting, $\vect{W}_k$ and $\vect{W}_v$ become low-rank matrices. Eq.~\ref{eqn:deltanet_analytical_solution} is thus simplified to:
\begin{align}
    \vect{S}^{*} = \vect{W}_v\,\mathbb{E}_{\vect{x}}[\vect{x}\vect{x}^{\top}]\,\vect{W}_k^{+}\vect{W}_k\,\mathbb{E}_{\vect{x}}[\vect{x}\vect{x}^{\top}]^{-1}\,\vect{W}_k^{+},
\end{align}
where $\vect{W}_k^{+}$ denotes the pseudo-inverse of $\vect{W}_k$. Since generally $\vect{W}_k^{+}\vect{W}_k \ne \vect{I}$, the term $\mathbb{E}_{\vect{x}}[\vect{x}\vect{x}^{\top}]$ cannot be eliminated. This implies that the optimal memory $\vect{S}^{*}$ depends on the data distribution and may not be fixed. \emph{The low-rank property of projection matrices (e.g., $\vect{W}_k$) introduced by multi-head attention may alleviate the degeneration of the model's in-context learning capability.}

\paragraph{\textbf{Softmax Attention.}}
Given that the multi-head structure can prevent ICL degradation, does this mean there are no theoretical obstacles remaining in constructing models supporting infinite context lengths? At least for softmax attention, this is not the case.

According to Table~\ref{tab:mem_update_forms}, the optimization objective of softmax attention is:
\begin{align}
    \mathcal{L}_t(\vect{S}_{t-1}) = \frac{1}{t} \left( \frac{1}{2} \|\vect{S}_{t-1}\|^2_F - \langle \vect{S}_{t-1}\phi(\vect{k}_{t}), \vect{v}_{t}\rangle \right).
    \label{eqn:long_opt_obj_softmax_attn}
\end{align}
When $t$ is sufficiently large, due to the presence of the $\frac{1}{t}$ term, we have $\mathcal{L}_t(\vect{S}_{t-1}) \to 0$, causing the gradient to approach zero and $\vect{S}$ to gradually cease updating. \emph{This implies that the ICL capability of softmax attention will still gradually diminish as the context length increases.} Intuitively, when the context, i.e., the KV cache, becomes excessively long, the information stored in the context resembles a ``vast ocean'' with a low signal-to-noise ratio, making it difficult to precisely recall the target information, thus rendering ICL infeasible.

If we remove the normalization, although the gradient does not vanish as $t$ increases, softmax attention without normalization corresponds to an unbounded optimization objective, which would eventually lead to model collapse.

\paragraph{\textbf{Summary.}}
Associative memory models theoretically risk converging gradually toward a fixed value as sequence length increases. Specifically for the attention layer, memory tends to progressively stop updating, leading to degraded ICL capability. Hence, \emph{constructing models capable of infinite contexts may be challenging}. But it should be acknowledged that practical scenarios are considerably more complex: mechanisms such as multi-head attention, memory decay, and short key-value convolutions can effectively prevent memory updates from stagnating. Moreover, large neural networks typically exhibit multi-layered structures, where fluctuations in single-layer optimization (e.g., variability introduced by $\mathbb{E}_{\vect{x}}[\vect{x}\vect{x}^{\top}]$) may accumulate and amplify across layers, thus obstructing memory from smoothly approaching the theoretical analytical solution.
\section{Experiments}

\subsection{Multi-head Efficiency Trade-off} 
\label{sec.capex}

\begin{figure}[t]
  \centering
  \includegraphics[width=0.5\textwidth]{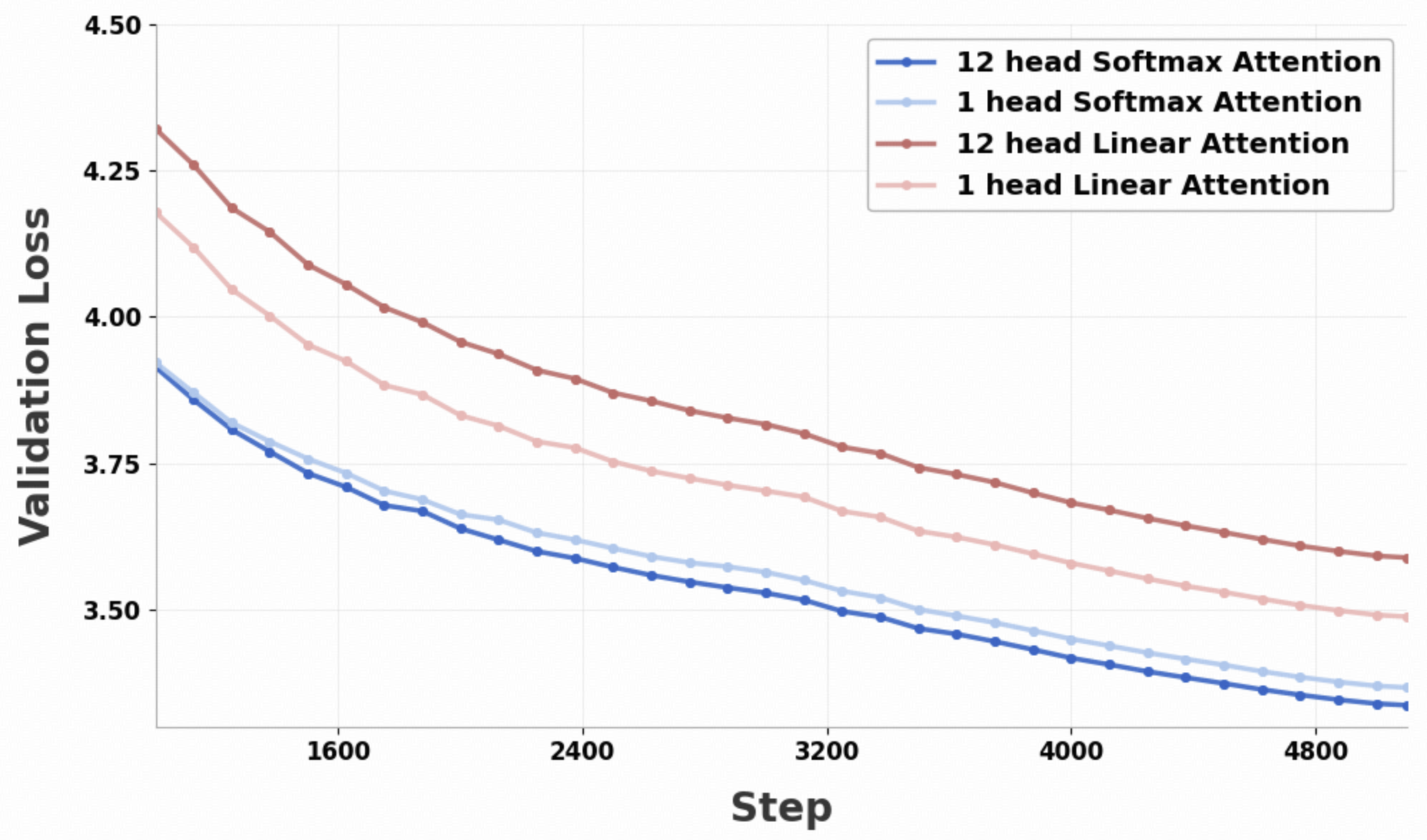} 
  \caption{{Effect of dimension under different attention mechanisms.} Validation loss with GPT-2 small (12L, 768D). Reducing head numbers benefits Linear Attention but harms Softmax Attention, likely due to differences in their retrieval and expressive properties.}
  \label{fig:head_tradeoff}
\end{figure}

We conduct experiments on a GPT-2 small Transformer with 12 layers and 768 hidden dimensions to study how attention mechanisms respond to head count versus per-head dimensionality. Our setup follows the open-source BlaGPT implementation~\footnote{\url{https://github.com/erogol/BlaGPT}}, including dataset and training configuration.

For the linear attention setting, we simply replace the Softmax operation with a linear kernel function and apply per-head RMS normalization to maintain training stability. To ensure fair comparison, we also remove the original normalization term from the Softmax baseline and apply per-head RMS norm instead.

As shown in Figure~\ref{fig:head_tradeoff}, {Linear Attention benefits significantly from reducing the number of heads to 1, thereby increasing head dimensionality}, while {Softmax Attention degrades under the same condition}. This supports our hypothesis that Softmax relies more on its expressiveness through multi-head, while Linear Attention gains more from greater memory capacity. These findings provide preliminary empirical validation for the hypothesis in Sec.~\ref{subsec:mem_cap}.

\subsection{Track the Exchange of Elements} 
\label{sec.trackex}

\begin{figure}[h]
  \centering
  \includegraphics[width=0.8\textwidth]{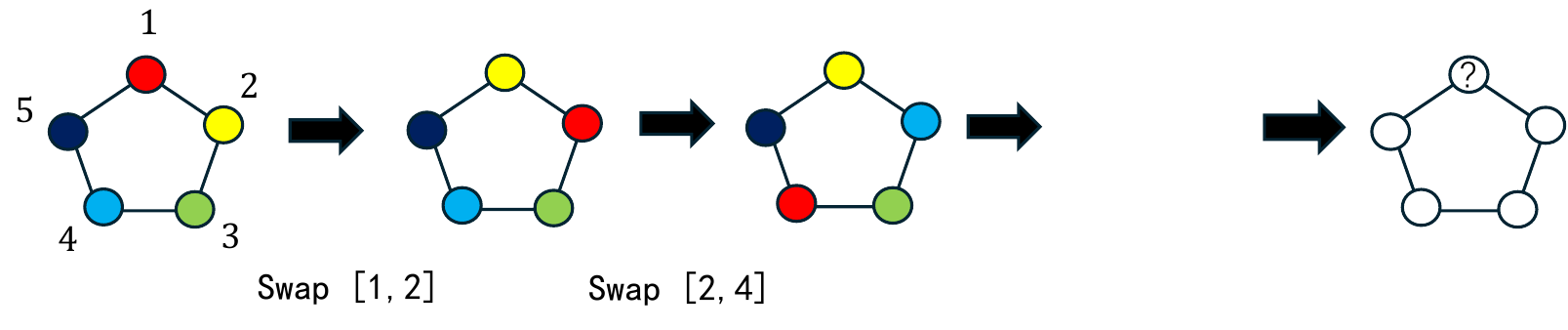} 
  \caption{Swap task diagram. At the beginning, tokens of different colors are placed at positions $1$ to $5$, and the tokens of two positions are exchanged at each step. We expect the model to query what the token for each position is at each step. Simply but without loss of generality, we default to outputting the token at the first position to avoid introducing a "query token". This task can also be tokenized into a task with an input vocabulary size of $C_5^2 = 10$ and an output vocabulary size of $5$.} 
  \label{fig:s5}
\end{figure}

Although our Theorem \ref{th:state_exchange} proves that DeltaFormer we describe in Section \ref{subsubsec:delta_attn} can track the exchange of $n$ objects, it still needs to be proven through experiments. By using gradient descent, can DeltaFormer learn to track the exchange of $n$ objects from the data? For this purpose, we conduct an experiment in this section to verify this. The setting is shown in Figure \ref{fig:s5}, and the default context length is $16$. \footnote{For simplicity, we omit the temperature coefficient used in the similarity function in this section.} 

\begin{figure}[ht]
    \centering
    \begin{subfigure}[b]{0.3\textwidth}
        \includegraphics[width=\textwidth]{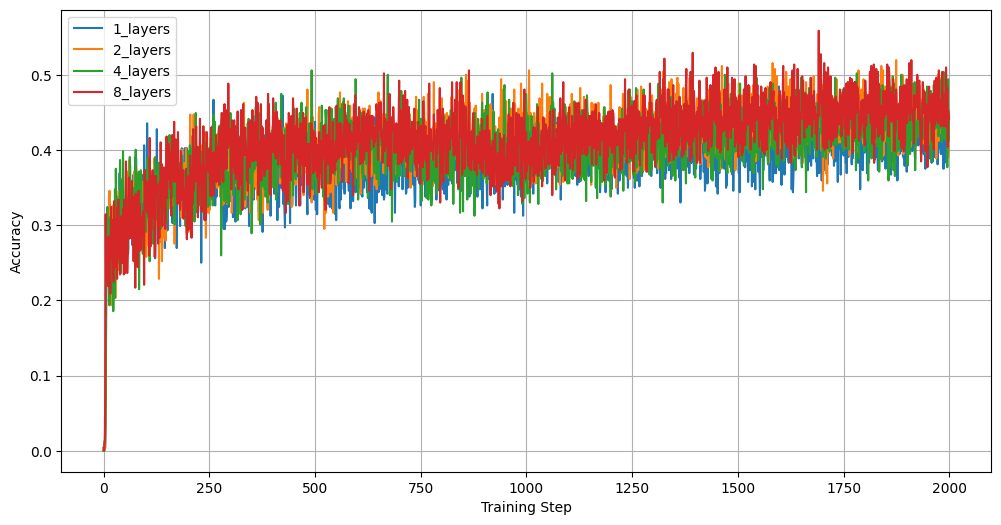}
        \caption{Transformer. \\ \ }
        \label{fig:sub1}
    \end{subfigure}
    ~ 
    \begin{subfigure}[b]{0.3\textwidth}
        \includegraphics[width=\textwidth]{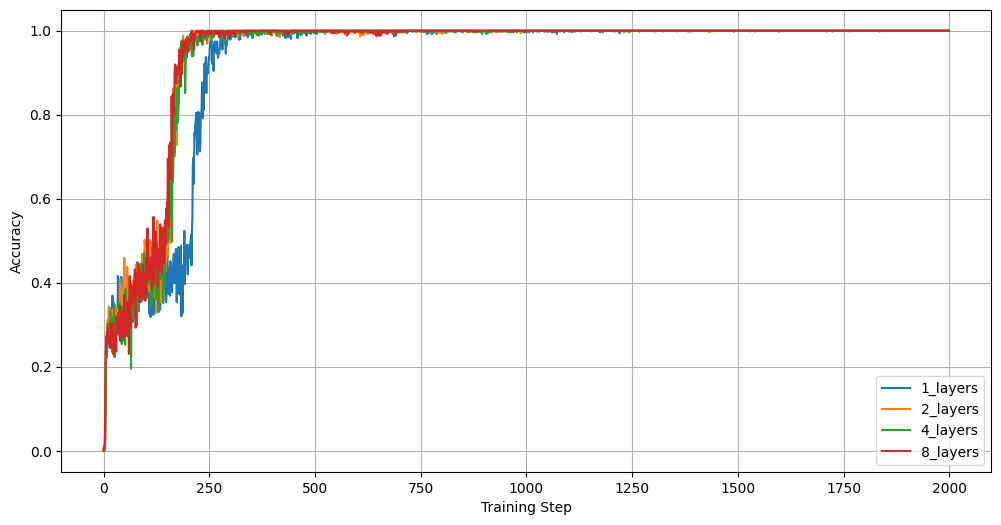}
        \caption{DeltaFormer. ($\kappa_1(\vect{x},\vect{y})=\lfloor \vect{x}^\top \vect{y} \rfloor$)}
        \label{fig:sub2}
    \end{subfigure}
    ~ 
    \begin{subfigure}[b]{0.3\textwidth}
        \includegraphics[width=\textwidth]{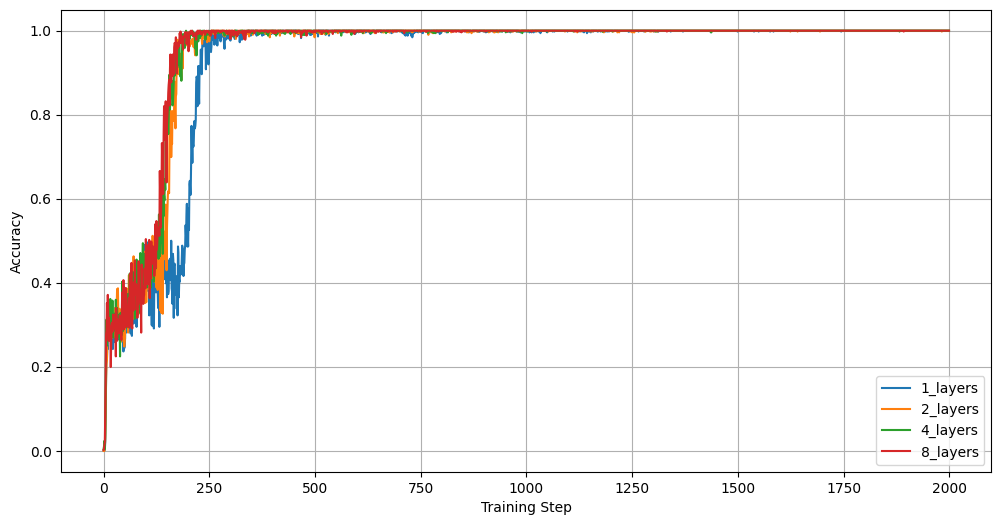}
        \caption{DeltaFormer. ($\kappa_1(\vect{x},\vect{y})= \vect{x}^\top \vect{y}$) \\ \ }
        \label{fig:sub3}
    \end{subfigure}
    \\
    \begin{subfigure}[b]{0.3\textwidth}
        \includegraphics[width=\textwidth]{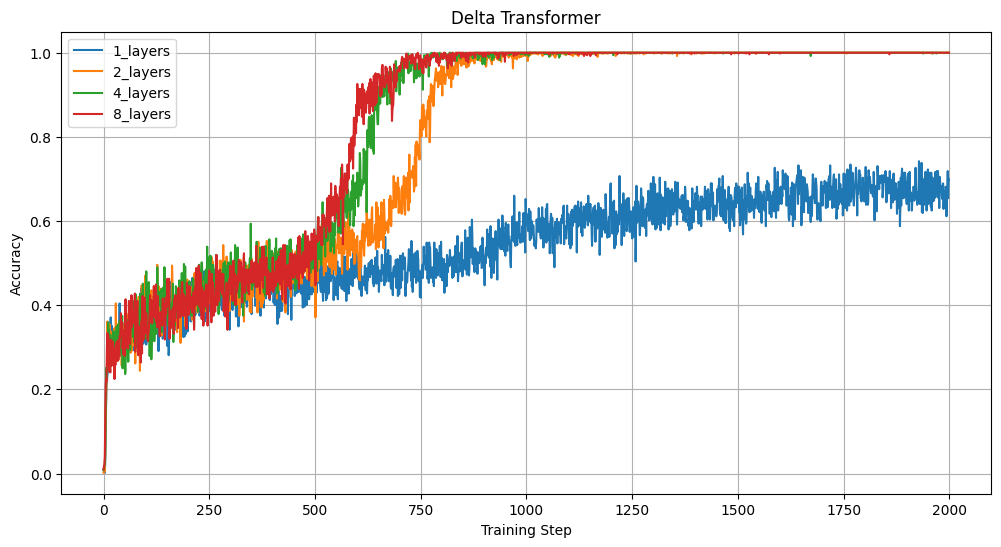}
        \caption{DeltaFormer. ($\kappa_1(\vect{x},\vect{y})= \max(\vect{x}^\top \vect{y},0)$)}
        \label{fig:sub4}
    \end{subfigure}
    ~ 
    \begin{subfigure}[b]{0.3\textwidth}
        \includegraphics[width=\textwidth]{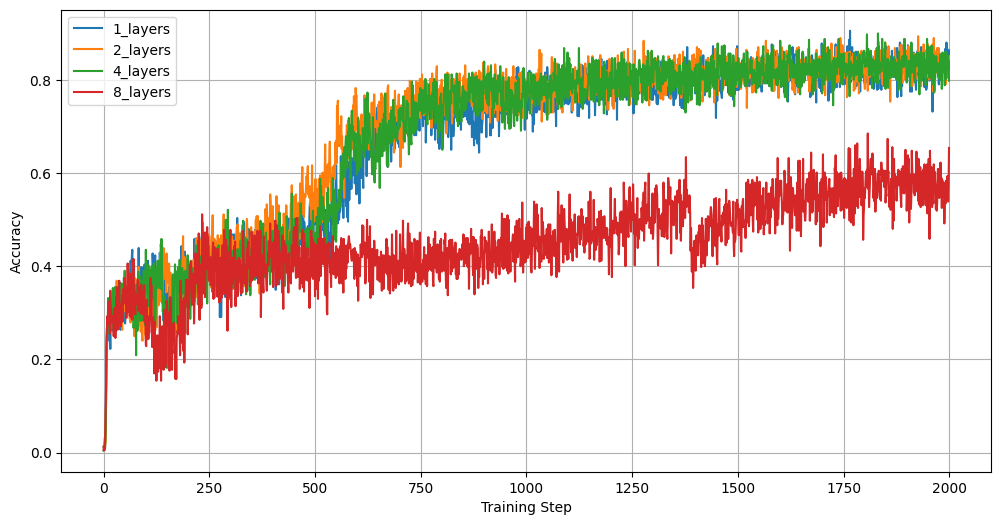}
        \caption{DeltaFormer. ($\kappa_1(\vect{x},\vect{y})= \exp(\vect{x}^\top \vect{y})$)}
        \label{fig:sub4}
    \end{subfigure}
    ~ 
    \begin{subfigure}[b]{0.3\textwidth}
        \includegraphics[width=\textwidth]{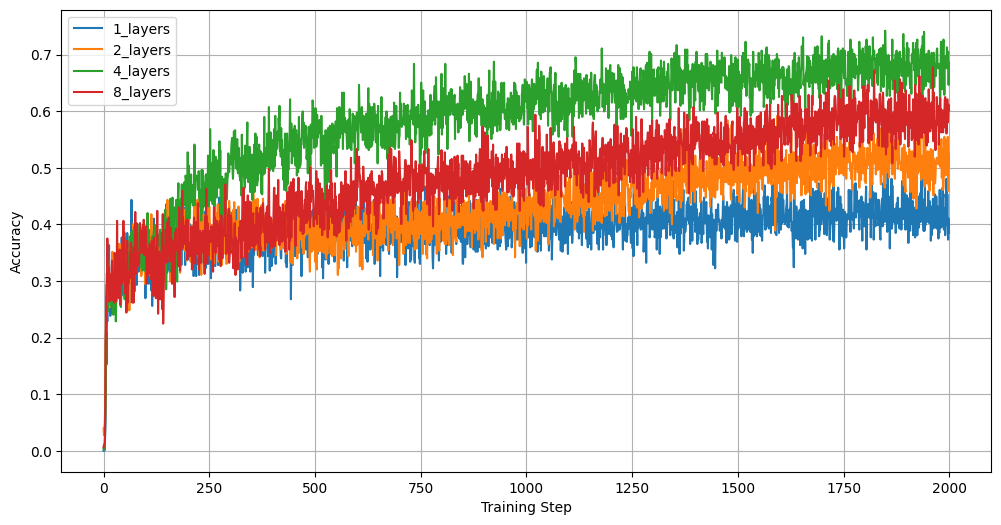}
        \caption{DeltaFormer. ($\kappa_1(\vect{x},\vect{y})= softmax(\vect{x}^\top \vect{y}) $)}
        \label{fig:sub4}
    \end{subfigure}
    \caption{Comparison of Transformer and DeltaFormer using different similarity functions $\kappa_1(\cdot)$ for performing swapping tasks. For $\kappa_2(\cdot)$, we use the softmax function to maintain consistency with Transformer. Pay attention to the scale of the y-axis. To ensure convergence, $\lfloor\cdot \rfloor$ means round to two decimal, such as $\lfloor 1.236 \rfloor = 1.24$.}
    \label{fig:full}
\end{figure}

\textbf{DeltaFormer Can Track the Exchange of Elements.} We compared DeltaFormer and Transformer under different similarity function designs, as shown in Figure \ref{fig:full}. Under almost imaginable simple $\kappa_1(\cdot)$ designs, DeltaFormer has achieved better results than Transformer models. And the 1-layer DeltaFormer can execute and track the exchange operations of $5$ elements. But increasing the number of layers in the transformer did not improve either. We speculate that there may be optimization issues involved, even with DeltaFormer. We will explore this hypothesis in the following sections.

\textbf{The Kernel $\kappa_1$ Used in $\vect{u}_t$ Is important.} Another noteworthy point is that the choice of different similarity functions also has a significant impact on the final effect. According to our construction proof in Theorem \ref{th:state_exchange}, the similarity function can effectively track $5$ elements. The closer the similarity function is to the constructive method, the better the performance. The normalization term of $softmax$ has a negative impact on the similarity calculation of $exp$. And the $\kappa_2$ used in our reading method is not based on the constructed similarity to ensure that it is the same as the standard attention, but instead uses softmax. Even so, a suitable $\kappa_1$ can achieve 100\% effectiveness. Because our Theorem \ref{th:state_exchange} actually proves that there is a way to read the elements at each position in a certain form of $\vect{u}$. This means that the exchange of elements is implicitly included in the update of $\vect{u}$. Intuitively speaking, if the similarity selection of $\kappa_1$ is not appropriate, it will cause more cumulative errors in the update of $\vect{u}$. Mathematically speaking, it actually reflects the perturbation of a inverse matrix is likely to be ill-conditioned. We rewrite the calculations for $\vect{u}$ and $\vect{o}$ as follows:
\begin{align}
    \vect{u} &= \vect{A}_1^{-1} \vect{v} \nonumber \\
    \vect{o} &= \vect{A}_2 \vect{u},
\end{align}
then we will have:
\begin{align}
    \|(\vect{A}_1+\Delta \vect{A})^{-1}\vect{V} - \vect{A}_1^{-1}\vect{V}\| \approx \|\vect{A}_1^{-1}(\Delta \vect{A})\vect{A}_1^{-1}\vect{V}\| \leq \|\vect{A}_1^{-1}\|\|\Delta \vect{A}\|\|\vect{A}_1^{-1}\|\|\vect{V}\| = \|\vect{A}_1^{-1}\|^2\|\Delta \vect{A}\|\|\vect{V}\|,
\end{align}
and
\begin{align}
    \|(\vect{A}_2+\Delta \vect{A}_2)\vect{U}-\vect{A}_2\vect{U}\| = \|(\Delta \vect{A}_2)\vect{U}\| \leq \|\Delta \vect{A}_2\|\|\vect{U}\|.
\end{align}
The stability of the calculation for $\vect{u}$ is weaker than that for $\vect{o}$, so the calculation for $\vect{u}$ need to balance stability and expressivity, and we provide a group query attention like approach to balance expressiveness and training stability as shown in the Appendix \ref{appendix:re-exam}.

\begin{figure}[ht]
    \centering
    \begin{subfigure}[b]{0.3\textwidth}
        \includegraphics[width=\textwidth]{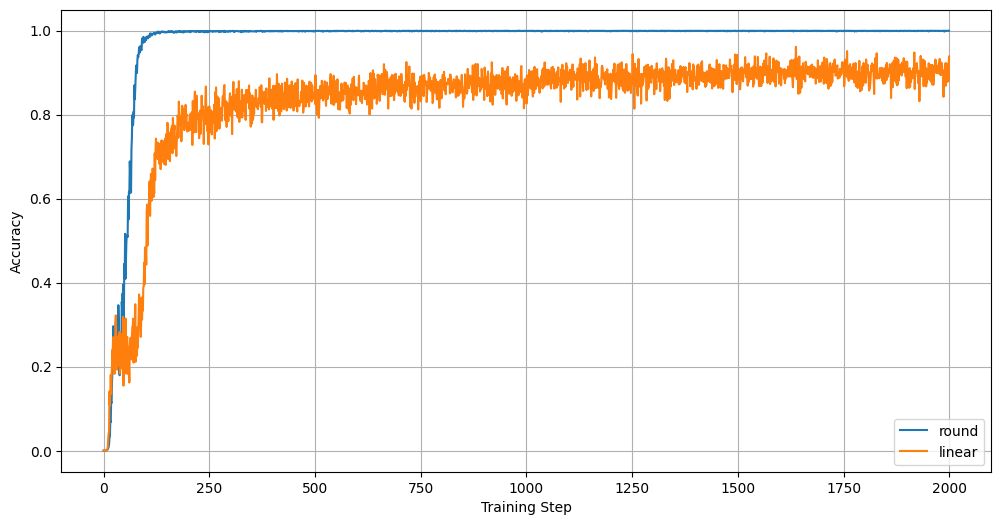}
        \caption{$d=128$, $n=128$.     }
        \label{fig:y}
    \end{subfigure}
    ~ 
    \begin{subfigure}[b]{0.3\textwidth}
        \includegraphics[width=\textwidth]{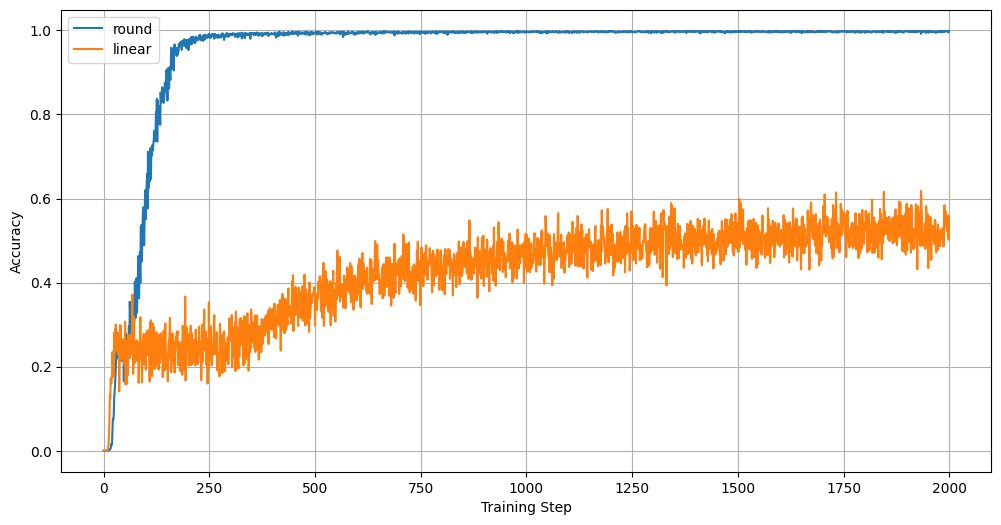}
        \caption{$d=128$, $n=256$.}
        \label{fig:x}
    \end{subfigure}
    ~
    \begin{subfigure}[b]{0.3\textwidth}
        \includegraphics[width=\textwidth]{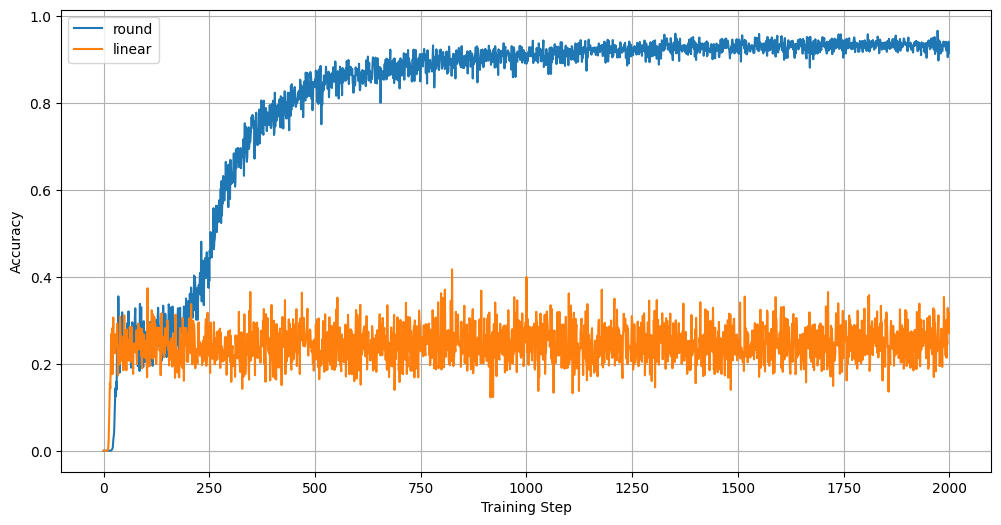}
        \caption{$d=128$, $n=512$.}
        \label{fig:x}
    \end{subfigure}
    \caption{Comparison of  DeltaFormer with $\kappa(\vect{x},\vect{y})=\vect{x}^\top \vect{y}$ and $\kappa(\vect{x},\vect{y})=\lfloor \vect{x}^\top \vect{y} \rfloor$. }
    \label{fig:press}
\end{figure}

\textbf{Stress Testing of Linear and Nonlinear Kernels.} We also conducted stress tests on the round and linear function in this Section, with $d=128$. The setting is similar to Section \ref{sec.trackex},  but with experiments where $n$ is greater than or equal to $128$. We use $n \in \{128,256,512\}$, and the corresponding training length is $\{256,512,1024\}$ to ensure as much as possible that most elements participate in the exchange. In addition, to avoid optimization issues, we adopted the almost orthogonal vectors used in our Theorem \ref{th:state_exchange} to set key and value of the model and the model only needs to learn to read information from the state space. The results is shown in Figure \ref{fig:press}.  We can observe that when $d$ is fixed, as $n\geq d$ increases, the performance of the linear kernel is severely degraded.  This essentially involves the famous Thompson problem, which is how to place as many orthogonal vectors as possible on the d-dimensional unit sphere. However linear functions cannot have superposition, and nonlinear functions can store a large amount of information through superposition \cite{elhage2022superposition}.

\textbf{Curriculum Learning Is Important.} As shown in the figure, we initially trained at a length of $256$, and the convergence speed of the model was very slow. So we decided to gradually lengthen the window from $32$, that is, gradually increase the difficulty. We find that on the basis of such curriculum learning, the model can achieve better performance with fewer computation and samples. A similar phenomenon has also been observed in using Transformers in-context~\cite{garg2022can}. 

\begin{figure}[ht]
    \centering
    \begin{subfigure}[b]{0.4\textwidth}
        \includegraphics[width=\textwidth]{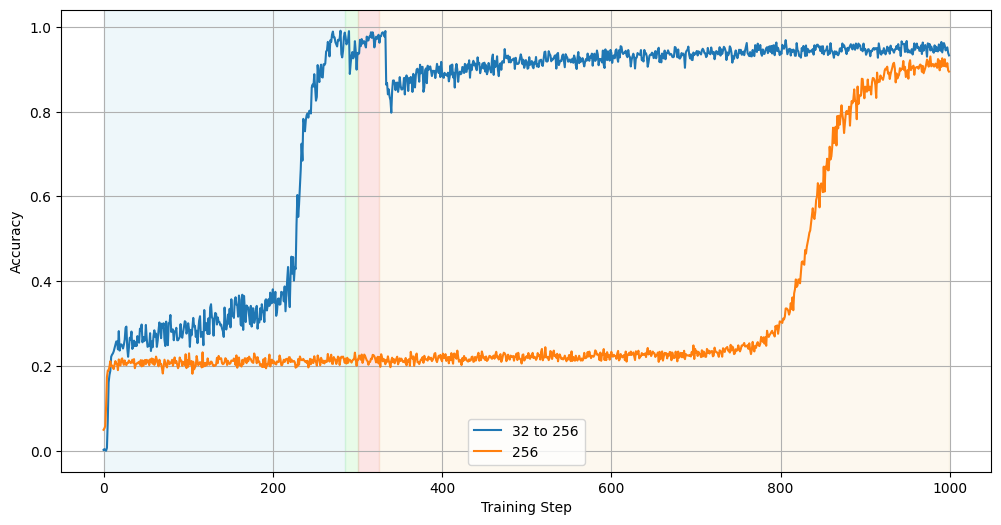}
        \caption{DeltaFormer with RoPE.    }
        \label{fig:curriculum_rope}
    \end{subfigure}
    ~ 
    \begin{subfigure}[b]{0.4\textwidth}
        \includegraphics[width=\textwidth]{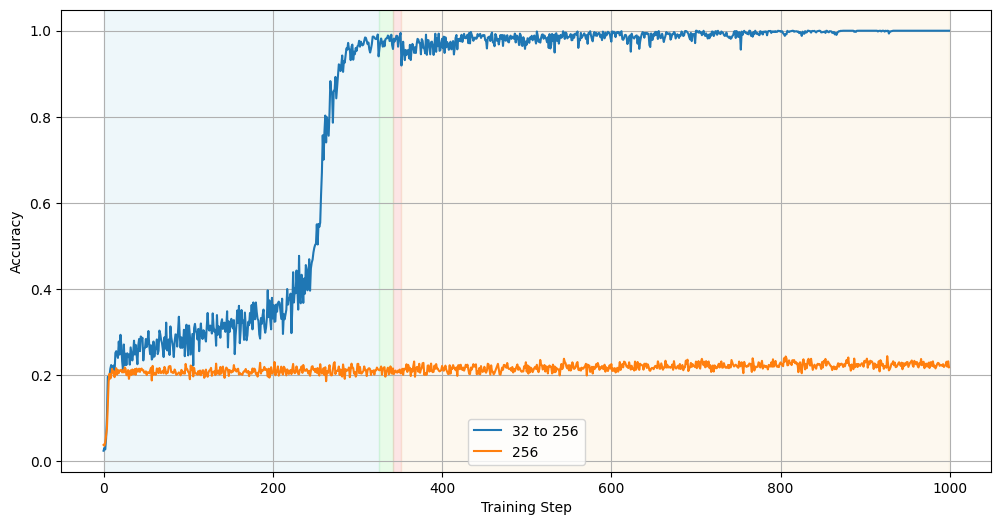}
        \caption{DeltaFormer with NoPE.}
        \label{fig:curriculum_nope}
    \end{subfigure}
    \caption{Comparison of  DeltaFormer using different learning strategy and position embedding. Each use $\kappa_1(\vect{x},\vect{y})=\lfloor \vect{x}^\top \vect{y} \rfloor$. ``32 to 256'' means that the initial training length  is 32, which means the number of swaps is 32. When the accuracy reaches 0.99, the training length will be doubled until it reaches 256. Each color gradient in the image represents a doubling of the training length. And ``256'' means that the model is trained on a training length of 256 from the beginning. The y-axis reflects the accuracy at the current training length.}
    \label{fig:nope}
\end{figure}

\textbf{The Role of Rotary Embeddings~\cite{su2024roformer}.}  Because our proof of Theorem \ref{th:state_exchange} does not require positional embeddings. Therefore, another interesting experiment is that we removed the default rotary position embeddings which is widely used in modern Transformer. We find that after removing the positional embedding, the convergence of the model slowed down, and even when trained directly at a length of $256$, the model get a random score. On the other hand, after removing RoPE, ``32 to 256'' can achieve 100\% accuracy. And during the extension process, the performance degradation of NoPE at jump points decreases, indicating that NoPE has better length generalization. There are also some studies~\cite{kazemnejad2023impact} that have made similar findings regarding the length generalization of position embedding. Therefore, we speculate that RoPE may have damaged the expression of  model and extrapolation ability, but it is more conducive to optimization.

\subsection{Reachability of Directed Acyclic Graphs}
Furthermore, we have a simple graph connectivity task, which involves determining reachability on a directed acyclic graph. For simplicity, we only consider whether other nodes can be reached by the first node in a certain topology ranking. And each node encodes at most its neighboring node information at the beginning.  Due to the final output being true and false, in order to avoid class imbalance, we construct the input data by dividing $n$ points into $2$ classes on average, with one tree for each class. So we only judge the reachability of other nodes starting from a certain root node. Then we only encode the parent node information for each node.  

\begin{figure}[ht]
    \centering
    \begin{subfigure}[b]{0.45\textwidth}
        \includegraphics[width=\textwidth]{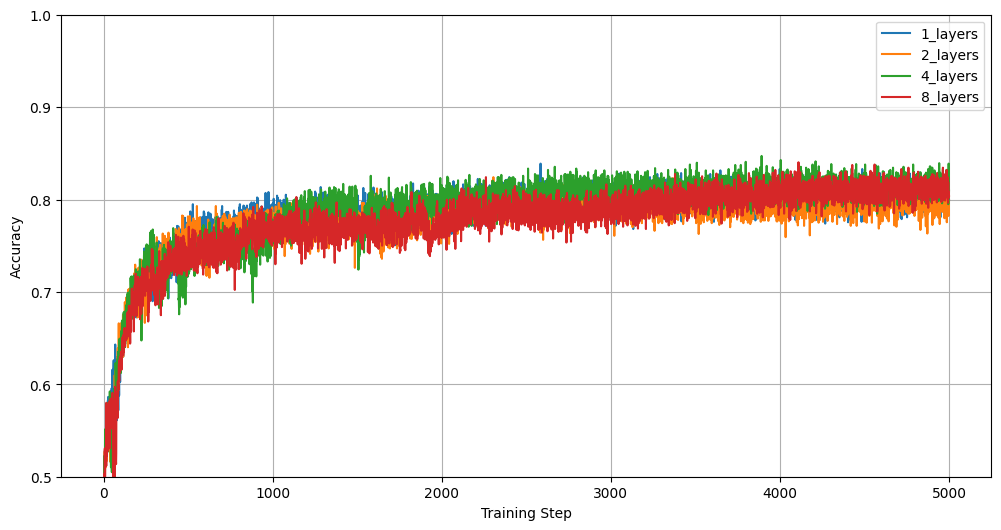}
        \caption{Transformer.}
        \label{fig:graphtr}
    \end{subfigure}
    \begin{subfigure}[b]{0.45\textwidth}
        \includegraphics[width=\textwidth]{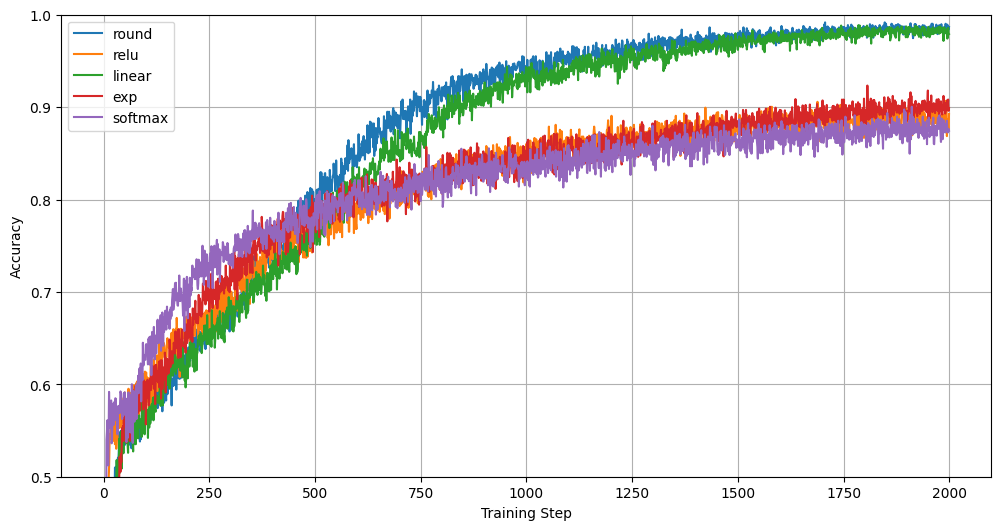}
        \caption{DeltaFormer.}
        \label{fig:sub2}
    \end{subfigure}
    \caption{Comparison of Transformer and DeltaFormer using different similarity functions $\kappa_1(\cdot)$ for performing the reachability of directed acyclic graphs tasks. For $\kappa_2(\cdot)$, we use the softmax function to maintain consistency with the standard Transformer.}
    \label{fig:graph}
\end{figure}

\textbf{DeltaFormer Performs Better Than Transformer.} We conducted experiments on $32$ nodes, as shown in Figure \ref{fig:graph}. A multi-layer transformer model is also difficult to achieve 100\% accuracy, but a single-layer DeltaFormer can do very well. In theory, a transformer requires \(\mathcal{O}(\log n)\)  layers to perform connectivity checks with $n$ nodes \cite{sanford2024understanding}. However, based on Figure \ref{fig:graphtr}, we speculate that the optimization problem also constrains the effectiveness of the Transformer in this task.  

\textbf{The Power of Matrix Inverse.} As shown in Eq. \ref{eq:pa}, the calculation of $\vect{u}$ can be rewrite by matrix inverse.  If  the adjacency matrix $\vect{A}$ is known, then to determine whether nodes $i$ and $j$ are connected, simply calculate $\vect{A},\vect{A}^2,\vect{A}^3,...,\vect{A}^n$ in sequence, and then observe whether the $(i, j)$ elements of this matrix are greater than 0. And $(\vect{I}-\vect{A})^{-1}$ can be seen as an approximation of $\vect{I}+\vect{A}+ \dots+\vect{A}^n$. Therefore, the operation of matrix inversion greatly improves the expressivity of the model. We can expect a model with matrix inverse to achieve better performance in graph related tasks. \citet{fagnou2024chain} also use matrix inversion for efficient entity tracking. Essentially, the operation of matrix inversion, which goes far beyond $TC^ 0$, enhances the expressiveness of Transformer.
\section{Conclusion}
This exploration, viewing Transformers through the prism of associative memory, represents but a modest step in our ongoing quest to unravel the full potential and intricate workings of these powerful architectures. While we believe this perspective offers valuable clarity and a unifying framework for understanding phenomena like memory capacity and update dynamics, it simultaneously opens the door to a landscape teeming with further questions and avenues for deeper investigation. We find ourselves particularly intrigued by, and eager to further explore, the following fundamental inquiries:

\textbf{The Architectural Endgame}: Is there an ultimate, optimal design for model architectures? Is associative memory, as we've discussed it, the most effective paradigm for achieving artificial intelligence? If this framework is indeed central, are our current Transformer structures—with their distinct attention and feed-forward layers—approaching a fundamental optimum, or are there radically different, yet more potent, instantiations of associative memory waiting to be discovered?

\textbf{Bridging Contextual and Persistent Memory}: Why do current Transformers necessitate a separation between dynamic Context Memory (Attention) and static Persistent Memory (FFNs)? Human cognition features sophisticated mechanisms, like the role of the hippocampus, for filtering, consolidating, and transferring salient contextual experiences into long-term, persistent knowledge. When might our models acquire analogous capabilities? Such an advancement could allow them to learn continuously and adaptively during test time, potentially mitigating the challenges posed by infinitely long contexts. A critical sub-question here is: if such a consolidation mechanism were to be developed, what would be the intrinsic reward signal or loss function to guide its learning and refinement?

\textbf{Parallelism, Hardware, and the Next Leap in Intelligence}: The inherent parallel complexity of existing Transformer models (largely within  $TC^0$) imposes limits on the types of problems they can efficiently learn and solve. This limitation is intrinsically linked to the capabilities of our current parallel hardware. Is there significant merit in pursuing models with higher parallel complexity classes (e.g.,  $TC^1$ or even P-complete tasks)? Does the advent of next-generation intelligence hinge upon a paradigm shift in hardware—perhaps the maturation of technologies like quantum computing, which theoretically promise efficient solutions to problems currently intractable within $TC^k$?

These questions underscore the vast and fertile ground that still lies before us. The journey to understand and replicate intelligence is far from over. Each answer we find tends to illuminate a dozen new questions, each more profound than the last. It is our hope that by posing these challenges and sharing our current understanding, we can stimulate further discussion, collaboration, and innovation within the research community, collectively pushing the boundaries of what artificial intelligence can achieve. The path ahead is complex, but the pursuit of these fundamental questions promises a future rich with discovery.

\section{Acknowledgements}
{
We would like to thank Songlin Yang, Bencheng Liao, Lai Wei, and Xiangyu Yu for their valuable feedback and insightful comments throughout the development of this work. We are also grateful to Shilong Liu for providing experimental support related to Deltanet, and to Guhao Feng for helpful discussions on circuit complexity theory.
We sincerely thank Tianle Cai for his constructive suggestions on many of the formula derivations and high-level conceptual thinking.
We further acknowledge the support of Jiaao He, Jianqiao Lu, and Heyuan Yao in various aspects of implementation, including algorithmic design and code-level experimentation.
Some of the names mentioned above and in the author list are internal aliases used within the company.
}

\clearpage

\bibliographystyle{plainnat}
\bibliography{main}

\clearpage

\beginappendix

\section{Derivation of the Inverse SNR Formula}
Since the linear form of Associative Memory is a special case of Eq.~\ref{eqn:kernel_recall}, we first derive Eq.~\ref{eqn:snr_kernel}, and then specialize the derivations for the linear and exponential kernels.

\paragraph{Inverse SNR with kernel.} 
\label{sec:inverse_snr}
First, we can write the associative recall equation with the kernel function $\kappa$ as follows:
\begin{align}
    \vect{S}_t \phi(\vect{k}_i) = \vect{v}_i \kappa(\vect{k}_i, \vect{k}_i) + \sum_{j \ne i} \vect{v}_j \kappa(\vect{k}_j,  \vect{k}_i).
    \label{eqn:associative_recall_kernel}
\end{align}
According to Eq.~\ref{eqn:def_SNR}, let
\[
    \vect{r} = \sum_{j \ne i} \vect{v}_j \kappa(\vect{k}_j, \vect{k}_i), \quad c = \kappa(\vect{k}_i, \vect{k}_i),
\]
then we have
\begin{align}
    \mathrm{SNR}^{-1} 
    &= \mathbb{E}_{\vect{v}_j, \vect{k}_j}\left[\frac{\|\vect{r}\|^2}{c^2 \|\vect{v}_i\|^2}\right] \notag \\
    &= \frac{
    \mathbb{E}_{\vect{v}_j, \vect{k}_j}\left[\left\| \sum_{j \ne i} \vect{v}_j \kappa(\vect{k}_j, \vect{k}_i) \right\|^2\right]
    }{
    c^2 \|\vect{v}_i\|^2
    } .
    \label{eqn:snr_kernel_start}
\end{align}
Continuing from Eq.~\ref{eqn:snr_kernel_start}, we expand the squared term inside the expectation:
\begin{align}
    \mathrm{SNR}^{-1} 
    &= \mathbb{E}_{\vect{v}_j, \vect{k}_j} \left[ \frac{\left\| \sum_{j \ne i} \vect{v}_j \kappa(\vect{k}_j, \vect{k}_i) \right\|^2}{c^2 \|\vect{v}_i\|^2} \right] \notag \\
    &= \frac{1}{c^2 \|\vect{v}_i\|^2} \, \mathbb{E}_{\vect{v}_j, \vect{k}_j} \left[ \left( \sum_{j \ne i} \vect{v}_j \kappa(\vect{k}_j, \vect{k}_i) \right)^\top \left( \sum_{j' \ne i} \vect{v}_{j'} \kappa(\vect{k}_{j'}, \vect{k}_i) \right) \right].
\end{align}

Since the vectors $\vect{v}_j$ are assumed to be i.i.d. standard Gaussian, the cross terms $j \ne j'$ vanish in expectation, and only the diagonal terms survive. Thus, we have
\begin{align}
    &= \frac{1}{c^2 \|\vect{v}_i\|^2} \, \mathbb{E}_{\vect{k}_j} \left[ \sum_{j \ne i} \mathbb{E}_{\vect{v}_j} \left[ \|\vect{v}_j\|^2 \right] \kappa^2(\vect{k}_j, \vect{k}_i) \right] \notag \\
    &= \frac{1}{c^2 \|\vect{v}_i\|^2} \, \mathbb{E}_{\vect{k}_j} \left[ (N-1) \, \mathbb{E}_{\vect{v}_j} \left[\|\vect{v}_j\|^2\right] \, \kappa^2(\vect{k}_j, \vect{k}_i) \right],
\end{align}
where $N$ denotes the total number of stored keys.

Assuming that $\mathbb{E}[\|\vect{v}_j\|^2] = \|\vect{v}_i\|^2$, which holds when the stored values have approximately the same magnitude, we further simplify:
\begin{align}
    &= \frac{(N-1)}{c^2} \, \mathbb{E}_{\vect{k}_j} \left[ \kappa^2(\vect{k}_j, \vect{k}_i) \right].
\end{align}

Finally, for large $N$, we use $N$ instead of $(N-1)$ for abbreviation, leading to
\begin{align}
    \mathrm{SNR}^{-1} 
    &= N \, \frac{ \mathbb{E}_{\vect{k}_j} \left[ \kappa^2(\vect{k}_j,\vect{k}_i) \right] }{ \kappa^2(\vect{k}_i, \vect{k}_i) }.
\end{align}

\paragraph{Linear kernel.}
\label{sec:linear_kernel}
Fix a probe key $\vect{k}_i\in\mathbb{R}^{d_k}$ and draw each other key
$\vect{k}_j\sim\mathcal{N}(\mathbf{0},\mathbf{I}_{d_k})$.
With the linear kernel $\kappa(\vect{x},\vect{y})=\vect{x}^{\top}\vect{y}$,
the random inner product $X_j:=\vect{k}_j^{\top}\vect{k}_i$ is obtained by
projecting $\vect{k}_j$ onto the direction of $\vect{k}_i$, hence
\[
   X_j \;\big|\;\vect{k}_i \;\sim\; \mathcal{N}\!\bigl(0,\,
   \lVert\vect{k}_i\rVert_2^{2}\bigr), 
   \qquad
   \text{so}\quad
   \mathbb{E}[X_j^{2}] = \lVert\vect{k}_i\rVert_2^{2}.
\]
Substituting this variance into the definition gives
\[
   \mathrm{SNR}^{-1}_{\text{Linear}}
   =N\,\frac{\mathbb{E}[X_j^{2}]}
{(\vect{k}_i^{\top}\vect{k}_i)^{2}}
   =\frac{N}{\lVert\vect{k}_i\rVert_2^{2}}\;.
\]
Because $\lVert\vect{k}_i\rVert_2^{2}\sim\chi^{2}_{d_k}$ has mean
$d_k$ and concentrates sharply around this value when $d_k$ is large, we
replace it by its expectation to obtain the high-dimensional estimate
\[
   \mathrm{SNR}^{-1}_{\text{Linear}}\;=\;\dfrac{N}{d_k+\mathcal{O}(\sqrt{d_k})}.
\]

\paragraph{Exponential kernel.}
\label{sec:exp_kernel}
For $\kappa_\text{exp}(\vect{x},\vect{y}) = \exp(\frac{\vect{x}^\top \vect{y}}{\tau})$ the inverse SNR is  
\[
   \mathrm{SNR}^{-1}_{\exp}
   =N\,
     \frac{\mathbb{E}_{\vect{k}_j}[\exp(\frac{2}{\tau}\vect{k}_j^{\top}\vect{k}_i)]}
          {\exp(\tfrac{2}{\tau}\lVert\vect{k}_i\rVert_2^{2})}.
\]
Conditional on the probe key $\vect{k}_i$, the inner product  
$X_j=\vect{k}_j^{\top}\vect{k}_i\sim\mathcal{N}\bigl(0,\lVert\vect{k}_i\rVert_2^{2}\bigr)$.  
The Gaussian moment–generating function therefore gives  
$\mathbb{E}[\exp(\frac{2X_j}{\tau})]=\exp(\frac{2\lVert\vect{k}_i\rVert_2^{2}}{\tau^{2}})$, and thus  
\[
   \mathrm{SNR}^{-1}_{\exp}
   =N\,
     \exp\Bigl(\tfrac{2}{\tau^{2}}-\tfrac{2}{\tau}\Bigr)
       \lVert\vect{k}_i\rVert_2^{2}
   =N\,
     \exp\Bigl(\tfrac{2(1-\tau)}{\tau^{2}}\lVert\vect{k}_i\rVert_2^{2}\Bigr).
\]
Replacing the squared norm of $\vect{k}_i$ by $d_k$ yields
\[
   \mathrm{SNR}^{-1}_{\text{exp}}\;=\;
     \frac{N}{\exp\bigl(\tfrac{2(\tau-1)}{\tau^{2}}(d_k+\mathcal{O}(\sqrt{d_k}))\bigr)}.
\]

\paragraph{ReLU kernel.}
\label{sec:relu_kernel}
Set $\kappa_{\text{ReLU}}(\vect{x},\vect{y})
      =\mathrm{ReLU}\bigl(\vect{x}^{\top}\vect{y}\bigr)
      =\max\bigl(0,\vect{x}^{\top}\vect{y}\bigr)$
and the inner product  
\(X_j=\vect{k}_j^{\top}\vect{k}_i\) is Gaussian,
\(X_j\mid\vect{k}_i\sim\mathcal{N}\!\bigl(0,\lVert\vect{k}_i\rVert_2^{2}\bigr)\).
Because the Gaussian density is symmetric,
\[
   \mathbb{E}\bigl[\mathrm{ReLU}(X_j)^{2}\,\big|\,\vect{k}_i\bigr]
   =\mathbb{E}\bigl[X_j^{2}\mathbf{1}_{\{X_j>0\}}\bigr]
   =\tfrac{1}{2}\,\mathbb{E}[X_j^{2}]
   =\frac{\lVert\vect{k}_i\rVert_2^{2}}{2}.
\]
Meanwhile
\(\kappa_{\text{ReLU}}(\vect{k}_i,\vect{k}_i)
      =\mathrm{ReLU}\bigl(\lVert\vect{k}_i\rVert_2^{2}\bigr)
      =\lVert\vect{k}_i\rVert_2^{2}\),
so the inverse SNR is
\[
   \mathrm{SNR}^{-1}_{\text{ReLU}}
   =N\,\frac{\lVert\vect{k}_i\rVert_2^{2}/2}{\lVert\vect{k}_i\rVert_2^{4}}
   =\frac{N}{2\,\lVert\vect{k}_i\rVert_2^{2}}.
\]
Replacing the squared norm of $\vect{k}_i$ by $d_k$ yields
\[
   {\mathrm{SNR}^{-1}_{\text{ReLU}}\;=\;
          \dfrac{N}{2(d_k+\mathcal{O}(\sqrt{d_k}))}.}
\]

\paragraph{SoLU kernel.}
\label{sec:solu_kernel}
For the Softmax–Linear-Unit (SoLU) non-linearity we define the kernel  
\[
  \kappa_{\text{SoLU}}(\vect{x},\vect{y})
  \;=\;
  (\vect{x}^{\top}\vect{y})
  \,\exp\bigl(\tfrac{\vect{x}^{\top}\vect{y}}{\tau}\bigr),
  \qquad \tau\in\mathbb{R}_{>0}.
\]
Let \(X_j=\vect{k}_j^{\top}\vect{k}_i\) as before.  
Conditioned on the query key \(\vect{k}_i\), the random variable
\(X_j\mid\vect{k}_i\sim\mathcal{N}(0,s^{2})\) with
\(s^{2}\stackrel{\mathrm{def}}{=}\lVert\vect{k}_i\rVert_2^{2}\).

Because \(X_j\) is Gaussian we may use the moment–generating function
\(\mathbb{E}[e^{aX_j}]=\exp(a^{2}s^{2}/2)\) and its derivatives:
\[
  \mathbb{E}\bigl[X_j^{2}\,e^{aX_j}\bigr]
  \;=\;
  \bigl(s^{2}+a^{2}s^{4}\bigr)\;
  \exp\bigl(\tfrac{a^{2}s^{2}}{2}\bigr).
\]
With \(a=2/\tau\) (because \(\kappa^{2}\) contains
\(\exp(2X_j/\tau)\)) we obtain
\[
  \mathbb{E}\!\left[\kappa_{\text{SoLU}}(\vect{k}_j,\vect{k}_i)^{2}\,\Bigm|\,\vect{k}_i\right]
  \;=\;
  \mathbb{E}\!\left[X_j^{2}\,e^{2X_j/\tau}\right]
  \;=\;
  s^{2}\!\Bigl(1+\frac{4s^{2}}{\tau^{2}}\Bigr)\,
  \exp\Bigl(\frac{2s^{2}}{\tau^{2}}\Bigr),
\]
\[
  \kappa_{\text{SoLU}}(\vect{k}_i,\vect{k}_i)
  \;=\;
  s^{2}\,\exp\bigl(\tfrac{s^{2}}{\tau}\bigr)
  \;\;\Longrightarrow\;\;
  \kappa_{\text{SoLU}}(\vect{k}_i,\vect{k}_i)^{2}
  \;=\;
  s^{4}\,\exp\bigl(\tfrac{2s^{2}}{\tau}\bigr).
\]

Plugging the two ingredients into
\(\mathrm{SNR}^{-1}=N\,\mathbb{E}\!\bigl[\kappa^{2}(\vect{k}_j,\vect{k}_i)\bigr]/\kappa^{2}(\vect{k}_i,\vect{k}_i)\)
gives
\[
  \mathrm{SNR}^{-1}_{\text{SoLU}}
  \;=\;
  N\,
  \frac{s^{2}\!\bigl(1+\tfrac{4s^{2}}{\tau^{2}}\bigr)\,
        \exp\bigl(\tfrac{2s^{2}}{\tau^{2}}\bigr)}
       {s^{4}\,\exp\bigl(\tfrac{2s^{2}}{\tau}\bigr)}
  \;=\;
  N\,
  \frac{1+\tfrac{4s^{2}}{\tau^{2}}}{s^{2}}\;
  \exp\Bigl[\tfrac{2s^{2}}{\tau^{2}}-\tfrac{2s^{2}}{\tau}\Bigr].
\]

For i.i.d.\ standard–normal keys the squared norm is
\(\lVert\vect{k}_i\rVert_2^{2}\sim\chi^{2}_{d_k}\) whose mean is \(d_k\).
Replacing \(s^{2}\) by \(d_k\) yields the large-width estimate
\[
  {\;
  \mathrm{SNR}^{-1}_{\text{SoLU}}
  \;\approx\;
  N\,
  \frac{1+\tfrac{4d_k}{\tau^{2}}}{d_k}\;
  \exp\Bigl[-\,\frac{2d_k}{\tau^{2}}\bigl(\tau-1\bigr)\Bigr]\; }.
\]
Setting the temperature \(\tau=\sqrt{d_k}\) gives:
\[
  \mathrm{SNR}^{-1}_{\text{SoLU}}
  \approx
  N\,
  \frac{1+\tfrac{4d_k}{d_k}}{d_k}\,
  \exp\bigl[-2(\sqrt{d_k}-1)\bigr]
  \;\approx\;
  \frac{5N}{d_k\exp(2\sqrt{d_k})}\;
  .
\]

\section{Optimization Objective of Associative Memory}
\label{sec:associative_mem_optimization_objective}

\subsection{General Form}
\label{subsec:general_ass_mem_opt_obj}
Consider a general recurrent form of memory update:
\begin{align}
    \vect{S}_{t} = \vect{A}_t \vect{S}_{t-1} \vect{B}_t + \vect{C}_t,
\end{align}
where $\vect{A}_t$, $\vect{B}_t$, and $\vect{C}_t$ are parameter matrices.

We aim to construct a scalar optimization objective $\mathcal{L}_{t}(\vect{S}_{t-1})$ such that:
\begin{align}
    \vect{S}_{t} = \vect{S}_{t-1} - \frac{\partial \mathcal{L}_{t}(\vect{S}_{t-1})}{\partial \vect{S}_{t-1}}.
\end{align}
In other words, we require:
\begin{align}
    \frac{\partial \mathcal{L}_{t}(\vect{S}_{t-1})}{\partial \vect{S}_{t-1}} = \vect{S}_{t-1} - \vect{A}_t \vect{S}_{t-1} \vect{B}_t - \vect{C}_t.
\end{align}

The linear portion of this gradient field can be expressed as:
\begin{align}
    \vect{S}_{t-1} \mapsto \vect{S}_{t-1} - \vect{A}_t \vect{S}_{t-1} \vect{B}_t.
    \label{eqn:linear_gradient_field}
\end{align}

To ensure that a vector field is a valid gradient field in matrix space (equipped with the Frobenius inner product $\langle \vect{X}, \vect{Y} \rangle = \mathrm{tr}(\vect{X}^\top \vect{Y})$), it must be self-adjoint. Specifically, for arbitrary matrices $\vect{X}, \vect{Y}$, we need:
\begin{align}
    \langle \vect{X}, \vect{A}_t \vect{Y} \vect{B}_t \rangle = \langle \vect{A}_t \vect{X} \vect{B}_t, \vect{Y} \rangle.
\end{align}

This condition requires $\vect{A}_t$ and $\vect{B}_t$ to be symmetric matrices. If they are not symmetric, generally there does not exist a scalar potential function whose gradient yields the desired vector field (Eq.~\ref{eqn:linear_gradient_field}).

Assuming that $\vect{A}_t$ and $\vect{B}_t$ are symmetric, we can explicitly construct $\mathcal{L}_{t}(\vect{S}_{t-1})$ as follows:
\begin{align}
    \mathcal{L}_{t}(\vect{S}_{t-1}) = \frac{1}{2}\operatorname{tr}(\vect{S}_{t-1}^\top \vect{S}_{t-1}) - \frac{1}{2}\operatorname{tr}(\vect{S}_{t-1}^\top \vect{A}_t \vect{S}_{t-1}\vect{B}_t) - \operatorname{tr}(\vect{C}_t^\top \vect{S}_{t-1}),
    \label{eqn:general_opt_obj}
\end{align}
where the gradient of each term is:
\begin{itemize}
    \item gradient of the first term:
    \[
        \frac{\partial}{\partial \vect{S}_{t-1}}\frac{1}{2}\mathrm{tr}(\vect{S}_{t-1}^\top \vect{S}_{t-1}) = \vect{S}_{t-1};
    \]
    \item gradient of the second term (assuming symmetric $\vect{A}_t$ and $\vect{B}_t$):
    \[
        \frac{\partial}{\partial \vect{S}_{t-1}}\frac{1}{2}\mathrm{tr}(\vect{S}_{t-1}^\top \vect{A}_t \vect{S}_{t-1} \vect{B}_t) = \vect{A}_t \vect{S}_{t-1} \vect{B}_t;
    \]
    \item gradient of the third term:
    \[
        \frac{\partial}{\partial \vect{S}_{t-1}}\mathrm{tr}(\vect{C}_t^\top \vect{S}_{t-1}) = \vect{C}_t.
    \]
\end{itemize}

\subsection{Representative Cases}
\label{subsec:ass_mem_opt_obj_cases}
\paragraph{Linear Attention}
By comparing with the memory update formula of linear attention:
\begin{align}
    \vect{S}_t = \vect{S}_{t-1} + \vect{v}_t \vect{k}_t^\top,
\end{align}
we obtain: $\vect{A}_t = \vect{I}$, $\vect{B}_t = \vect{I}$, and $\vect{C}_t = \vect{v}_t \vect{k}_t^\top$. Substituting these into Eq.~\ref{eqn:general_opt_obj}, we have:
\begin{align}
    \mathcal{L}_{t}(\vect{S}_{t-1}) = \frac{1}{2}\operatorname{tr}(\vect{S}_{t-1}^\top \vect{S}_{t-1}) - \frac{1}{2}\operatorname{tr}(\vect{S}_{t-1}^\top \vect{S}_{t-1}) - \operatorname{tr}((\vect{v}_t \vect{k}_t^\top)^\top \vect{S}_{t-1}).
\end{align}
Since,
\begin{align}
    \operatorname{tr}((\vect{v}_t \vect{k}_t^\top)^\top \vect{S}_{t-1}) &= \operatorname{tr}(\vect{k}_t \vect{v}_t^\top \vect{S}_{t-1}) \nonumber \\
    &= \operatorname{tr}(\vect{S}_{t-1}\vect{k}_t\vect{v}_t^\top) \nonumber \\
    &= \vect{v}_t^\top\vect{S}_{t-1}\vect{k}_t \nonumber \\
    &= \langle \vect{S}_{t-1}\vect{k}_t, \vect{v}_t \rangle,
\end{align}
we thus have:
\begin{align}
    \mathcal{L}_{t}(\vect{S}_{t-1}) = -\langle \vect{S}_{t-1}\vect{k}_t, \vect{v}_t \rangle.
\end{align}

If we introduce the gating matrix $\text{diag}(\vect{\lambda}_t)$, i.e., $\vect{A}_t = \text{diag}(\vect{\lambda}_t)$, then:
\begin{align}
    \mathcal{L}_{t}(\vect{S}_{t-1}) &= \frac{1}{2}\operatorname{tr}(\vect{S}_{t-1}^\top \vect{S}_{t-1}) - \frac{1}{2}\operatorname{tr}(\vect{S}_{t-1}^\top \text{diag}(\vect{\lambda}_t)\vect{S}_{t-1}) - \operatorname{tr}((\vect{v}_t \vect{k}_t^\top)^\top \vect{S}_{t-1}) \nonumber \\
    &= \frac{1}{2} \text{diag}(1 - \vect{\lambda}_t) \|\vect{S}_{t-1} \|_{F}^2 - \langle \vect{S}_{t-1}\vect{k}_t, \vect{v}_t \rangle.
\end{align}

\paragraph{DeltaNet}
By comparing with the memory update formula of DeltaNet:
\begin{align}
    \vect{S}_t = \vect{S}_{t-1}(\vect{I} - \vect{k}_t \vect{k}_t^\top) + \vect{v}_t \vect{k}_t^\top,
\end{align}
we obtain: $\vect{A}_t = \vect{I}$, $\vect{B}_t = \vect{I} - \vect{k}_t \vect{k}_t^\top$, and $\vect{C}_t = \vect{v}_t \vect{k}_t^\top$. Substituting these into Eq.~\ref{eqn:general_opt_obj}, we have:
\begin{align}
    \mathcal{L}_{t}(\vect{S}_{t-1}) = \frac{1}{2}\operatorname{tr}(\vect{S}_{t-1}^\top \vect{S}_{t-1} \vect{k}_t \vect{k}_t^\top) - \operatorname{tr}(\vect{k}_t \vect{v}_t^\top \vect{S}_{t-1}),
\end{align}
where
\begin{align}
    \operatorname{tr}(\vect{S}_{t-1}^\top \vect{S}_{t-1} \vect{k}_t \vect{k}_t^\top) &= \operatorname{tr}((\vect{S}_{t-1} \vect{k}_t) (\vect{S}_{t-1} \vect{k}_t)^\top) \nonumber \\
    &= \|\vect{S}_{t-1} \vect{k}_t \|^2.
\end{align}
Thus,
\begin{align}
    \mathcal{L}_{t}(\vect{S}_{t-1}) &= \frac{1}{2} \|\vect{S}_{t-1} \vect{k}_t \|^2 - \langle \vect{S}_{t-1}\vect{k}_t, \vect{v}_t \rangle \nonumber \\
    &= \frac{1}{2} \|\vect{S}_{t-1} \vect{k}_t - \vect{v}_t \|^2 - \frac{1}{2} \|\vect{v}_t \|^2.
\end{align}
Since $- \frac{1}{2} \|\vect{v}_t\|^2$ is independent of $\vect{S}_{t-1}$ and thus vanishes after taking the gradient, the optimization objective can be expressed as:
\begin{align}
    \mathcal{L}_{t}(\vect{S}_{t-1}) = \frac{1}{2} \|\vect{S}_{t-1} \vect{k}_t - \vect{v}_t \|^2.
\end{align}

Introducing momentum, i.e., $\vect{C}_t = \eta_t \vect{C}_{t-1} + \vect{v}_t \vect{k}_t^\top$, where $\eta$ is the momentum coefficient and $\eta \in (0, 1)$, we expand this recurrent formulation as follows:
\begin{align}
    \vect{C}_t = \sum_{i=1}^{t}\left(\prod_{j=i+1}^{t}\eta_j\right)\vect{v}_i\vect{k}_i^\top.
\end{align}
This corresponds to adding an extra constant-coefficient term to the original $\vect{C}_t$, updating the optimization objective to:
\begin{align}
    \mathcal{L}_{t}(\vect{S}_{t-1}) = \frac{1}{2}\|\vect{S}_{t-1}\vect{k}_t\|^2 - \sum_{i=1}^{t}\left(\prod_{j=i+1}^{t}\eta_j\right)\langle \vect{S}_{t-1}\vect{k}_t, \vect{v}_t\rangle.
\end{align}

\paragraph{Softmax Attention}
Consider the output $\vect{o}_t$:
\begin{align}
    \vect{o}_t &= \sum_{i=1}^t \frac{\exp(\vect{q}_t^\top \vect{k}_i)}{\sum_{j=1}^t \exp(\vect{q}_t^\top \vect{k}_j)} \vect{v}_i \nonumber \\
    &= \sum_{i=1}^t \frac{\phi(\vect{q}_t)^\top \phi(\vect{k}_i)}{\sum_{j=1}^t \exp(\vect{q}_t^\top \vect{k}_j)} \vect{v}_i,
\end{align}
where $\phi(\cdot)$ is defined by Eq.~\ref{eqn:softmax_kernel}. For sufficiently large $t$, according to the law of large numbers, we have $\sum_{j=1}^t \exp(\vect{q}_t^\top \vect{k}_j) \approx t c(\vect{q}_t)$, where $c(\vect{q}_t)$ is a query-specific constant. Thus, the output can be approximated as:
\begin{align}
    \vect{o}_t \approx \frac{1}{c(\vect{q}_t)} \sum_{i=1}^t \frac{\phi(\vect{q}_t)^\top \phi(\vect{k}_i)}{t} \vect{v}_i.
\end{align}
Let $\vect{S}_t = \sum_{i=1}^t \frac{\vect{v}_i \phi(\vect{k}_i)^\top}{t}$; then we have:
\begin{align}
    \vect{S}_t = \frac{t-1}{t} \vect{S}_{t-1} + \frac{1}{t} \vect{v}_t \phi(\vect{k}_t)^\top.
\end{align}
This implies that $\vect{A}_t = \frac{t-1}{t}\vect{I}$, $\vect{B}_t = \vect{I}$, and $\vect{C}_t = \frac{1}{t}\vect{v}_t \phi(\vect{k}_t)^\top$. Substituting these into Eq.~\ref{eqn:general_opt_obj}, we obtain:
\begin{align}
    \mathcal{L}_{t}(\vect{S}_{t-1}) = \frac{1}{t} \left( \frac{1}{2} \|\vect{S}_{t-1} \|_{F}^2 - \langle \vect{S}_{t-1} \phi(\vect{k}_t), \vect{v}_t \rangle \right).
\end{align}

If we introduce the gating matrix $\text{diag}(\vect{\lambda}_t)$, i.e., $\vect{A}_t = \frac{t-1}{t}\text{diag}(\vect{\lambda}_t)$, then:
\begin{align}
    \mathcal{L}_{t}(\vect{S}_{t-1}) = \frac{1}{t}\left(\frac{1}{2}\text{diag}(1 - \vect{\lambda}_t)\|\vect{S}_{t-1}\|_{F}^{2} - \langle \vect{S}_{t-1}\phi(\vect{k}_t), \vect{v}_t \rangle\right).
\end{align}

\section{Delta Rule with Kernel Function}
\label{sec:appendix_delta_rule_with_kernel}
We consider the kernel function $\kappa(\vect{x}, \vect{y}) = f(\vect{x})^\top f(\vect{y})$, where $f(\vect{x})$ is a mapping from $d$ to infinite dimensions. Then the delta-rule-based update form and the corresponding read-out equation can be re-write as:
\begin{align}
    \vect{S}_t &= \vect{S}_{t-1} {\left(\vect{I} - f(\vect{k}_t)f(\vect{k}_t)^\top\right)} + \vect{v}_t f(\vect{k}_t)^\top, \\
    \vect{o}_t &= {\vect{S}_t f(\vect{q}_t)}.
\end{align}
Hypothesis:
\begin{align}
    \vect{S}_t = \sum_{i=1}^t \vect{u}_i \vect{w}_i^\top,
\end{align}
where $\vect{u}_i$ and $\vect{w}_i$ is pending. Then we have:
\begin{align}
    \sum_{i=1}^t \vect{u}_i \vect{w}_i^\top &= \sum_{i=1}^{t-1} \vect{u}_i \vect{w}_i^\top {\left({\vect{I}} - f(\vect{k}_t)f(\vect{k}_t)^\top\right)} + \vect{v}_t f(\vect{k}_t)^\top, \\
    \vect{u}_t \vect{w}_t^\top &= \sum_{i=1}^{t-1} \vect{u}_i \vect{w}_i^\top {\left( -f(\vect{k}_t)f(\vect{k}_t)^\top\right)} + \vect{v}_t f(\vect{k}_t)^\top,
\end{align}
take the pending $\vect{w}_i = f(\vect{k}_i)$:
\begin{align}
     \vect{u}_t f(\vect{k}_t)^\top &= \sum_{i=1}^{t-1} \vect{u}_i f(\vect{k}_i)^\top {\left( - f(\vect{k}_t)f(\vect{k}_t)^\top\right)} + \vect{v}_t f(\vect{k}_t)^\top \nonumber \\
     &= - \sum_{i=1}^{t-1} f(\vect{k}_i)^\top f(\vect{k}_t) \vect{u}_i f(\vect{k}_t)^\top + \vect{v}_t f(\vect{k}_t)^\top \nonumber \\
     &= \left(- \sum_{i=1}^{t-1} f(\vect{k}_i)^\top f(\vect{k}_t) \vect{u}_i + \vect{v}_t \right) f(\vect{k}_t)^\top.
\end{align}
Thus, we get the pending $\vect{u}_t$:
\begin{align}
    \vect{u}_t \bcancel{f(\vect{k}_t)^\top} &= \left(- \sum_{i=1}^{t-1} f(\vect{k}_i)^\top f(\vect{k}_t) \vect{u}_i + \vect{v}_t \right) \bcancel{f(\vect{k}_t)^\top} \nonumber \\
    &= - \sum_{i=1}^{t-1} \kappa(\vect{k}_i, \vect{k}_t) \vect{u}_i + \vect{v}_t.
\end{align}
Then we have the final update form and the corresponding read-out equation:
\begin{align}
    \vect{S}_t &= \sum_{i=1}^t \vect{u}_i f(\vect{k}_i)^\top = \vect{S}_{t-1} + \vect{u}_t \phi(\vect{k}_t)^\top, \\
    \vect{o}_t &= \sum_{i=1}^t \kappa(\vect{k}_i, \vect{q}_t) \vect{u}_i.
\end{align}

\section{Re-examining DeltaNet from the Perspective of Information Aggregation}
\label{sec:appendix_info_aggregation}
In this section, we present our initial motivation and derivation for the combination of DeltaNet \cite{schlag2021linear,yang2024deltanet} and softmax attention. But this differs significantly from the narrative logic in the main text and maybe not as easy to understand as the main text, so it is included in the appendix.

\subsection{Information Aggregation}
We first consider the vanilla Transformer attention. For simplicity, ignoring the exponential kernel and normalization term, the corresponding associative memory update is:
\begin{align}
    \vect{S}_t = \vect{S}_{t-1} + \vect{v}_t \vect{k}_t^\top.
\end{align}

In this form, the stored information $\{\vect{v}_1, \vect{v}_2, \dots\}$ is independent across time steps and can be computed in parallel. Now, consider a new form of value representation $\vect{u}$ defined as:
\begin{align}
    \vect{u}_t = \vect{v}_t - \sum_{i=1}^{t-1} (\vect{k}_i^\top \vect{k}_t)\,\vect{u}_i.
    \label{eqn:deltanet_info_agg}
\end{align}

In this case, the stored information $\{\vect{u}_1, \vect{u}_2, \dots\}$ is no longer independent but has a recurrent dependency structure. Consequently, the associative memory update becomes:
\begin{align}
    \vect{S}_t &= \vect{S}_{t-1} + \vect{u}_t \vect{k}_t^\top \nonumber \\
    &= \vect{S}_{t-1} + \left(\vect{v}_t - \sum_{i=1}^{t-1} (\vect{k}_i^\top \vect{k}_t)\,\vect{u}_i\right) \vect{k}_t^\top \nonumber \\
    &= \vect{S}_{t-1} + \left(\vect{v}_t - \left(\sum_{i=1}^{t-1} \vect{u}_i \vect{k}_i^\top\right)\vect{k}_t\right)\vect{k}_t^\top \nonumber \\
    &= \vect{S}_{t-1} + \left(\vect{v}_t - \vect{S}_{t-1}\vect{k}_t\right)\vect{k}_t^\top \nonumber \\
    &= \vect{S}_{t-1}\left(\vect{I} - \vect{k}_t\vect{k}_t^\top\right) + \vect{v}_t\vect{k}_t^\top.
\end{align}
Surprisingly, we find that this formulation recovers DeltaNet. We have derived DeltaNet from an alternative perspective. The information aggregation method in Eq.~\ref{eqn:deltanet_info_agg} essentially corresponds to DeltaNet.

Intuitively, vanilla attention simply writes each timestep's value into memory. In contrast, the delta rule (Eq.~\ref{eqn:deltanet_info_agg}) first removes redundant information from the current value based on key similarity with historical values, writing only incremental information. Thus, the latter aggregation mechanism supports both memory writing and deletion, whereas the former supports only writing.

\subsection{Enhancing Retrieval Capability of DeltaNet}
Further, we can enhance the information retrieval capability of vanilla DeltaNet by introducing kernel functions. Specifically, we replace the key similarity computation in Eq.~\ref{eqn:deltanet_info_agg} with an exponential kernel function:
\begin{align}
    \vect{u}_t = \vect{v}_t - \sum_{i=1}^{t-1} \frac{\exp(\vect{k}_i^\top \vect{k}_t)}{\vect{Z}_{t}^{(1)}} \vect{u}_i,
    \label{eqn:deltanet_softmax}
\end{align}
where, similar to softmax attention, we introduce a normalization term $\vect{Z}_{t}^{(1)} = \sum_{i=1}^{t-1}\exp(\vect{k}_i^\top\vect{k}_t)$. Consequently, the current output $\vect{o}_t$ is computed as:
\begin{align}
    \vect{o}_t = \sum_{i=1}^{t}\frac{\exp(\vect{k}_i^\top \vect{q}_t)}{\vect{Z}_{t}^{(2)}}\vect{u}_i,
\end{align}
where $\vect{Z}_{t}^{(2)} = \sum_{i=1}^{t}\exp(\vect{k}_i^\top\vect{q}_t)$. Note that the normalization terms $\vect{Z}_{t}^{(1)}$ and $\vect{Z}_{t}^{(2)}$ could alternatively use RMSNorm \cite{zhang2019root} to stabilize training.

In summary, we successfully integrate the delta rule with softmax attention.

\subsection{Efficient Chunk-wise Implementation} \label{sec.appendix.chunkwise}
Consider $\vect{q}, \vect{k}, \vect{v}, \vect{u} \in \mathbb{R}^{d}$, sequence length $T$, and assume $T \gg d$. If we compute $\{\vect{u}_1, \dots, \vect{u}_T\}$ according to Eq.~\ref{eqn:deltanet_softmax}, the computational complexity is $\mathcal{O}(T^2 d)$. Note that, since each $\vect{u}_t$ depends on all previous $\vect{u}_{<t}$, this algorithm cannot be parallelized, resulting in a training-time complexity of $\mathcal{O}(T)$, significantly slower than the $\mathcal{O}(1)$ complexity of standard softmax attention. Consequently, it fails to utilize GPUs' parallel computation capabilities.

In fact, the computations of $\{\vect{u}_1, \dots, \vect{u}_T\}$ can be parallelized at the cost of increased computes. Rewriting Eq.~\ref{eqn:deltanet_softmax} in matrix form, we have:
\begin{align}
    \vect{U} = \vect{V} - \vect{A}\vect{U},
\end{align}
where the $t$-th rows of matrices $\vect{U}$ and $\vect{V}$ are $\vect{U}_{t,:} = \vect{u}_t$ and $\vect{V}_{t,:} = \vect{v}_t$, respectively. The similarity matrix $\vect{A}$ is defined as $\vect{A}_{t,i} = \frac{\exp(\vect{k}_i^\top \vect{k}_t)}{\vect{Z}_t^{(1)}}$ and is lower-triangular. Thus, $\vect{U}$ can be solved via matrix operations:
\begin{align}\label{eq:pa}
    \vect{U} = (\vect{I} + \vect{A})^{-1}\vect{V}.
\end{align}
This approach requires a matrix inversion operation, increasing computational complexity to $\mathcal{O}(T^3)$, but can be paralleled in GPUs.

Intuitively, we can seek a trade-off between the two approaches described above. By dividing the sequence of length $T$ into $N$ equally sized chunks, each having length $C = \lfloor \frac{T}{N} \rfloor$, we can sequentially compute the sequence of $\vect{u}$ for each chunk using parallelized matrix operations. As a result, the recurrent steps from $\mathcal{O}(T)$ to $\mathcal{O}(N)$, while the computational complexity transitions from $\mathcal{O}(T^2 d)$ to $\mathcal{O}(T^2d + TCd + TC^2)$. Essentially, this approach trades computational resources for reduced runtime.

Below is a simple PyTorch implementation, serving as pseudo-code. We can easily modify the selection of the kernel function or remove the normalization term.

\begin{lstlisting}[language=Python, caption=PyTorch-style pseudo-code.]
import torch
import torch.nn.functional as F
import math

def flash_attn(K_chunk, K_prev, V_prev): 
    attn = K_chunk @ K_prev.transpose(-1, -2)/math.sqrt(K_chunk.shape[-1])
    z_intra = torch.logsumexp(attn, dim=-1)
    return torch.softmax(attn,dim=-1)@V_prev, z_intra

def naive_implementation(k, n, d_model): """n is the previous v, v is actually new v. """
    B, H, T, D = k.shape
    v = torch.zeros_like(n)
    for t in range(T):
        if t == 0:
            v[:, :, 0] = n[:, :, 0]
        else:
            scores = torch.matmul(k[:, :, :t], k[:, :, t].unsqueeze(-1)).squeeze(-1) / math.sqrt(d_model)
            attn_probs = F.softmax(scores, dim=-1)
            v[:, :, t] = n[:, :, t] - torch.sum(attn_probs.unsqueeze(-1) * v[:, :, :t], dim=-2)
    return v

def optimized_chunked_implementation(K, N, d_model, C): 
    B, H, T, D = K.shape
    V = torch.zeros(B, H, T, D)
    chunk_nums = T // C
    mask = torch.tril(torch.ones(C, C),diagonal=-1).unsqueeze(0).unsqueeze(0).to(K.device)
    for chunk_num in range(chunk_nums):
        start = chunk_num * C
        end = (chunk_num + 1) * C
        K_chunk = K[:, :, start:end]
        N_chunk = N[:, :, start:end]
        if chunk_num > 0:
            intra_output, Z_intra = flash_attn(K_chunk, K[:, :, :start], V[:, :, :start])#\mathcal{O}(TCD)
            A = (K_chunk @ K_chunk.transpose(-2, -1)).masked_fill(mask[:, :, :C, :C] == 0, float("-inf"))  / math.sqrt(d_model)#\mathcal{O}(C^2D)
            Z_inter = torch.logsumexp(A, dim=-1)
            P = N_chunk - intra_output * (1/(1 + torch.exp((Z_inter-Z_intra).unsqueeze(-1))))
            A = F.softmax(A, dim=-1) * (1/(1 + torch.exp((Z_intra-Z_inter).unsqueeze(-1))))
            A[:,:,0,:] = 0
        else:
            A = (K_chunk @ K_chunk.transpose(-2, -1)).masked_fill(mask[:, :, :C, :C] == 0, float("-inf"))  / math.sqrt(d_model)
            A = F.softmax(A, dim=-1)
            A[:,:,0,:] = 0
            P = N_chunk
        Ti = torch.eye(C).unsqueeze(0).unsqueeze(0).unsqueeze(0).to(K.device) + A
        Ti_inverse = torch.inverse(Ti) ## Forward substitution method \mathcal{O}(C^3) Each block can be solved in parallel if we don't use the normalization of softmax. To fully utilize GPUs,  we can use \mathcal{O}(log C) iteration, each iteration use \mathcal{O}(C^3) flops, such as $a_1,b_1 = I - A, A^2$, $a_{i+1},b_{i+1} = a_{i}@b_{i} + b_{i}@b_{i}$, return a_{log C - 1} 
        V[:, :, start:end] = Ti_inverse @ P     # \mathcal{O}(C^2D)                     
    return V    #\mathcal{O}(T/C * (TCD + C^2D)) = \mathcal{O}(T^2D + TCD + TC^2)

def verify_equivalence():
    B = 2
    H = 2
    T = 1024
    D = 64
    C = 32
    K = torch.randn(B, H, T, D)
    N = torch.randn(B, H, T, D)
    naive_output = naive_implementation(K, N, D)
    optimized_output = optimized_chunked_implementation(K, N, D, C)
    equivalence = torch.allclose(naive_output, optimized_output, atol=1e-5)
    print(f"{equivalence}")
\end{lstlisting}

\subsubsection{Efficient Matrix Inversion on GPUs.}
In our method, the matrix inversion of a block is an important part. Previous work \cite{yang2024deltanet} uses a hybrid approach of forward substitution methods and block matrix inversion, but fail to address the issue of forward substitution method not efficiently utilizing GPU parallel capabilities. To solve this problem, we study the efficient implementation of inverse of a lower triangular matrix with diagonal 1 on GPU.

Assuming the matrix we are studying is $\vect{I}+\vect{A}$, where $\vect{A}$ is a strict lower triangular matrix with a diagonal of 0 and a size of $C \times C$. Under the setting of \cite{yang2024deltanet}, $C$ is generally 64. We can find that $\vect{A}^C = 0$.

Then we can get:
\begin{align}
    (\vect{I}+\vect{A})(\vect{I}-\vect{A})\prod_{i=1}^{\log_2 C - 1}(\vect{I}+\vect{A}^{2^i}) = \vect{I} - \vect{A}^C = \vect{I},
\end{align}
which means
\begin{align}
    (\vect{I}+\vect{A})^{-1} = (\vect{I}-\vect{A})\prod_{i=1}^{\log_2 C - 1}(\vect{I}+\vect{A}^{2^i})
\end{align}
Then we can use the following iteration to calculate $(\vect{I}+\vect{A})^{-1}$:
\begin{align}
    [\vect{X}_0,\vect{Y}_0] &= [\vect{I}-\vect{A}, \vect{A}^2] \nonumber \\
    [\vect{X}_{i},\vect{Y}_{i}] &= [\vect{X}_{i-1},\vect{0}] + [\vect{X}_{i-1},\vect{Y}_{i-1}]\vect{Y}_{i-1}, i=1, \dots, \log_2 C - 1,
\end{align}
finally, $(\vect{I}+\vect{A})^{-1} = \vect{X}_{\log_2C - 1}$.

\section{State Tracking of DeltaFormer}
\label{sec:state_tracking_delt_attn}
We first propose and prove a theorem related to state exchange. Then, we discuss that under certain conditions, DeltaFormer could express the attention described by the aforementioned theorem. Finally, we analyze the spatial complexity advantages of DeltaFormer in state tracking.

\begin{tcolorbox}[colback=gray!10, colframe=gray!70]
\begin{theorem}[State Exchange]
\label{th:state_exchange}
$\\$

\textbf{Assumption 1:} There exist $n$ state points on a $d$-dimensional unit sphere, and the absolute value of the inner product of any two distinct state points is less than or equal to $\epsilon(d,n)$, which means:
\begin{align}
    \exists \, \vect{x}_1, \vect{x}_2, \dots, \vect{x}_n \in \mathbb{R}^{d} \quad \text{s.t.} \quad \|\vect{x}_i\|_2 = 1 (\forall i), \quad \max_{i \neq j} |\vect{x}_i^\top \vect{x}_j| \leq \epsilon(d,n) < \frac{1}{8}.
    \nonumber
\end{align}

\textbf{Assumption 2:} There is a function $f$ satisfies: 
\begin{align}
    \forall \vect{x} \in \{-1,0,1,2\}, \, \forall \tilde{\vect{x}} \in U(\vect{x}, 4\epsilon(d,n)): \quad f(\tilde{\vect{x}}) = \vect{x}.
    \nonumber
\end{align}

$\\$

Consider initializing $n$ key-value pairs as $\{(\vect{k}_1, \vect{v}_1), \dots,(\vect{k}_n, \vect{v}_n)\}$. The keys $\{\vect{k}_1, \dots, \vect{k}_n\}$ lie on a $d$-dimensional unit sphere and satisfies Assumption 1, which means:
\[
    \forall \, i,j\in\{1,\dots,n\}, \, i\neq j: \quad \|\vect{k}_i\|_2 = 1, \quad |\vect{k}_i^\top \vect{k}_j| \leq \epsilon(n,d).
\]

$\\$

Define an attention mechanism as follows:
\[
    \vect{u}_t = \vect{v}_t - \sum_{i=1}^{t-1} f(\vect{k}_i^\top \vect{k}_t) \vect{u}_i,\quad \vect{o}_t = \sum_{i=1}^{t} f(\vect{q}_t^\top \vect{k}_i) \vect{u}_i,
\]
where $f(\cdot)$ satisfies Assumption 2 and it is noted that $\forall \, i \in \{1,\dots,n\}$, since $f(\vect{k}_i^\top \vect{k}_i) = 0$, we have $\vect{u}_i = \vect{v}_i$.

$\\$

At the current step $t, \, t>n$, the value corresponding to $\vect{k}_i$ is denoted by $\tilde{\vect{v}}_i$, $i \in \{1, \dots, n\}$. Note that, after $t-1-n$ exchanges, $\tilde{\vect{v}}_i$ is not necessarily equal to the initially assigned $\vect{v}_i$. $\forall \, 1 \leq t_2 < t_1 \leq n$, to exchange the stored values $\tilde{\vect{v}}_{t_1}$ and $\tilde{\vect{v}}_{t_2}$ corresponding to $\vect{k}_{t_1}$ and $\vect{k}_{t_2}$, it suffices to construct:
\[
    \vect{k}_t = \vect{k}_{t_1} - \vect{k}_{t_2}, \quad \vect{v}_t = 0
\]

$\\$

When retrieving the values:
\begin{itemize}
    \item Query $\vect{q}_t = \vect{k}_{t_1}$, then $\vect{o}_t = \tilde{\vect{v}}_{t_2}$;
    \item Query $\vect{q}_t = \vect{k}_{t_2}$, then $\vect{o}_t = \tilde{\vect{v}}_{t_1}$;
    \item Query $\vect{q}_t = \vect{k}_{t_3}$, $1\leq t_3\leq n$, $t_3\neq t_1, t_2$, then $\vect{o}_t = \tilde{\vect{v}}_{t_3}$.
\end{itemize}
This implies the exchange of values corresponding to $k_{t_1}$ and $k_{t_2}$ is completed.
\end{theorem}
\end{tcolorbox}

\subsection{Proof of Theorem~\ref{th:state_exchange}}
Before proving Theorem~\ref{th:state_exchange}, we introduce an auxiliary lemma for facilitating the proof.

\begin{tcolorbox}[colback=gray!10, colframe=gray!70]
\begin{lemma}\label{lemma:auxiliary}
Consider Theorem~\ref{th:state_exchange}, the set of keys $\{\vect{k}_i\}_{i=1}^n$ satisfies Assumption 1, and $\vect{k}_{>n}$ is the difference between two keys chosen from $\{\vect{k}_i\}_{i=1}^{n}$. If the function $f(\cdot)$ satisfies Assumption 2, then the following identity holds:
\begin{align}
    \forall\, 1 \le j < i \le n,\,\forall\, l \ge 1:\quad f((\vect{k}_i - \vect{k}_j)^\top \vect{k}_l) = f(\vect{k}_i^\top\vect{k}_l) - f(\vect{k}_j^\top\vect{k}_l).
    \nonumber
\end{align}
\end{lemma}
\end{tcolorbox}

\subsubsection{Proof of Lemma~\ref{lemma:auxiliary}}
We distinguish two separate cases according to the value of the index $l$:

\textbf{Case 1: $1 \le l \le n$.}
Consider the following subcases:
\begin{itemize}
    \item[i.] If $\vect{k}_i = \vect{k}_l$, then we obtain
    \[
        f((\vect{k}_i - \vect{k}_j)^\top \vect{k}_l) = 
        f(1 - \vect{k}_j^\top \vect{k}_l) = 
        f(U(1, \epsilon)) =
        1
    \]
    \[
        f(\vect{k}_i^\top\vect{k}_l) - f(\vect{k}_j^\top\vect{k}_l) = 
        1 - f(U(0, \epsilon)) = 1.
    \]

    \item[ii.] If $\vect{k}_j = \vect{k}_l$, then we have
    \[
        f((\vect{k}_i - \vect{k}_j)^\top \vect{k}_l) = 
        f(\vect{k}_i^\top \vect{k}_l - 1) = 
        f(U(-1, \epsilon)) = -1
    \]
    \[
        f(\vect{k}_i^\top\vect{k}_l) - f(\vect{k}_j^\top\vect{k}_l) = f(U(0,\epsilon)) - 1 = -1.
    \]

    \item[iii.] If $\vect{k}_i \ne \vect{k}_l$, $\vect{k}_j \ne \vect{k}_l$, then
    \[
        f((\vect{k}_i - \vect{k}_j)^\top \vect{k}_l) = 
        f(\vect{k}_i^\top \vect{k}_l - \vect{k}_j^\top \vect{k}_l) = 
        f(U(0,2\epsilon)) = 0
    \]
    \[
        f(\vect{k}_i^\top\vect{k}_l) - f(\vect{k}_j^\top\vect{k}_l) = 
        f(U(0,\epsilon)) - f(U(0,\epsilon)) = 0.
    \]
\end{itemize}

\textbf{Case 2: $l > n$.}
In this case, denote $\vect{k}_l = \vect{k}_{l_1} - \vect{k}_{l_2}$, where $1 \le l_2 < l_1 \le n$. Consider the following possibilities regarding the number of equalities among indices $i, j$ and $l_1, l_2$:
\begin{itemize}
    \item[i.] If no pair among $(i, j)$ and $(l_1, l_2)$ is equal, then we have
    \[
        f((\vect{k}_i - \vect{k}_j)^\top \vect{k}_l) = 
        f(U(0,4\epsilon)) = 0
    \]
    \[
        f(\vect{k}_i^\top\vect{k}_l) - f(\vect{k}_j^\top\vect{k}_l) = 
        f(U(0,2\epsilon)) - f(U(0,2\epsilon)) = 0.
    \]

    \item[ii.] If exactly one pair is equal, we analyze further:
    \begin{enumerate}
        \item If $i = l_1$, then we have
        \[
            f((\vect{k}_i - \vect{k}_j)^\top \vect{k}_l) = 
            f(U(1,3\epsilon)) = 1
        \]
        \[
            f(\vect{k}_i^\top\vect{k}_l) - f(\vect{k}_j^\top\vect{k}_l) = 
            f(U(1,\epsilon)) - f(U(0,2\epsilon)) = 1.
        \]

        \item If $i = l_2$, then we have
        \[
            f((\vect{k}_i - \vect{k}_j)^\top \vect{k}_l) = 
            f(U(-1,3\epsilon)) = -1
        \]
        \[
            f(\vect{k}_i^\top\vect{k}_l) - f(\vect{k}_j^\top\vect{k}_l) = 
            f(U(-1,\epsilon)) - f(U(0,2\epsilon)) = -1.
        \]

        \item If $j = l_1$, then similarly
        \[
            f((\vect{k}_i - \vect{k}_j)^\top \vect{k}_l) = 
            f(U(-1,3\epsilon)) = -1
        \]
        \[
            f(\vect{k}_i^\top\vect{k}_l) - f(\vect{k}_j^\top\vect{k}_l) = 
            f(U(0,2\epsilon)) - f(U(1,\epsilon)) = -1.
        \]

        \item If $j = l_2$, then similarly
        \[
            f((\vect{k}_i - \vect{k}_j)^\top \vect{k}_l) = 
            f(U(1,3\epsilon)) = 1
        \]
        \[
            f(\vect{k}_i^\top\vect{k}_l) - f(\vect{k}_j^\top\vect{k}_l) = 
            f(U(0,2\epsilon)) - f(U(-1,\epsilon)) = 1.
        \]
    \end{enumerate}

    \item[iii.] If two pairs are equal simultaneously:
    \begin{enumerate}
        \item If $i = l_1, j = l_2$, we have
        \[
            f((\vect{k}_i - \vect{k}_j)^\top \vect{k}_l) = 
            f(U(2,2\epsilon)) = 2
        \]
        \[
            f(\vect{k}_i^\top\vect{k}_l) - f(\vect{k}_j^\top\vect{k}_l) = 
            f(U(1,\epsilon)) - f(U(-1,\epsilon)) = 2.
        \]

        \item If $i = l_2, j = l_1$, this contradicts the ordering condition $j < i, l_2 < l_1$ and thus cannot occur.
    \end{enumerate}
\end{itemize}

Combining all the above cases, we have completed the proof.

\subsubsection{Formally Prove Theorem~\ref{th:state_exchange}}
We use mathematical induction to prove Theorem~\ref{th:state_exchange}.

When $t = n + 1$:
\begin{align}
    \vect{k}_t &= \vect{k}_{t_1} - \vect{k}_{t_2}, \\
    \vect{u}_t &= -\sum_{i=1}^{t-1} f(\vect{k}_i^\top \vect{k}_t) \vect{u}_i = -\vect{u}_{t_1} + \vect{u}_{t_2}.
\end{align}

If we read the state at $t_1$, i.e., $\vect{q}_t = \vect{k}_{t_1}$,
\begin{equation}
    \sum_{i=1}^{t} f(\vect{q}_t^\top \vect{k}_i) \vect{u}_i = \sum_{i=1}^{t} f(\vect{k}_{t_1}^\top \vect{k}_i) \vect{u}_i = \vect{u}_{t_1} + (-\vect{u}_{t_1} + \vect{u}_{t_2}) = \vect{u}_{t_2}.
\end{equation}

If we read the state at $t_2$, i.e., $\vect{q}_t = \vect{k}_{t_2}$,
\begin{equation}
    \sum_{i=1}^{t} f(\vect{q}_t^\top \vect{k}_i) \vect{u}_i = \sum_{i=1}^{t} f(\vect{k}_{t_2}^\top \vect{k}_i) \vect{u}_i = \vect{u}_{t_2} + (\vect{u}_{t_1} - \vect{u}_{t_2}) = \vect{u}_{t_1}.
\end{equation}

If we read other states, i.e., the state at $j$, where $j \neq t_1, t_2$,
\begin{equation}
    \sum_{i=1}^{t} f(\vect{q}_t^\top \vect{k}_i) \vect{u}_i = \sum_{i=1}^{t} f(\vect{k}_j^\top \vect{k}_i) \vect{u}_i = \vect{u}_j.
\end{equation}

In summary, at step $t = n + 1$, according to our rules, it is possible to trace the states exchanged between $t_1$ and $t_2$.

Assuming the proposition holds for $t - 1$, we consider the case for $t$ ($t > n + 1$).

At the $t$-th step,
\begin{align}
    \vect{k}_t = \vect{k}_{t_1} - \vect{k}_{t_2}.
\end{align}
According to Lemma 1, we have
\begin{align}
    \vect{u}_t &= -\sum_{i=1}^{t-1} f(\vect{k}_t^\top \vect{k}_i) \vect{u}_i \nonumber \\
    &= -\sum_{i=1}^{t-1} f(\vect{k}_{t_1}^\top \vect{k}_i) \vect{u}_i + \sum_{i=1}^{t-1} f(\vect{k}_{t_2}^\top \vect{k}_i) \vect{u}_i \nonumber \\
    &= -\vect{\tilde{v}}_{t_1} + \vect{\tilde{v}}_{t_2}.
\end{align}

If we read the state at $t_1$, i.e., $\vect{q}_t = \vect{k}_{t_1}$,
\begin{align}
    \sum_{i=1}^{t} f(\vect{q}_t^\top \vect{k}_i) \vect{u}_i &= \sum_{i=1}^{t} f(\vect{k}_{t_1}^\top \vect{k}_i) \vect{u}_i \nonumber \\
    &= \sum_{i=1}^{t-1} f(\vect{k}_{t_1}^\top \vect{k}_i) \vect{u}_i + f(\vect{k}_{t_1}^\top \vect{k}_t) \vect{u}_t \nonumber \\
    &= \tilde{\vect{v}}_{t_1} + (-\tilde{\vect{v}}_{t_1} + \tilde{\vect{v}}_{t_2}) \nonumber \\
    &= \tilde{\vect{v}}_{t_2}.
\end{align}

If we read the state at $t_2$, i.e., $\vect{q}_t = \vect{k}_{t_2}$,
\begin{align}
    \sum_{i=1}^{t} f(\vect{q}_t^\top \vect{k}_i) \vect{u}_i &= \sum_{i=1}^{t} f(\vect{k}_{t_2}^\top \vect{k}_i) \vect{u}_i \nonumber \\
    &= \sum_{i=1}^{t-1} f(\vect{k}_{t_2}^\top \vect{k}_i) \vect{u}_i + f(\vect{k}_{t_2}^\top \vect{k}_t) \vect{u}_t \nonumber \\
    &= \tilde{\vect{v}}_{t_2} + (\tilde{\vect{v}}_{t_1} - \tilde{\vect{v}}_{t_2}) \nonumber \\
    &= \tilde{\vect{v}}_{t_1}.
\end{align}

If we read other states, i.e., the state at $j$, where $j \neq t_1, t_2$,
\begin{align}
    \sum_{i=1}^{t} f(\vect{q}_t^\top \vect{k}_i) \vect{u}_i &= \sum_{i=1}^{t} f(\vect{k}_j^\top \vect{k}_i) \vect{u}_i \nonumber \\
    &= \sum_{i=1}^{t-1} f(\vect{k}_j^\top \vect{k}_i) \vect{u}_i + f(\vect{k}_j^\top \vect{k}_t) \vect{u}_t \nonumber \\
    &= \tilde{\vect{v}}_{j}.
\end{align}

In summary, at step $t$, according to our rules, the retrieved states corresponding to $\{\vect{k}_1, \dots, \vect{k}_n\}$ is correct.

By mathematical induction, regardless of how large the exchange step $t$ is, the model can always trace the exchange of $n$ states.

\subsection{Re-exam Assumptions} \label{appendix:re-exam}
Now we re-exam the two assumptions in Theorem~\ref{th:state_exchange}. 

What relationship between $d$ and $n$ must hold for Assumption 1 to be satisfied? According to Theorem 5.2.1 in \cite{zhao2022probabilistic},
\begin{tcolorbox}[colback=gray!10, colframe=gray!70]
    \textit{For every $\alpha \in (0, 1)$ and $\varepsilon > 0$, there exists $c > 0$ such that for every $d$, one can find at least $2^{cd}$ unit vectors in $\mathbb{R}^d$ whose pairwise inner products all lie in $[\alpha - \varepsilon, \alpha + \varepsilon]$.}
\end{tcolorbox}
Setting $\alpha = 0.01$ and $\varepsilon = 0.1$, we have $\epsilon(d,n) \leq 0.11 < \frac{1}{8}$. This implies that for Assumption 1 to hold, it is required that $d = \mathcal{O}(\log n)$.

Regarding the choice of $f(\cdot)$ in Assumption 2, intuitively, we can use the rounding function $round(\cdot)$, i.e., setting the input to the nearest integer. Under appropriate rounding precision, this function can satisfy the assumption. But it faces optimization problems that require careful design. Therefore, a natural idea is to use a combination of multiple easily optimized kernels to approximate the round kernel. Given the KU cache size, it is natural to think of using grouped-query attention (GQA)~\cite{ainslie2023gqa} to implement it, which is why we decouple $\vect{k}$ in $\kappa_1$ into $\vect{w}$ and $\vect{k}$ in the general form. After decoupling, $\vect{w}$ can use GQA to improve performance without increasing KU cache:
\begin{align}
    \vect{u}_t &= \alpha_t \vect{v}_t - \beta_t\sum_{i=1}^{t-1}  \sum_{j =1}^{G}a_j \kappa_1(\vect{k}_i, \vect{w}_t^j) \vect{u}_i,
\end{align}
where $G$ is the group number, $a_j \in R$ is the weight of each query is similar to the role of $\vect{o}\ proj$ in standard attention. For example, when $G = 4$, this method can fully fit at least the four points of $\{ -1, 0, 1, 2\}$. It means that, theoretically, a group query attention with four shared heads can express an $f(\cdot)$ that fulfills Assumption 2. If we consider $\vect{w}_t^j = j \vect{k}_t$,  we can always set $a_1,a_2,a_3,a_4$ to make $\sum_{j=1}^4 a_j \exp(j\vect{x}) = \vect{x}$, when $\vect{x} \in \{-1,0,1,2\}$, then we can find that we can track at least $\vect{S}_d$.  On the other hand, this group query attention on $\vect{u}$ does not improve the ability of linear, so the combination of linear is still linear. 

\begin{figure}[ht]
    \centering
    \begin{subfigure}[b]{0.3\textwidth}
        \includegraphics[width=\textwidth]{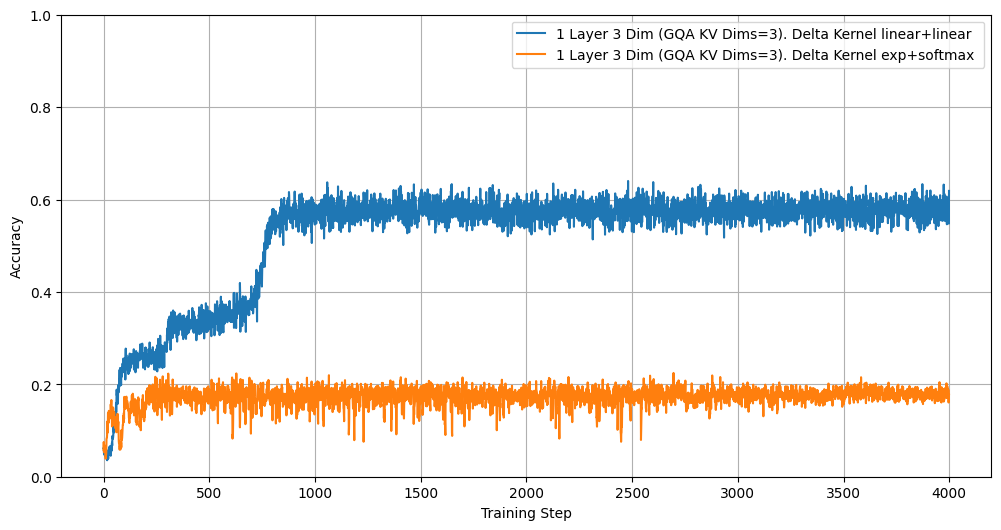}
        \caption{Query number = 1. }
    \end{subfigure}
    ~ 
    \begin{subfigure}[b]{0.3\textwidth}
        \includegraphics[width=\textwidth]{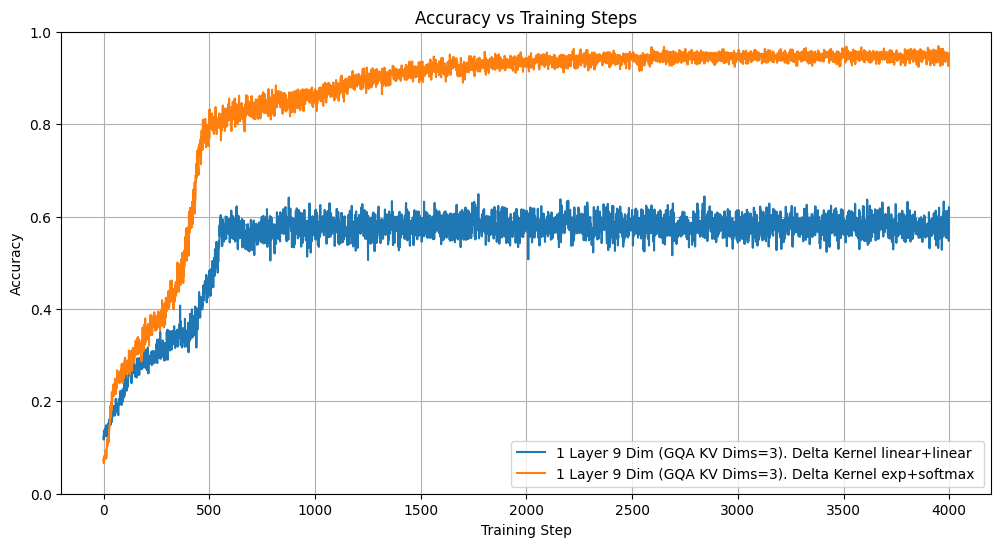}
        \caption{Query number = 3.}
    \end{subfigure}
        ~
    \begin{subfigure}[b]{0.3\textwidth}
        \includegraphics[width=\textwidth]{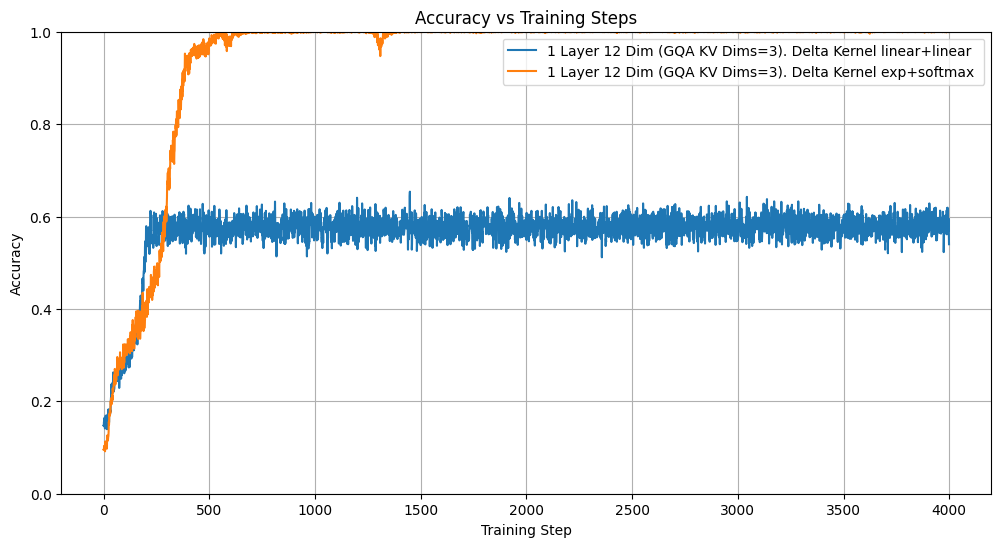}
        \caption{Query number = 4.}
    \end{subfigure}
    \caption{Comparison of  DeltaFormer with different $\kappa_1$ and $\kappa_2$ under different query heads numbers. }
    \label{fig:gqa}
\end{figure}

Next is an interesting experiment to verify this thing. We adopt the GQA-like method, head dim=3, to track the exchange of 5 elements, exchange 16 times. As shown in Figure \ref{fig:gqa}, as the number of queries increases, DeltaFormer with nonlinear kernel's ability to perform $S_5$ becomes stronger and stronger.  So, in the case of using nonlinear kernels,  GQA-like method can also implicitly increase the complexity of the kernel, thereby increasing the capacity of the model.

In summary, according to Theorem~\ref{th:state_exchange}, DeltaFormer can achieve state exchange between historical time steps $t_1$ and $t_2$, provided that $d = \mathcal{O}(\log n)$ and the projection matrices $\vect{W}_k$ and $\vect{W}_v$ learn mappings such that $\vect{k}_t = \vect{k}_{t_1} - \vect{k}_{t_2}$ and $\vect{v}_t = 0$.

\subsection{Stronger Compression Capability}
What is the space cost in Theorem \ref{th:state_exchange}, if we want to track $T$ exchanges of $n$ states? Obviously, we need to set $d = \mathcal{O}(\log n)$ and use KU cache with length $T$. So it needs $\mathcal{O}(T\log n)$ spaces.  

If we read out $\{\tilde{\vect{v}}_1, \dots, \tilde{\vect{v}}_n\}$ based on $\{\vect{k}_1, \dots, \vect{k}_n\}$ every $\mathcal{O}(n)$ steps, then re-write $\{(\vect{k}_i, \tilde{\vect{v}}_i)\}_{i=1}^n$ into KU cache. We only use the KU cache with length $\mathcal{O}(n)$, so the total space cost is $\mathcal{O}(n \log n)$, which is much smaller than the $\mathcal{O}(n^2)$ space when $f(\cdot)$ is an identity mapping, in which \citet{peng2025rwkv} use a $5 \times 5$ matrix to track the exchange of $5$ elements. 

For comparison, employing non-linear kernels  unlocks the powerful potential of the delta rule: we can track the exchange of exponentially many states instead of merely the previous $n = \mathcal{O}(d)$ states. Analogously, softmax attention can retrieve richer information from substantially longer contexts compared to linear attention.

\subsection{Toy Experiment}
\label{subsec:deltaformer_toy_model}
We provide a reference code for DeltaFormer to complete the state tracking toy experiment.
\begin{lstlisting}[language=Python, caption=Toy model code for exploration.]
import torch
import torch.nn as nn
import math
import torch
from torch.utils.data import Dataset, DataLoader
import random
from transformers import AdamW, get_scheduler
from tqdm import tqdm
import numpy as np
from apex.normalization import FusedRMSNorm as RMSNorm
from matplotlib import pyplot as plt
# seed all
def seed_all(seed):
    random.seed(seed)
    np.random.seed(seed)
    torch.manual_seed(seed)
    torch.cuda.manual_seed_all(seed)
    torch.backends.cudnn.deterministic = True
    torch.backends.cudnn.benchmark = False


seed_value = 42
seed_all(seed_value)

class RoundSTE(torch.autograd.Function):
    @staticmethod
    def forward(ctx, x):
        return x.round()
    @staticmethod
    def backward(ctx, grad_output):
        return grad_output 

class Embedding(nn.Module):
    def __init__(self, vocab_size, dim):
        super().__init__()
        self.wte = nn.Embedding(vocab_size, dim)
    def forward(self, x):
        return self.wte(x)

class Unembedding(nn.Module):
    def __init__(attn, dim, vocab_size):
        super().__init__()
        attn.lm_head = nn.Linear(dim, vocab_size, bias=False)
    def forward(self, x):
        return self.lm_head(x)

# Generalized Attention supporting MQA, GQA, Full MHA
class MultiHeadAttention(nn.Module):
    def __init__(self, dim, num_q_heads=8, num_kv_heads=None, rotary_base=10000):
        super().__init__()
        self.num_q_heads = num_q_heads
        self.num_kv_heads = num_kv_heads or num_q_heads  # if None, full MHA
        self.rotary_base = rotary_base
        self.head_dim = dim // self.num_q_heads
        self.q_proj = nn.Linear(dim, dim)
        self.k_proj = nn.Linear(dim, self.head_dim * num_kv_heads)
        self.v_proj = nn.Linear(dim, self.head_dim * num_kv_heads)
        self.out = nn.Linear(dim, dim)

    def forward(self, x):
        b, t, d = x.size()
        # project
        q = self.q_proj(x).view(b, t, self.num_q_heads, self.head_dim).transpose(1, 2)
        k = self.k_proj(x).view(b, t, self.num_kv_heads, self.head_dim).transpose(1, 2)
        v = self.v_proj(x).view(b, t, self.num_kv_heads, self.head_dim).transpose(1, 2)
        # compute attention scores: broadcast kv heads to q heads if needed
        k_expanded = k.unsqueeze(2).expand(-1, -1, self.num_q_heads // self.num_kv_heads, -1, -1).reshape(b, -1, t, self.head_dim)
        v_expanded = v.unsqueeze(2).expand(-1, -1, self.num_q_heads // self.num_kv_heads, -1, -1).reshape(b, -1, t, self.head_dim)
        # q: b,t,Qh,hd, k_expanded: b,t,Qh,hd
        scores = torch.einsum('bhtd,bhsd->bhts', q, k_expanded) / math.sqrt(d // self.num_q_heads)
        attn_mask = torch.tril(torch.ones(t, t)).unsqueeze(0).unsqueeze(1).to(x.device)
        scores = scores.masked_fill(attn_mask == 0, float('-inf'))
        attn = torch.softmax(scores, dim=-1).to(x.dtype)
        out = torch.einsum('bhts,bhsd->bthd', attn, v_expanded)
        out = out.reshape(b, t, d)
        return self.out(out)

class DeltaAttention(MultiHeadAttention):
    def __init__(self, dim, num_q_heads=8, num_kv_heads=None, rotary_base=10000, kernel_1='linear', kernel_2='linear'):
        super().__init__(dim, num_q_heads, num_kv_heads, rotary_base)
        self.alpha = nn.Parameter(torch.ones(1), requires_grad=True)
        self.beta = nn.Parameter(torch.ones(1), requires_grad=True)
        self.kernel_1 = kernel_1
        self.kernel_2 = kernel_2
        self.w_proj = nn.Linear(dim, dim)
        self.w_weight =  nn.Parameter(torch.randn(1,num_q_heads,1,1), requires_grad=True)
    def forward(self, x):
        b, t, d = x.size()
        # project
        q = self.q_proj(x).view(b, t, self.num_q_heads, self.head_dim).transpose(1, 2)
        w = self.w_proj(x).view(b, t, self.num_q_heads, self.head_dim).transpose(1, 2)
        k = self.k_proj(x).view(b, t, self.num_kv_heads, self.head_dim).transpose(1, 2)
        v = self.v_proj(x).view(b, t, self.num_kv_heads, self.head_dim).transpose(1, 2)
        # compute attention scores: broadcast kv heads to q heads if needed
        k_expanded = k.unsqueeze(2).expand(-1, -1, self.num_q_heads // self.num_kv_heads, -1, -1).reshape(b, -1, t, self.head_dim)
        # v_expanded = v.unsqueeze(2).expand(-1, -1, self.num_q_heads // self.num_kv_heads, -1, -1).reshape(b, -1, t, self.head_dim)

        A = torch.einsum("b h s d, b h t d-> b h s t", w, k_expanded) / math.sqrt(d // self.num_q_heads)

        attn_mask = torch.tril(torch.ones(t, t),-1).unsqueeze(0).unsqueeze(1).to(x.device)
        A = A.masked_fill(attn_mask == 0, 0)
        if self.kernel_1 == 'linear':
            A = A.float()
        elif self.kernel_1 == 'round':
            A = (RoundSTE.apply(A)).float()
        elif  self.kernel_1 == 'exp':
            A = torch.exp(A).float()
        elif self.kernel_1 == 'softmax':
            A = torch.softmax(A, dim=-1).float()
        elif self.kernel_1 == 'relu':
            A = torch.relu(A).float()


        A = A*self.w_weight
        group_size = self.num_q_heads // self.num_kv_heads
        A_reshaped = A.view(b, self.num_kv_heads, group_size, t, t)
        A = A_reshaped.mean(dim=2)  # shape: (B, num_kv_heads, s, s)


        u = torch.linalg.solve_triangular(self.beta * A, self.alpha * v, upper=False, unitriangular=True).to(k_expanded)
        u = u.unsqueeze(2).expand(-1, -1, self.num_q_heads // self.num_kv_heads, -1, -1).reshape(b, -1, t, self.head_dim)
        scores = torch.einsum('bhtd,bhsd->bhts', q, k_expanded) / math.sqrt(d // self.num_q_heads)
        attn_mask = torch.tril(torch.ones(t, t)).unsqueeze(0).unsqueeze(1).to(x.device)

        if self.kernel_2 == 'linear':
            attn = scores.masked_fill(attn_mask == 0, 0)
            out = torch.einsum('bhts,bhsd->bthd', attn, u)
        elif self.kernel_2 == 'round':
            attn = (RoundSTE.apply(scores)).masked_fill(attn_mask == 0, 0)
            out = torch.einsum('bhts,bhsd->bthd', attn, u)
        elif self.kernel_2 == 'softmax':
            scores = scores.masked_fill(attn_mask == 0, float('-inf'))
            attn = torch.softmax(scores, dim=-1).to(x.dtype)
            out = torch.einsum('bhts,bhsd->bthd', attn, u)
        elif self.kernel_2 =='relu':
            attn = torch.relu(scores).masked_fill(attn_mask == 0, 0)
            out = torch.einsum('bhts,bhsd->bthd', attn, u)
        out = out.reshape(b, t, d)
        return self.out(out)

class TransformerBlock(nn.Module):
    def __init__(self, dim, num_q_heads=8, num_kv_heads=None, rotary_base=10000, kernel_1=None, kernel_2=None):
        super().__init__()
        if kernel_1 is None:
            self.attn = MultiHeadAttention(dim, num_q_heads, num_kv_heads, rotary_base)
        else:
            self.attn = DeltaAttention(dim, num_q_heads, num_kv_heads, rotary_base, kernel_1, kernel_2)

    def forward(self, x):
        x = x + self.attn(x)
        return x

class Transformer(nn.Module):
    def __init__(self, vocab_size, dim, n_layers=1,
                 num_heads=8, num_kv_heads=None, rotary_base=10000, kernel_1=None, kernel_2=None):
        super().__init__()
        self.emb = Embedding(vocab_size, dim)
        self.blocks = nn.ModuleList([
            TransformerBlock(dim, num_heads, num_kv_heads, rotary_base, kernel_1, kernel_2)
            for _ in range(n_layers)
        ])
        self.unemb = Unembedding(dim, vocab_size)

    def forward(self, x):
        x = self.emb(x)
        for block in self.blocks:
            x = block(x)
        return self.unemb(x)


n_elements = 5
swap_pairs = [(i, j) for i in range(n_elements) for j in range(i+1, n_elements)]

def apply_swap(perm, swap_idx):
    i, j = swap_pairs[swap_idx]
    perm = list(perm)
    perm[i], perm[j] = perm[j], perm[i]
    return tuple(perm)

def generate_S5_data(k, num_samples):
    data = []

    for _ in range(num_samples):
        swap_sequence = [random.randint(0, len(swap_pairs)-1) for _ in range(k)]
        current_perm = tuple(range(n_elements))
        first_elements = []

        for swap_idx in swap_sequence:
            current_perm = apply_swap(current_perm, swap_idx)
            first_elements.append(current_perm[0])

        input_ids = torch.tensor(swap_sequence, dtype=torch.long)
        labels = torch.tensor(first_elements, dtype=torch.long)
        data.append((input_ids, labels))

    return data

class S5Dataset(Dataset):
    def __init__(self, k=4, num_samples=3):
        self.k = k
        self.num_samples = num_samples
        self.data = generate_S5_data(self.k, self.num_samples)
    def __len__(self):
        return self.num_samples

    def __getitem__(self, idx):

        input_ids, labels = self.data[idx]
        if idx == 0:
            self.data = generate_S5_data(self.k, self.num_samples)
        return {
            "input_ids": input_ids,
            "labels": labels
        }




def train_model(model, dataloader, run_name="run", log_dict=None,k=16,bs = 128):
    device = "cuda" if torch.cuda.is_available() else "cpu"
    model.to(device).to(torch.bfloat16)


    num_epochs = 4000

    optimizer = AdamW(model.parameters(), lr=0.01, weight_decay=0.01)
    num_training_steps = num_epochs * len(dataloader)
    lr_scheduler = get_scheduler(
        name="linear", optimizer=optimizer, num_warmup_steps=128, num_training_steps=num_training_steps
    )

    progress_bar = tqdm(range(num_training_steps), desc=f"Training {run_name}")
    acc_log = []
    loss_log = []
    loss = nn.CrossEntropyLoss()
    model.train()
    step = 0
    for epoch in range(num_epochs):
        for batch in dataloader:
            input_ids = batch["input_ids"].to(device)
            labels = batch["labels"].to(device)
            # print(input_ids, labels)
            logits = model(input_ids)
            loss = torch.nn.functional.cross_entropy(logits.view(-1, logits.shape[-1]), labels.view(-1))
            predictions = torch.argmax(logits, dim=-1)
            correct = (predictions == labels).sum().item()
            total = labels.numel()
            accuracy = correct / total

            loss.backward()
            optimizer.step()
            lr_scheduler.step()
            optimizer.zero_grad()

            acc_log.append(accuracy)
            loss_log.append(loss.item())

            progress_bar.set_postfix({"loss": f"{loss.item():.4f}", "acc": f"{accuracy:.4f}"})

            step += 1
            progress_bar.update(1)

        if accuracy > 0.999:
            k = k*2
            dataset = S5Dataset(k=k, num_samples=bs)
            dataloader = DataLoader(dataset, batch_size=bs, shuffle=False)
            print(f"step: {step}, length come to {k}")
    log_dict[run_name] = {"acc": acc_log, "loss": loss_log}



bs = 32*4
k = 16
dataset = S5Dataset(k=k, num_samples=bs)
dataloader = DataLoader(dataset, batch_size=bs, shuffle=False)


n_layers = 1
vocab_size = math.comb(n_elements, 2)
dim = 12
num_heads = 4
num_kv_heads = 1
model_base = Transformer(
    vocab_size=vocab_size,
    dim=dim,
    n_layers=n_layers,
    num_heads=num_heads,
    num_kv_heads=num_kv_heads,
    rotary_base=None,
)
kernel_cmb = [
    # ['linear', 'softmax'],
    #    ['round','round'],
    ['linear','linear'],
    ['exp', 'softmax'],
    #   ['softmax', 'softmax'],
    # ['round', 'softmax'],
]
delta_models = []
for kernels in kernel_cmb:
    delta_models.append(Transformer(
        vocab_size=vocab_size,
        dim=dim,
        n_layers=n_layers,
        num_heads=num_heads,
        num_kv_heads=num_kv_heads,
        rotary_base=None,
        kernel_1=kernels[0],
        kernel_2=kernels[1],
    ))


log_dict = {}
# train_model(model_base, dataloader, run_name=f'{n_layers} Layer {dim} Dim Transformer', log_dict=log_dict)
for i, kernels in enumerate(kernel_cmb):
    train_model(delta_models[i], dataloader, run_name=f'{n_layers} Layer {dim} Dim (GQA KV Dims=3). Delta Kernel {kernels[0]}+{kernels[1]} ', log_dict=log_dict,k=k,bs =bs)


plt.figure(figsize=(12, 6))
for key in log_dict:
    accs = log_dict[key]['acc']
    steps = np.arange(len(accs))
    plt.plot(steps, accs, label=key)

plt.title("Accuracy vs Training Steps")
plt.xlabel("Training Step")
plt.ylabel("Accuracy")
plt.ylim(0,1)
plt.legend()
plt.grid(True)
plt.show()
\end{lstlisting}

\section{Analytical Solution of Associative Memory}
\label{sec:associative_mem_analytical_solution}
Without loss of generality, consider the memory optimization objective in terms of the L2 norm:
\begin{align}
    \mathcal{L}_t(\vect{S}_{t-1}) = \frac{1}{2}\|\vect{S}_{t-1}\vect{k}_t - \vect{v}_t\|^2.
\end{align}

We seek the optimal memory matrix $\vect{S}^{*}$ by minimizing:
\begin{align}
    \vect{S}^{*} = \arg\min_{\vect{S}} \mathbb{E}_{\vect{x}}\left[\frac{1}{2}\|\vect{S}\vect{W}_k\vect{x}-\vect{W}_v\vect{x}\|^2\right],
\end{align}
where $\vect{x}$ denotes the single-layer input, $\vect{k} = \vect{W}_k\vect{x}$, $\vect{v} = \vect{W}_v\vect{x}$, and $\vect{W}_k$, $\vect{W}_v$ are fixed model weights at test time.

The objective can be reformulated as:
\begin{align}
    \mathcal{L}(\vect{S}) &= \mathbb{E}_{\vect{x}}\left[\frac{1}{2}\|(\vect{S}\vect{W}_k-\vect{W}_v)\vect{x}\|^2\right]  \nonumber \\
    &= \mathbb{E}_{\vect{x}}\left[\frac{1}{2}\vect{x}^\top(\vect{S}\vect{W}_k - \vect{W}_v)^\top(\vect{S}\vect{W}_k - \vect{W}_v)\vect{x}\right] \nonumber \\
    &= \mathrm{tr}\left(\frac{1}{2}(\vect{S}\vect{W}_k - \vect{W}_v)^\top(\vect{S}\vect{W}_k - \vect{W}_v)\mathbb{E}_{\vect{x}}[\vect{x}\vect{x}^\top]\right),
\end{align}
where $\mathrm{tr}(\cdot)$ denotes the trace operator. Since $\mathcal{L}(\vect{S})$ is a scalar, we have $\mathcal{L}(\vect{S}) = \mathrm{tr}(\mathcal{L}(\vect{S}))$. Taking the derivative of $\mathcal{L}(\vect{S})$ with respect to $\vect{S}$ yields:
\begin{align}
    \frac{\partial \mathcal{L}(\vect{S})}{\partial\vect{S}} &= \frac{\partial}{\partial\vect{S}}\mathrm{tr}\left(\frac{1}{2}(\vect{S}\vect{W}_k - \vect{W}_v)^\top(\vect{S}\vect{W}_k - \vect{W}_v)\mathbb{E}_{\vect{x}}[\vect{x}\vect{x}^\top]\right) \nonumber \\
    &= (\vect{S}\vect{W}_k - \vect{W}_v)\mathbb{E}_{\vect{x}}[\vect{x}\vect{x}^\top]\vect{W}_k^\top.
\end{align}

Setting the gradient to zero, we obtain the optimal solution:
\begin{align}
    \vect{S}^{*} = \left(\vect{W}_v\,\mathbb{E}_{\vect{x}}[\vect{x}\vect{x}^{\top}]\,\vect{W}_k^{\top}\right)\left(\vect{W}_k\,\mathbb{E}_{\vect{x}}[\vect{x}\vect{x}^{\top}]\,\vect{W}_k^{\top}\right)^{-1}.
\end{align}

\end{document}